\documentclass[10pt]{article} %
\usepackage[accepted]{tmlr}

\usepackage{amsmath,amsfonts,bm}

\def\eqref#1{equation~\ref{#1}}

\def\1{\bm{1}}

\DeclareMathAlphabet{\mathsfit}{\encodingdefault}{\sfdefault}{m}{sl}
\SetMathAlphabet{\mathsfit}{bold}{\encodingdefault}{\sfdefault}{bx}{n}

\newcommand{\R}{\mathbb{R}}

\DeclareMathOperator*{\argmin}{arg\,min}

\usepackage{hyperref}
\usepackage{url}

\title{Tags for DAGs:\\Graph Refinement with Meta-Informed Relations}

\author{\name Florian Peter Busch${}^\mathbf{*}$ \email florian\_peter.busch@tu-darmstadt.de \\
    \addr Department of Computer Science, Technical University of Darmstadt\\
    Hessian Center for AI (hessian.AI)
    \AND
    \name Moritz Willig${}^\mathbf{*}$ \email moritz.willig@cs.tu-darmstadt.de \\
    \addr Department of Computer Science, Technical University of Darmstadt
    \AND
    \name Florian Guldan \email  florian.guldan@stud.tu-darmstadt.de \\
    \addr Department of Computer Science, Technical University of Darmstadt
    \AND
    \name Kristian Kersting \email kristian.kersting@tu-darmstadt.de \\
    \addr Department of Computer Science, Technical University of Darmstadt\\
    Centre for Cognitive Science, Technical University of Darmstadt\\
    Hessian Center for AI (hessian.AI)\\
    German Research Center for AI (DFKI)
    \AND
    \name Devendra Singh Dhami \email d.s.dhami@tue.nl \\
    \addr Department of Mathematics and Computer Science, Eindhoven University of Technology
      }

\usepackage{booktabs}

\usepackage{bm}
\usepackage{amsmath}  %
\usepackage{amssymb}
\usepackage{amsthm}

\usepackage{algorithm}
\usepackage{algorithmicx}
\usepackage{algpseudocode}

\usepackage{tikz}
\usetikzlibrary{arrows.meta}
\usetikzlibrary{fadings}
\usetikzlibrary{calc}
\usetikzlibrary{positioning}

\usepackage{wrapfig}
\usepackage{caption}
\usepackage{cancel}

\usepackage{subcaption}
\usepackage{multirow}
\usepackage{rotating}

\usepackage[dvipsnames]{xcolor}

\usepackage[normalem]{ulem} %
\usepackage{multicol}
\usepackage{enumitem}

\usepackage[most]{tcolorbox}
\newtcolorbox{promptbox}[1][]
{
  enhanced,
  attach boxed title to bottom right={yshift=5mm,xshift=-3mm},
  colback=gray!5!white,
  colframe=gray!75!black,
  title={#1},
  fonttitle=\bfseries,
  coltitle=black,
  boxed title style={
    size=small,
    colback=gray!5!white,
    colframe=gray!75!black
  }
}

\DeclareMathOperator{\CVarDomains}{\bm{\mathcal{X}}}

\DeclareMathOperator{\sampleSpace}{\bm{\mathcal{Y}}}
\DeclareMathOperator{\CVars}{\mathbf{X}}
\DeclareMathOperator{\Edges}{\mathbf{E}}
\DeclareMathOperator{\Edge}{E}

\DeclareMathOperator{\tagset}{\mathbf{t}}
\DeclareMathOperator{\tg}{t}

\DeclareMathOperator{\tagFunc}{\text{tag}}
\DeclareMathOperator{\taggedFunc}{\text{tagged}}

\newcommand{\causesm}[2]{$#1{\rightarrow}#2$}
\newcommand{\causesmath}[2]{#1{\rightarrow}#2}

\usepackage{pifont}
\newcommand{\cmark}{\ding{51}}
\newcommand{\xmark}{\ding{55}}

\DeclareMathOperator{\N}{\mathbb{N}}

\newtheorem{definition}{Definition}
\newtheorem{assumption}{Assumption}

\newcommand\extrafootertext[1]{%
    \bgroup
    \renewcommand\thefootnote{\fnsymbol{footnote}}%
    \renewcommand\thempfootnote{\fnsymbol{mpfootnote}}%
    \footnotetext[0]{#1}%
    \egroup
}

\def\month{09}  %
\def\year{2026} %
\def\openreview{\url{https://openreview.net/forum?id=a399gBzXic}} %

\begin{document}

\maketitle
\extrafootertext{${}^\mathbf{*}$ Indicates shared co-first authorship.}
\begin{abstract}
Causal discovery has shifted from data-centric methods to hybrid strategies that integrate semantic knowledge from experts or large language models (LLMs). Such external information is vital for identifying causal structures beyond the Markov Equivalence Class (MEC), which data alone cannot resolve. However, expert availability is often limited, and LLMs frequently misidentify causal directions in specialized domains. To overcome such shortcomings, we propose a tag-based approach that leverages semantically meaningful labels while deriving causal directionality directly from data. Using variable-level tag assignments from available sources (e.g., LLMs), our tags for DAGs method learns from identifiable data structures to extract higher-level causal relations. These are then used to orient undirected edges, enabling causal discovery to move beyond the MEC without reliance on fallible external knowledge.
\end{abstract}

\section{Introduction}
Understanding variables not just as sole columns of data, but rather as semantic entities that can share common properties with each other is something well understood by humans, who have been shown to often leverage meta-information about them when making predictions \citep{bonawitz2010just,meltzoff2007infants,bailey2024causal}.
Knowledge about a particular semantic \textit{type} of a variable can be highly indicative of certain causal interactions. For example, demographic factors usually cause consumer behavior, socioeconomic policies typically drive individual financial behaviors, and biological traits are a cause of symptoms.
Even without domain knowledge, the relation between \textit{Bandwidth} and \textit{Latency} can be better explained by knowing meta-statistics of \textit{Infrastructure} commonly causing \textit{Performance Metric}.

\begin{wrapfigure}{r}{0.5\linewidth}
    \centering
    \includegraphics[width=0.95\linewidth]{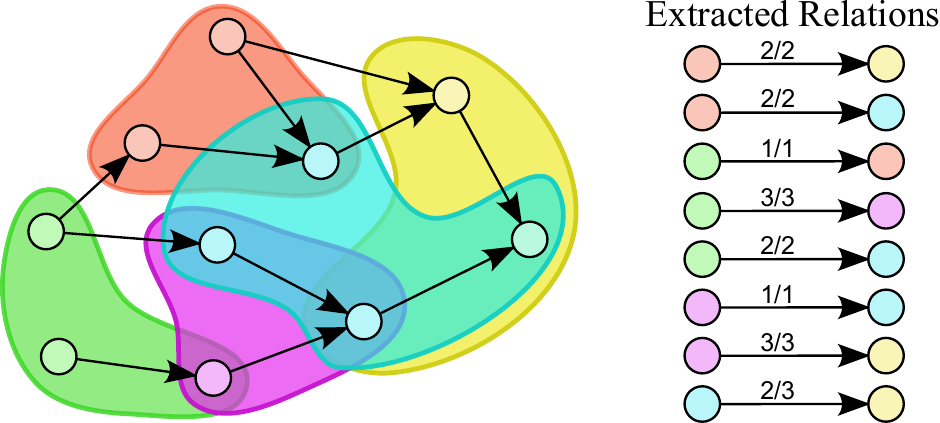}
    \vspace{2mm}
    \caption{\textbf{Collecting Tag Information.} Our tagging approach considers the assignment of multiple tags per variable, (1) allowing for the flexible assignment of light-weight meta-information, and (2) repurposing already discovered directed edges for identifying higher-level causal relations.}
    \label{fig:motivation}
\end{wrapfigure}

The task of causal discovery (CD) aims to recover causal structures from data \citep{spirtes1991algorithm,hoyer2008nonlinear,mooij2011causal,scholkopf2021toward} and while extensive consideration has been directed at recovering causal graphs from data alone, results derived from Pearl's do-calculus \citep{verma1990equivalence,frydenberg1990chain,pearl2009causality} limit the identification of causal structures from purely observational data, such that the direction of some edges can remain unknown. Such edges could, for example, be directed with the aid of experts or through additional experimental data. However, expert time is scarce, and additional data sources can be costly to obtain. To still improve graph recovery, more accessible forms of information need to be considered to further utilize the already available data. Towards this, we propose a tagging-based approach on top of standard causal discovery that further leverages meta-level information induced by relations formed between the semantic types of groups of variables (Fig.~\ref{fig:motivation}).
\vspace{3mm}

We follow the philosophy that it is more economical to leverage available data to its full extent by supplementing it with light-weight tagging information before requesting additional experimental data or repeatedly querying external experts for their support in providing additional information on individual causal relations.
The idea of using the semantic knowledge of the variables for causal discovery is not new.
Following the ongoing development of large language models (LLMs), causal discovery has been opened up to incorporate ubiquitous external knowledge that was previously hard to provide in practice.
With arguments for and against these models yielding notions of `true' causal understanding~\citep{zhiheng2022can,zevcevic2023causal,lampinen2024passive,joshi2024llms,chiunveiling}, LLMs have been utilized to predict the direction of edges in causal discovery.
While such information can be helpful, approaches based purely on LLMs have been found to struggle in practice when tasked with performing causal inference or discovery \citep{kiciman2023causal,jin2023cladder,zevcevic2023causal}.
More recently, hybrid approaches that jointly leverage existing data and the knowledge learned by LLMs have been shown to be more reliable \citep{brouillard2022typing,wang2024rcagent,clivio2025learning}.
Our tagging approach falls into this latter category. By making use of classical CD to recover the initial causal graph and only then to determine the edge direction based on tagging-induced statistics, our approach is `data first', circumventing possible deficits in LLMs' learned causal domain knowledge.

\vspace{2mm}
\textbf{Contributions.}
\begin{enumerate}
    \item We introduce a theory for extracting high-level causal relations from existing causal graphs by leveraging meta information in the form of tags and using this to narrow down the MEC.
    \item We propose an algorithm that implements this idea and includes several strategies for increasing robustness to graph faults and noisy tagging information.
    \item We demonstrate the conservative robustness of our tagging approach through ablations, inducing severe failures in the underlying recovered causal graph structure and the tagging assignment itself.
    \item We show how the discovered tag relations extract meaningful causal relations on abstract concepts.
\end{enumerate}
\vspace{2mm}
The code is made available at \url{https://github.com/olfub/tagging-causality}.
\clearpage

\section{Background and Related Work}

\textbf{Notation.} Throughout the paper, we denote univariate variables with uppercase letters $(X,Y)$ and their realizations as lowercase letters $(x,y)$. (Indexed) sets are indicated in boldface, ($\mathbf{X}$; $\CVarDomains$), and their realizations as lowercase boldface letters ($\mathbf{x}$). Individual elements are denoted with subscript, $X_i \in \mathbf{X}$; $X_i \in \CVarDomains$.

\textbf{Causal Discovery.} A causal graph $\mathcal{G}$ is an acyclic directed graph (DAG) that represents the structure of cause-effect relations over an indexed set of variables $\CVarDomains$ \citep{spirtes2000causation, pearl2009causality}. For every cause-effect relation between variables $X_i, X_j \in \CVarDomains$, individual edges $(X_i,X_j) \in \CVarDomains \times \CVarDomains$ in the graph point from the cause $X_i$ to the effect $X_j$. Causal discovery is concerned with identifying the underlying causal graph from data \citep{peters2017elements}.
Even when disregarding the presence of hidden confounders, causal models can only be discovered up to Markov equivalence \citep{verma2022equivalence,verma1992algorithm} from observational data without further assumptions. Completed partially directed acyclic graphs (CPDAGs) comprise all Markov equivalent DAGs that are admissible under the given data but leave some edges undirected \citep{andersson1997characterization}. 
The approach in this paper aims to refine CPDAGs in an attempt to further direct remaining undirected edges. A primary class of causal discovery methods is constraint-based algorithms, which utilize independence tests to recover the graph skeleton. We will rely on the well-known PC algorithm \citep{spirtes1991algorithm}, which performs skeleton and immorality discovery followed by Meek rules \citep{meek1995causal}. %
Another primary class is score-based algorithms that find the best graph by maximizing a particular scoring function. A common representative of this class of methods is greedy equivalence search (GES), which greedily adds and removes edges, starting from an empty graph \citep{chickering2002optimal}.

\textbf{Breaking Markov Equivalence.} Additional assumptions are needed to break Markov equivalence when trying to recover causal graphs from observational data and moving beyond CPDAGs by leveraging background knowledge~\citep{perkovic2017interpreting}.
Here, the provision of interventional data~\citep{hauser2012characterization,brouillard2020differentiable} or additional knowledge about the types of causal relations and noise can be used \citep{shimizu2006linear,hoyer2008nonlinear}. Furthermore, meta-considerations about the structure of graphs \citep{mansinghka2012structured} or edge directionality based on CPDAG conforming model counting \citep{long2023causal} have been proposed. Recently, the use of meta-information based on variable names has been suggested. Here, \citet{brouillard2022typing} put forward the idea of assigning particular \textit{types} to variables and using these types to enforce a \textit{type consistency} for edges such that one type can not be a cause and an effect of a second type at the same time.
While our work is close to \citet{brouillard2022typing} in that we equip variables with meta-information, our tag-based strategy has several advantages over the purely type-based approach.
For one, we can assign as many tags as we can derive from meta knowledge, without being constrained to a single type per variable.
Additionally, \citet{brouillard2022typing}'s \textit{type consistency} assumption, which enforces a unique causal edge direction between types, is often violated in practice and results in incorrect graphs.
Our tagging approach employs a probabilistic approach for dealing with tagging inconsistencies, calibrating the impact of individual tagging pairs on the recovered graph. Since `faulty' tag relations can be overruled by others, our algorithm becomes robust to faults in graph recovery and tag assignment.

\textbf{Relation to Causal Abstractions and Tiered Background Knowledge.} Section~\ref{sec:homogeneity} is concerned with the extraction of abstract causal relations from known, tag-annotated graphs. %
Intuitively, one could think of our tagging approach as a way to extract higher-level causal models \citep{rubenstein2017causal,beckers2019abstracting} with the help of individual tags as the new nodes. As such, our work would be related to works on cluster DAGs \citep{anand2023causal,wahl2023foundations,massidda2024learning,de2025cluster} that are primarily concerned with the identifiability of causal effects, or view tags as a form of self-inducing tiered constraints on the causal graph \citep{andrews2020completeness,mooij2020joint,bang2023we}. In contrast to these prior works, our approach does not require `hard' knowledge about any higher-level clusters or tiered background knowledge constraints. The `higher-level' tag statistics arise solely from variable annotation and causal graph discovery on the lower level. In this regard, concurrent work of \citet{brouillard2026causal} similarly exploits metadata-informed structures to improve the precision of causal discovery. Furthermore, our approach neither requires nor enforces assumptions about the acyclicity of the higher-level model. Most importantly, most prior works use the strong notion of `constructive' causal abstractions \citep{beckers2019abstracting} that assume compact, non-overlapping partitions of low-level variables. In these cases, typing of \citet{brouillard2022typing} would suffice, as every variable would be uniquely assigned to a partition.
With tagging, however, variables not only share tags but are usually assigned multiple tags at the same time, such that individual nodes do not belong to a unique cluster or tier, breaking the partition requirement of constructive abstractions. %

\textbf{Causal Discovery and LLM.} Multiple works have investigated the abilities of LLM to reason about causal concepts \citep{kiciman2023causal,jin2023cladder,zevcevic2023causal,gendron2024large} including the direct questioning of LLMs for causal discovery \citep{long2022can,kiciman2023causal,zevcevic2023causal,jiralerspong2024efficient} concluding mixed performance and motivating the need for approaches that jointly leverage statistical methods and LLMs \citep{choi2022lmpriors,long2023causal,mathur2024modeling,clivio2025learning}.

\begin{figure*}[t]
  \centering
  \includegraphics[width=0.8\textwidth]{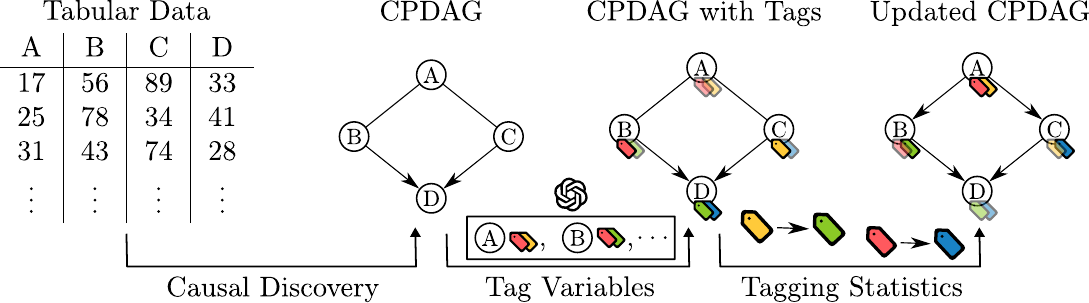}
  \caption{\textbf{Tag-Supported Causal Discovery.} The figure illustrates the tagging-based causal discovery described in this paper: First, a causal discovery algorithm such as PC or GES is used to obtain a CPDAG. Variables are tagged with semantically meaningful tags, for example, by using an LLM. Lastly, tag relations on the CPDAG are used to predict the direction of undirected edges.%
  }
  \label{fig:approach}
\end{figure*}

\section{Tagging Informed Edge Direction}
\label{sec:tagsInformative}

The underlying assumption of our approach is that observing causal directions between higher-level concepts can inform decisions about the direction of causality of their lower-level realizations.
In particular, the causal directions discovered by an initial CD algorithm can be used to compute preferences between higher-level tag relations, which can then be leveraged for predicting edge directions. Fig.~\ref{fig:approach} shows the basic approach.

One such approach by \citet{brouillard2022typing} assigns a single \textit{type} to each variable to gather the necessary statistics. Under the assumption of \textit{type consistency}, all edges between variables of type $A-B$ are assumed to be directed in the same way.
This rather strong assumption has several drawbacks. First, it can never provide information about the relations between variables of the same type.
Second, it cannot consider particular aspects shared between variables of different types since only one type can be assigned to any variable.
Therefore, in this work, we generalize the notion of types by associating multiple \textit{tags} $\tagset$ to every variable $X_i$. Tags are assigned by some external mechanism $\tagset := \text{tag}(X_i)$, such that individual tags $\tg^a \in \tagset$ are general concepts such as ``Symptom'' or ``Demographic Information'', expressed in natural language.
Tagging allows for drawing more evidence from a causal graph by considering multiple, possibly noisy tag relations between variables rather than a single type relation.

The starting point of the approach is an existing CPDAG.
An edge $E_{ij}$ between the variables $X_i, X_j$ either indicates $X_i$ causing $X_j$ ($\causesmath{X_i}{X_j}$), $X_j$ causing $X_i$ ($\causesmath{X_j}{X_i}$) or the edge being undecided ($X_i \text{---} X_j$). Edge direction is generally denoted as $d_{ij} \in \{-1,0,1\}$, where a value of $1$ indicates $\causesmath{X_i}{X_j}$, $-1$ indicates $\causesmath{X_j}{X_i}$ and $0$ indicates $X_i \text{---} X_j$. The following symmetry holds: $d_{ji} = -d_{ij}$.

\textbf{Notation:} The following section describes how tagging statistics are extracted from different subsets of variables. To differentiate various tags and tagging statistics, we use the following notation: (1) We write superscripts ($t^a, t^b, \dots$) to indicate unbound arbitrary tags. (2) We use subscripts to connect tags and variables: $\tagset_i$ is the set of tags assigned to variable $X_i$. (3) We use sub- and superscript in conjunction to reference specific tags of a variable's tag set: $t^a_i \in \tagset_i$.

\textbf{Structure.} This first part of the section is concerned with the theoretical modeling and the required underlying assumptions of the approach. The resulting practically applied formulas that no longer track noise statistics explicitly are presented Sec.~\ref{sec:taggingApproach}.

\textbf{Tag Informative Value.} In order to transfer information about the observation of one edge's direction onto another edge, the level of similarity between both needs to be assessed. Here, the overlap in the tagging sets of both variables serves as the common denominator of edge similarity. Consequently, the edge direction is assumed to be the result of some \textit{edge direction function} $f$ with $d_{ij} := f(\tagset_i, \tagset_j, N_{ij})$, taking the tag sets of both variables and an edge-specific noise term $N_{ij}$. %
Consequently, our approach assumes that $d_{ij}$ can be predicted from knowledge of $\tagset_i, \tagset_j$ and that the influence of individual $N_{ij}$ is rather small.

\begin{tcolorbox}[boxrule=0pt,frame hidden,sharp corners,enhanced,borderline west={1pt}{0pt}{black}]
    \textbf{Informal Idea:} The overall idea of the following \textit{Tag Informative Value} (Def.~\ref{def:tagInformativeValue}) is to compute the fraction of observed directed edges pointing from $t^a$-tagged nodes to $t^b$-tagged nodes (instead of pointing from $t^b$ to $t^a$).
\end{tcolorbox}

Beyond the informal description, the following notation is provided to formally capture the (possibly varying) tagging statistics within different subparts of the graph. These terms will be required for proving the correctness of the approach in the variable limit of $\left|\CVars\right|\to\infty$.

When observing a CPDAG, one is interested in quantifying how much the observation of a particular pair of tags attached to some edge informs about the directionality of other edges when observing that particular tag pair again. For this, a \textit{tag informative value} is defined, which measures the probability of observing a particular pair of tags with respect to a particular edge direction, $(t^a_i, t^b_j) \sim d_{ij}$.

The following tag informative value is defined as a conditional probability with respect to (some subset of) the graph. As we are trying to compute statistics on edge directions, a sample space for a node set $\sampleSpace \subseteq \CVarDomains$ is the set of all edges between variables in $\sampleSpace$ in that involve tags $\tg^a$ and $\tg^b$. We use superscript $\cdot^{\sampleSpace}$ to indicate expressions that are constraint to the (sub)set of edges between $\sampleSpace$. First, an auxiliary function is defined to determine whether any two variables are tagged with the tags of interest:
    $\taggedFunc(X_i,X_j,\tg^a,\tg^b) := (\tg^a \in \tagFunc(X_i)) \land (\tg^b \in \tagFunc(X_j))$.
Then, out of the set of edges between $\sampleSpace$, Eq.~\ref{eq:tagFilter} filters the set of directed edges where either endpoint carries tag $t^a$ and the other carries $t^b$:
\begin{align}
    \Edges^{\sampleSpace,\tg^a,\tg^b} = \{(X_i, X_j) \in \Edges~|~X_i \in \sampleSpace \land X_j \in \sampleSpace \land (\taggedFunc(X_i,X_j,\tg^a,\tg^b) \lor \taggedFunc(X_j,X_i,\tg^a,\tg^b))\}\label{eq:tagFilter}
\end{align}
and Eq.~\ref{eq:tagDirFilter} filters the edges where exactly the starting point carries tag $t^a$ and the endpoint carries tag $t^b$:
\begin{align}
    \Edges^{\sampleSpace,\tg^a\to\tg^b} = \{(X_i, X_j) \in \Edges~|~X_i \in \sampleSpace \land X_j \in \sampleSpace \land \taggedFunc(X_i,X_j,\tg^a,\tg^b)\}\label{eq:tagDirFilter}
\end{align}

In typing, directed edge sets are mutually exclusive ($\Edges^{\sampleSpace,\tg^a\to\tg^b}\cap\Edges^{\sampleSpace,\tg^b\to\tg^a}=\emptyset$) and exhaustive ($\Edges^{\sampleSpace,\tg^a\to\tg^b} \cup \Edges^{\sampleSpace,\tg^b\to\tg^a} = \Edges^{\sampleSpace,\tg^a,\tg^b}$), such that $\vert{}\Edges^{\sampleSpace,\tg^a\to\tg^b}\vert{} + \vert{}\Edges^{\sampleSpace,\tg^b\to\tg^a}\vert{} = \vert{}\Edges^{\sampleSpace,\tg^a,\tg^b}\vert{}$. Tagging, however, allows for overlapping sets ($\Edges^{\sampleSpace,\tg^a\to\tg^b}\cap\Edges^{\sampleSpace,\tg^b\to\tg^a}\neq\emptyset$), which requires the following tag information value to be normalized by the individual terms $\Edges^{\sampleSpace,\tg^a\to\tg^b}$ and $\Edges^{\sampleSpace,\tg^b\to\tg^a}$:

\begin{definition}
  \textbf{Tag Informative Value.}
  \label{def:tagInformativeValue}
  A pair of tags $(t^a, t^b)$ over a set of variables $\sampleSpace \subseteq \CVarDomains$ has tag informative value 
\begin{align*}
    I^{\sampleSpace}(t^a, t^b) := \frac{|\Edges^{\sampleSpace,\tg^a\to\tg^b}|}{%
    |\Edges^{\sampleSpace,\tg^a\to\tg^b}| + |\Edges^{\sampleSpace,\tg^b\to\tg^a}|}.
\end{align*}
\end{definition}
Put in simple words, $I^{\sampleSpace}(t^a, t^b)$ is the probability of any edge $X\text{---}Y$ in $\sampleSpace$ carrying tags $t^a, t^b$ to be directed as \causesm{X_i}{X_j}. In the following, we assume that the direction of an edge $d_{ij}$, annotated with tags $(\tg^a_i, \tg^b_j)$, is more likely to also be directed \causesm{X_i}{X_j} ($d_{ij} = 1$) if a high percentage of similar edges in $\sampleSpace$, carrying tags $\tg^a_i, \tg^b_j$, have previously been observed to point in the same direction \causesm{\tg^a_i}{\tg^b_j}.

Some particular combinations of tags $(t^a,t^b)$ might never be observed ($\Edges^{\sampleSpace,\tg^a,\tg^b} = \emptyset$), making $I^{\sampleSpace}(t^a, t^b)$ undefined. We filter out those cases by an indicator function
$o^{\sampleSpace}(t^a, t^b) := \mathbb{I}(\Edges^{\sampleSpace,\tg^a,\tg^b} \neq \emptyset)$, where $\mathbb{I}$ is $1$ if $\Edges^{\sampleSpace,\tg^a,\tg^b} \neq \emptyset$ and $0$ otherwise (with $I^{\sampleSpace}(t^a, t^b)$, in the absence of evidence, being defaulted to $0.5$). When observing a novel, previously undirected edge with tag sets $\tagset_i, \tagset_j$, we can now use the tag information value to reason about the true edge direction.

\begin{tcolorbox}[boxrule=0pt,frame hidden,sharp corners,enhanced,borderline west={1pt}{0pt}{black}]
    \textbf{Informal Idea:} For an undirected edge $X_i \text{---} X_j$, we infer the \textit{preferred edge direction} (Eq.~\ref{eq:direction}) as the average of tag information values between all tagging pairs $(\tg^a, \tg^b) \in \tagset_i \times \tagset_j$.
\end{tcolorbox}

When assuming the underlying edge direction function $f$ to be realized as the average tag informative value of all pairs of tags in $\tagset_i, \tagset_j$ --filtering out unobserved tagging pairs via the indicator function $I^{\sampleSpace}(\tg^a, \tg^b)$-- the \textit{preferred edge direction} is computed as follows:
\begin{align}
\label{eq:direction}
d'^{\sampleSpace}_{ij} = \frac{\sum_{(\tg^a, \tg^b) \in \tagset_i \times \tagset_j} I^{\sampleSpace}(\tg^a, \tg^b)o^{\sampleSpace}(\tg^a,\tg^b) + N_{ij}}{\sum_{(\tg^a, \tg^b) \in \tagset_i \times \tagset_j} o^{\sampleSpace}(\tg^a,\tg^b)}
\end{align}
Since $I^{\sampleSpace}(\tg^a, \tg^b)$ is already normalized, it does not require another normalization term in the denominator.
As $d'^{\sampleSpace}_{ij}$ is still a continuous quantity, we discretize it into one of the possible states captured by $d^{\sampleSpace}_{ij}$: $d^{\sampleSpace}_{ij}$ is set to $1$ (indicating \causesm{X_i}{X_j}) for $d'^{\sampleSpace}_{ij} > 0.5+\epsilon$, $d^{\sampleSpace}_{ij} := -1$ (indicating \causesm{X_j}{X_i}) for $d'^{\sampleSpace}_{ij} < 0.5-\epsilon$ and no decision on the direction ($d^{\sampleSpace}_{ij}=0$) is made whenever $0.5-\epsilon \leq d'^{\sampleSpace}_{ij} \leq 0.5+\epsilon$, for some small $\epsilon \in \R_{>0}$.

Different subsets of edges might yield different edge statistics. To reliably predict the direction of undirected edges from observed evidence, it must be assumed that edges in the directed and undirected cases both follow the same statistics:
\begin{assumption}
\label{assm:tagDistrConsist}
   \textbf{Tag Distribution Consistency.} For any two sufficiently large subsets $\sampleSpace', \sampleSpace'' \subseteq \CVarDomains$ and some small $\varepsilon \in \R_{>0}$, the tag informative values of both sets converge towards each other: $\forall (\tg^a, \tg^b) \in \tagset_i \times \tagset_j.$ $\lim_{|\sampleSpace'|,|\sampleSpace''|\to\infty} |I^{\sampleSpace'}(\tg^a, \tg^b) - I^{\sampleSpace''}(\tg^a, \tg^b)| < \varepsilon$.
\end{assumption}

The tag distribution consistency assumption asserts that information about the edge direction of yet undirected edges is related to that of already directed edges. I.e., it rejects scenarios in which the probability of directing edges correlates with their directionality. This assumption is usually not verifiable in practice, but we assume it holds under mild conditions, e.g., when applying causal discovery methods to domains with sufficiently diverse structures.
The experiments on our tested datasets seem to indicate support for this assumption, as we show that tag informative values between directed edges in the MEC do correlate with tag informative values based on the true direction of their undirected edges (App.~\ref{app:tagConsistencyAnalysis}).

\subsection{Tagging Approach}
\label{sec:taggingApproach}

Our algorithm follows the definitions of the tag informative value (Def.~\ref{def:tagInformativeValue}) and preferred edge direction (Eq.~\ref{eq:direction}) by first collecting statistics on the directional preferences between tagging pairs from the directed edges of a CPDAG and subsequently using this information to direct the remaining edges one by one. We assume the underlying CPDAG to be sound. Our algorithm, therefore, tackles the problem of refining the graph beyond the observational Markov equivalence with the use of additional tagging information. We do not intend to modify the underlying skeleton by adding or removing any existing edges based on our tagging statistics. %
Due to the error-prone nature of many causal baseline algorithms, the observed tagging statistics might be inaccurate. Several strategies are employed during the construction of our algorithm in Sec.~\ref{sec:strategies} to mitigate these issues. We provide the full pseudo-code of our algorithm in App.~\ref{sec:pseudo_code}.

\textbf{Granularity of Tags.} Part of the algorithm performance depends on the quality and sufficient number of provided tags. Providing too coarse tag sets generally groups together too many dissimilar edges, therefore weakening tag consistency and leading to fewer directed edges. Similarly, providing too specific tag sets will result in too few edges being selected to support individual tag pairs. Generally, our prompts (cf. App.~\ref{sec:llmPromptsAppx}) query the LLM to provide a moderate number of tags (``[\dots] Tags should be general enough to be applicable to multiple variables but specific enough to identify differences between similar variables. [\dots]''), to align tag assignment with the 'natural' complexity of the domain.

\textbf{Collect Tagging Evidence.} Similar to the previous tag informative value (Def.~\ref{def:tagInformativeValue}) we first count the number of edges $(X_i, X_j) \in \Edges$ carrying tags $(t^a, t^b)$ for all tag pairs between all tags, $\mathbf{T} = \tagset_1 \cup \dots \cup \tagset_N$, in the graph.
This is stored in an $\N^{|\mathbf{T}|\times|\mathbf{T}|}$-matrix $C$ with entries:
\begin{equation}
    C^{\sampleSpace}_{ab} := | \Edges^{\sampleSpace,\tg^a\to\tg^b} |
\end{equation}
The following convergence limit of Eq.~\ref{eq:limitThm} suggests that all statistics should be computed over the largest possible set of evidence, that is, over all variables ($C^{\CVarDomains}_{ab}$). If not stated otherwise, we therefore do not explicitly note the $\CVarDomains$-superscript in the following, and simply write $C_{ab}$ instead.
Different from the tag informative value, we do not directly normalize direction statistics, but retain the absolute edge counts in $C$. This helps us to later reject direction decisions that would be based on little evidence, e.g., a single observed edge pair, making the algorithm more robust in practice.
The observable finite sample corresponding to the tag informative value is then called $\hat{p}_{ab}$ and estimated as:
\begin{equation}
    \hat{p}_{ab} = \frac{C_{ab}}{C_{ab} + C_{ba}}
\end{equation}
Similar to $I^{\sampleSpace}(t^a, t^b)$, a $\hat{p}_{ab}$ of $0.5$ indicates no preference for either direction, while $\hat{p}_{ab} = 0$ and $\hat{p}_{ab} = 1$ indicate a full support for directing edges as \causesm{t^a}{t^b} or \causesm{t^b}{t^a}, respectively.

\textbf{Decision Rule.} Similar to before, and to eventually decide on the directionality of an edge $\Edge_{ij} \in \Edges$, we compute the \textit{evidence edge preference}, $Q(\Edge_{ij})$ as the finite sample correspondent to the preferred edge direction $d'^{\sampleSpace}_{ij}$ of Eq.~\ref{eq:direction}:
\begin{equation}
\label{eq:edgeQ}
    Q(\Edge_{ij}) = %
    \frac{\sum\nolimits_{(\tg^a, \tg^b) \in \tagset_i \times \tagset_j} \hat{p}_{ab} \cdot o(t^a,t^b)}{\sum\nolimits_{(\tg^a, \tg^b) \in \tagset_i \times \tagset_j} o(t^a,t^b)}
\end{equation}
All edges not introducing cycles are directed as $\hat{d}_{ij} := 1$ (the hat, again, indicates assignment based on the observed statistics, which is possibly different to the ground truth value) if $Q(\Edge_{ij}) > 0.5 + \alpha$; $\hat{d}_{ij} := -1$ if $Q(\Edge_{ij}) < 0.5 - \alpha$ and leaving the edge undirected ($\hat{d}_{ij} := 0$) otherwise for some $\alpha \in \R_{>0}$.
Due to acyclicity constraints and the possible application of Meek rules, the order of directing edges matters. To prioritize edges with higher confidence, the algorithm greedily directs edges with the highest $Q(\Edge_{ij})$ values first.

\textbf{Asymptotic Guarantees.} 
So far, we have illustrated our tagging approach and how it aims to predict the correct edge directions.
Under Assumption~\ref{assm:tagDistrConsist}, this can be shown to lead to correct edge predictions in the asymptotic limit.
Let the tag informative values $\mathbf{I}^{\sampleSpace} = \{I^{\sampleSpace}_1, I^{\sampleSpace}_2, \dots, I^{\sampleSpace}_n\}$ be sampled from a Beta distribution $\text{Beta}(\alpha, \beta)$. From the Central Limit Theorem, it follows that the variance of our prediction on $n$ is given as:
\begin{align}
    \text{Var}[&d'^{\sampleSpace}_{ij}] = \text{Var}\left[\frac{1}{n} \sum_{m=1}^n I^{\sampleSpace}_m\right] = \frac{\alpha \beta}{(\alpha + \beta)^2 (\alpha + \beta + 1) n}, \quad n = |\mathbf{t}_i \times \mathbf{t}_j|\label{eq:limitThm}
\end{align}
Full derivation in App.~\ref{app:theory}. An increasing number of tag pairs reduces the variance, thus improving the accuracy of the prediction.
In the limit, we have $\text{Var}[d'^{\sampleSpace}_{ij}]_{n \rightarrow \infty} = 0$, which even leads to perfect predictions $\mathbb{P}[d'^{\sampleSpace}_{ij} > 0.5] = 1$.
However, in practice, neither an infinite number of useful tag pairs nor enough edges to correctly estimate the tag informative scores are usually present. Similarly, our proof requires a strict i.i.d. assumption on the tag distribution, which might be violated in practice, but could be relaxed by adopting a more general notion of non-systematicity. Empirical evaluation indicates that the most severe errors usually stem from limited data artifacts (incorrect discovery of the CPDAG, insufficient tagging evidence due to a limited graph size, \dots), rather than obvious systematic tagging violations. We therefore expect the proof to `approximately' hold in practice under mild conditions. Further generalization might require a more careful reframing of the assumed (non-)systematicity of tagging and is left for future work. Next, we will introduce parameters aimed at mitigating these issues in practical scenarios with limited data.

\subsection{Robustness Strategies}
\label{sec:strategies}
While our convergence results are a strong theoretical basis for the tagging approach, data and tag limitations can result in incorrect edge directions.
To compensate for this, we introduce several strategies for leveraging the information provided by tags, while not relying on the perfect correctness of the CPDAG, and to increase robustness in settings with sparse and noisy tags.
In our experiments, we will show that our implementation is quite robust to such settings.

\textbf{Absence of V-Structures.} Taking further inspiration from the previous algorithm of \citet{brouillard2022typing} which enforced type consistency within variable triplets\footnote{Following the type consistency assumption, any unshielded triplet $A \mathrel{-} C \mathrel{-} B$ where $A$ and $B$ share a type that is different from $C$'s ($\tg^A = \tg^B \neq \tg^C$) must either be directed as a v-structure $A\to C\gets B$ or as a fork $A\gets C\to B$. Therefore, edges can even be directed when such a triplet does not represent a v-structure.},
we propose a variant of our method that considers nodes within triplets that contain the same tags as a source of information.
Following our tagging procedure, we assign tagging evidence to any tag pair supported by an identified v-structure $A \rightarrow C \leftarrow B$.
If an unshielded triplet is not identified as a v-structure, we assign evidence to any tag pairs in the opposite direction (corresponding to the fork structure $A \leftarrow C \rightarrow B$) for only those tag pairs where one tag is shared across both $A$ and $B$.
We collect evidence using this procedure and then start directing edges on the skeleton, where no edges are yet directed, not even v-structures.
The underlying idea is that, as all evidence still originates from v-structures, this information can be leveraged and combined in such a manner that the occasional incorrectly identified v-structure can possibly be overruled by overwhelming tag evidence from other structures.
We denote this variant as `AntiV' in our experiments, but find that it generally performs worse than directing edges on the CPDAG.

\textbf{Calibrating Conservativeness of the Algorithm.} Our theoretical guarantees hold in the sample limit and under sufficiently large graphs. Practical applications, however, are commonly limited by the availability of data and the size of the graph. Under these circumstances, evidence statistics might become noisy due to either erroneous tagging assignments or errors of the initial discovery graph.
The actual $\alpha$ value mentioned in conjunction with Eq.~\ref{eq:edgeQ} should depend on the system's inherent noise level and controls the `\textit{conservativeness}' of the prediction. A low $\alpha$ leads to more, but possibly noisy, decisions, while a high $\alpha$ can make the algorithm abstain too often, leaving most undirected edges unchanged.

The optimal $\alpha$-threshold for a newly discovered CPDAG is calibrated via leave-one-out cross-validation on the CPDAG itself. For every split with a held-out edge $\Edge_{ij}$, edges of the calibration set are used to compute the tagging statistics and, in turn, to predict the hold-out edge. A score function $s(\Edge^{GT}, \Edge^{pred})$ is introduced that returns an error of $0$ for correct and $1$ for incorrect predictions. To prevent the algorithm from always abstaining from making a decision, undirected edges are scored with an error of $0.5$. Possible values for $\alpha$ are all $Q(\Edge_{ij})$ encountered over all splits, $\bm{\alpha} := \bigcup_{i,j\in N\times N, i\neq j}Q(\Edge_{ij})$. For every $\alpha' \in \bm{\alpha}$, the cumulative score $S(\alpha') := \sum s(\Edge_{ij}^{GT}, \hat{d}_{ij}(Q(\Edge_{ij}))$ is computed and the optimal $\alpha^*$ is selected as $\alpha^* := \argmin_{\alpha' \in \bm{\alpha}}S(\alpha')$.
Importantly, this means that no new information (e.g., from the ground truth graph) is employed, as the entire procedure is applied on the provided CPDAG before further edge orientations begin.

\textbf{Minimal Evidence and Meek Rules.} In addition to the considered $\alpha$ conservativeness, we enforce the evidence for the final $Q(\Edge_{ij})$ to stem from either at least one or two different evidence edges. Our experiments found that a configuration with a two-edge boundary performed slightly more robustly.
Further, the algorithm allows for the optional application of Meek rules after each edge-direction step. The four best configurations, however, do not require their application.

\textbf{Redirecting Edges.} The recovered CPDAG serves as the baseline of our approach. However, it might contain errors, stemming from too little data or measurement noise. While we assume the overall graph is trustworthy, we might also use our tagging statistics to detect edges that are likely to be incorrectly inferred. Therefore, we optionally allow our algorithm to redirect such wrongly inferred edges before moving on to the undirected edges. (Edges that would induce cycles are still excluded.) Upon doing so, we allow the redirection to only take place at edges with a decision boundary of $Q(\Edge_{ij}) \geq 0.6$. As one might not trust the evidence stemming from the edge under consideration, one might not include it in the statistics.
Again, this is a hyperparameter that is up to the user to decide.
Our experiments showed a small preference for a conservative redirection strategy that includes the edge in the evidence.
Lastly, we provide two variants of a \textit{redirection strategy}, with either updating or not updating the evidence after redirecting edges.

\section{Experiments}
\label{sec:experiments}

We conduct an extensive empirical evaluation over eleven datasets. We consider all possible combinations of the described parameters, resulting in 560 runs per dataset and seed (-- 61,600 evaluations in total). We employ the PC \citep{spirtes1991algorithm} and GES \citep{chickering2002optimal} algorithms for initial CPDAG discovery, as those are the most common representatives of their respective classes of constraint-based and score-based CD algorithms.
Our tagging approach, on average, achieves the best results on top of GES and also outperforms variants of the \textit{naive} and \textit{majority} typing approaches by \citet{brouillard2022typing}.
We answer the following research questions:
\textbf{Q1.} Is tagging information a suitable indicator for determining edge directions with competitive performance?
\textbf{Q2.} Are tagging approaches robust under faulty information?
\textbf{Q3.} Does the Tag Distribution Consistency assumption (Assm.~\ref{assm:tagDistrConsist}) hold, and can it be used to identify higher-level causal relations?

\textbf{Tagging Annotation.}
The key driver of our algorithm's performance is meaningful variable annotations with sets of tags. These sets could, for example, be distilled by leveraging meta-information about the domain, such as variable names or short variable descriptions \citep{long2023causal}. 
In our experiments, we use LLMs as stand-in experts to provide a fair comparison across datasets, specifically GPT-5.2 \citep{openai2026gpt52}, Claude Opus 4.6 \citep{anthropic2026claude46}, Gemini 3 Pro \citep{google2026gemini3pro}, Llama 3.3 70b Instruct \citep{meta2025llama33}, Qwen3.5 397b-a17b Instruct \citep{qwen35blog}, Minimax M2.5 \citep{minimax2026m25}, and GLM-5 \citep{zeng2026glm}.
We make use of the fact that most modern LLMs are typically familiar with the given domains and therefore provide a good baseline across all datasets.
Results are queried with the respective standard parameters per model and a temperature of 1.0.
The LLMs are presented with a list of all variables per dataset and prompted to assign each variable to (possibly) multiple tags or a single type.
The exact prompts are given in App.~\ref{sec:llmPromptsAppx}. Answers from all LLMs are included in the code repository.

\textbf{Datasets \& Evaluation.} We evaluate algorithm performance over several datasets of the bnlearn repository~\citep{scutari2010learning,Taskesen_Learning_Bayesian_Networks_2020}. Additionally, we also include a Lung Cancer dataset (LUCAS; \citet{guyon2011time}). As we rely on language models to produce tags, we filter datasets that lack meaningful variable names. Eventually, we evaluate on the following datasets: Asia, Cancer, Earthquake, Survey, Alarm, Child, Insurance, Hailfinder, HEPAR2, Win95pts, and LUCAS.
All experiments are carried out over 10 seeds with 10,000 randomly sampled datapoints per dataset. (We use the 2,000 provided ground-truth samples for the LUCAS dataset.)
We include additional comparisons in the appendix (App.~\ref{sec:all_evals}), where we also compare to a different class of causal discovery approaches: orienting edges by prompting LLMs for the correct orientation directly. Compared to our tagging approach, such approaches require specific knowledge of causal directions rather than the annotation of variables independent of their causal relation. Consequently, these families of methods rely on distinctly different assumptions.

We measure the individual performance of edge prediction using Structural Hamming Distance (SHD; \citet{tsamardinos2006max}), Structural Intervention Distances (SID; \citet{peters2015structural}), as well as Precision, Recall, and $F_1$-score of all directed edges. SHD measures the correct or incorrect prediction of edges, with classical SHD increasing its score for every incorrect prediction. We also report an alternative ($\text{SHD}_\text{double}$), which counts a doubled error when encountering edges directed in the anti-causal direction. SID measures the similarity between two graphs based on their interventional distributions. Due to the exponential runtime scaling of SID, we could not evaluate it on the largest datasets: Hailfinder, HEPAR2, and Win95pts. (Evaluation did not conclude after 5 days.) %
We report upper and lower bounds for SID, which consider undirected edges to either be counted as correct or incorrect, respectively. Directing further edges can, therefore, never improve the lower SID bound but can reduce the upper bound.

\subsection{Tagging Performance}
\label{sec:main_eval}

\begin{table*}[t]
    \centering
\resizebox{\textwidth}{!}{
\begin{tabular}{l|ccccccc}
& SHD & SHD\textsubscript{double} & SID\textsubscript{min} & SID\textsubscript{max} & Precision & Recall & F\textsubscript{1} \\
 & Ranks {\footnotesize($\downarrow$)} & Ranks {\footnotesize($\downarrow$)} & Ranks {\footnotesize($\downarrow$)} & Ranks {\footnotesize($\downarrow$)} & Ranks {\footnotesize($\downarrow$)} & Ranks {\footnotesize($\downarrow$)} & Ranks {\footnotesize($\downarrow$)} \\
\hline
PC & $4.32 {\scriptstyle \pm 0.38}$ & $4.27 {\scriptstyle \pm 0.33}$ & $3.15 {\scriptstyle \pm 0.28}$ & $3.98 {\scriptstyle \pm 0.48}$ & $3.18 {\scriptstyle \pm 0.39}$ & $4.41 {\scriptstyle \pm 0.34}$ & $4.45 {\scriptstyle \pm 0.31}$ \\
GES & $3.54 {\scriptstyle \pm 0.51}$ & $3.07 {\scriptstyle \pm 0.47}$ & $1.27 {\scriptstyle \pm 0.13}$ & $3.17 {\scriptstyle \pm 0.51}$ & $2.69 {\scriptstyle \pm 0.51}$ & $3.17 {\scriptstyle \pm 0.42}$ & $3.13 {\scriptstyle \pm 0.51}$ \\
\hline
Typed-PC (Naive) & $3.42 {\scriptstyle \pm 0.34}$ & $3.62 {\scriptstyle \pm 0.33}$ & $3.10 {\scriptstyle \pm 0.27}$ & $3.20 {\scriptstyle \pm 0.33}$ & $3.07 {\scriptstyle \pm 0.42}$ & $3.47 {\scriptstyle \pm 0.34}$ & $3.59 {\scriptstyle \pm 0.39}$ \\
Typed-PC (Maj.) & $2.97 {\scriptstyle \pm 0.28}$ & $3.25 {\scriptstyle \pm 0.34}$ & $3.40 {\scriptstyle \pm 0.38}$ & $3.27 {\scriptstyle \pm 0.37}$ & $2.84 {\scriptstyle \pm 0.35}$ & $3.05 {\scriptstyle \pm 0.26}$ & $3.20 {\scriptstyle \pm 0.24}$ \\
\hline
Tagged-PC (AntiV) & $3.60 {\scriptstyle \pm 0.45}$ & $4.05 {\scriptstyle \pm 0.38}$ & $4.54 {\scriptstyle \pm 0.32}$ & $4.05 {\scriptstyle \pm 0.46}$ & $4.25 {\scriptstyle \pm 0.40}$ & $3.67 {\scriptstyle \pm 0.37}$ & $4.11 {\scriptstyle \pm 0.40}$ \\
Tagged-PC & $2.66 {\scriptstyle \pm 0.32}$ & $3.00 {\scriptstyle \pm 0.41}$ & $2.95 {\scriptstyle \pm 0.49}$ & $2.67 {\scriptstyle \pm 0.50}$ & $3.07 {\scriptstyle \pm 0.48}$ & $2.89 {\scriptstyle \pm 0.28}$ & $2.99 {\scriptstyle \pm 0.38}$ \\
Tagged-GES & $\mathbf{2.11} {\scriptstyle \pm 0.55}$ & $\mathbf{1.73} {\scriptstyle \pm 0.39}$ & $\mathbf{1.11} {\scriptstyle \pm 0.09}$ & $\mathbf{1.68} {\scriptstyle \pm 0.37}$ & $\mathbf{2.24} {\scriptstyle \pm 0.51}$ & $\mathbf{1.77} {\scriptstyle \pm 0.42}$ & $\mathbf{1.70} {\scriptstyle \pm 0.47}$ \\
\end{tabular}
}
\caption{\textbf{Average \underline{Ranks} for Best Configurations over all Datasets (lower is better).} Tagging improves upon PC and GES, and both tagging methods outperform their typed counterparts. Tagged-GES achieves the best results overall, yielding the lowest ranks across all metrics.
While the AntiV baseline offers a slight improvement over PC, it is surpassed by both typing and tagging approaches.
Detailed absolute results are presented in Tab.~\ref{tab:detailedOverallResultsAppx} in the appendix. ${}_{\pm y}$ indicates standard deviation.}
\label{tab:avgBestRankedResults}
\end{table*}

Results over all parameter configurations permitted by our method are evaluated on all datasets. For comparison with the PC, GES baselines, and typing approaches, we select the best-performing configuration for each method individually.
We will use the best-performing parameters in the follow-up ablations. Because SHD and SID metrics scale with the number of variables and the sparsity of the graphs, we use the average F\textsubscript{1}-score across all datasets to identify the best-performing configuration.
We find the overall best strategy is obtained by not employing the AntiV approach, not removing single tags, without the specificity prior, without applying Meek rules after every step, redirecting some existing edges, and using the tagging sets produced by Claude 4.6.
More information on the best performing parameter configurations can be found in App.~\ref{sec:best_configs}

We compare our method with PC and GES baselines (Tagged-PC and Tagged-GES) and with standard PC and GES, as well as with previous typing approaches using both naive and majority strategies.
\citet{brouillard2022typing} presented another \textit{t-Propagation} approach which, however, requires strict type consistency. As strict type consistency rarely holds for any non-constructed types, we omit this last step and additionally compare to this approach in Tab.~\ref{tab:detailedOverallResultsAppx} in the appendix.)
Contrary to the experimental section in \citet{brouillard2022typing}, which employed synthetic types, matching the ground-truth causal ordering, we again use LLMs to provide types for all variables.
Because SHD and SID scale non-linearly with dataset size, we rank absolute results per dataset and average the resulting ranks across all seeds. Results are presented in Tab.~\ref{tab:avgBestRankedResults}. A detailed report on all absolute metric scores across all datasets is provided in Tab.~\ref{tab:detailedOverallResultsAppx} in the appendix.

\textbf{Results.} We find that GES approaches generally perform better across most metrics. \textit{Both tagging approaches, ignoring the AntiV variant,  consistently improve upon their PC and GES baselines}. 
When comparing typing approaches, the `majority' variant performs somewhat better than the `naive' variant.
Among the PC-based approaches, Tagged-PC improves upon both PC and the typing approaches in most or all metrics.
Inspecting the datasets individually (Tab.~\ref{tab:detailedOverallResultsAppx}), we observe a larger benefit of Tagged-PC on datasets with a medium or larger number of variables.
This aligns with our theoretical results of Sec.~\ref{sec:taggingApproach} on the benefits of considering tag sets on sufficiently large graphs.
Further note that we also employ LLM-based directing strategies (Tab.~\ref{tab:detailedOverallResultsAppx}). With this comparison, Tagged-GES is no longer the clearly superior method; it both ``wins'' and ``loses'' against the direct LLM orientation on different datasets. Still, we believe our method is valuable for two main reasons: First, our assumption is less strict, only requiring meta information on variables, not on exact variable relations (causal information). Second, the tested datasets are well-known and have likely been part of the training data for the tested LLMs, making data contamination likely.

Comparing overall performance of the methods in Tab.~\ref{tab:avgBestRankedResults}, \textit{we find that average Tagged-GES results consistently rank the best across all methods and metrics}, even including $\text{SID}_{\text{min}}$ and Precision, which are both metrics that disproportionately favor keeping edges undirected. While areas of standard deviation sometimes overlap, Tagged-GES performs clearly better on $\text{SID}_{\text{max}}$, which is commonly considered an important measure for the ability of causal discovery methods to truthfully recover causal graphs. The results indicate our approach's ability to demonstrably reduce the Markov equivalence class of the initial CPDAGs and, therefore, reliably improve the given causal graph towards the underlying ground truth. We find the AntiV variant, adopted from the typing approaches, ranks worst among all typing and tagging methods, indicating that the heuristic rather harms than helps with variable tagging.
Overall, this answers \textbf{Q1} affirmatively.

\textbf{Further Experiments.}
We conducted several further experimental analyses using the same setup as in Tab.~\ref{tab:avgBestRankedResults}, each focusing on specific aspects.
For one, App.~\ref{sec:loocv} chooses the best-performing parameter configuration following a leave-one-out cross-validation procedure.
Results do not change substantially compared to Tab.~\ref{tab:avgBestRankedResults}.
We investigate different calibration thresholds in App.~\ref{sec:vary_thresholds}, observing how stricter thresholds increase abstentions generally and cause incorrect predictions in particular to vanish.
Lastly, applying Laplace correction on all tag pairs before computing the tag informative values results in a marginal, but insubstantial decrease in performance (App.~\ref{sec:laplace_correction}).

\textbf{Validity of Tag Distribution Consistency.} Our algorithm's performance is fundamentally linked to the assumption of Tag Distribution Consistency (Assm.~\ref{assm:tagDistrConsist}). While we demonstrated the general benefits of the algorithm against several baselines, quantifying adherence to the assumed tag distribution consistency provides important insights into the scenarios under which the algorithm can operate. The ability to transfer the knowledge gained from the directed edges of some CPDAG to predict the undirected edges primarily depends on the correlation (or difference) of the tag informative values between the two sets, $\text{r}(I^{\text{dir}}(\tg^a,\tg^b), I^{\text{undir}}(\tg^a,\tg^b))$. An on-average high correlation indicates that information obtained from the observed directed edges with observed high tag informative values generally provides an informative signal for the undirected edges. In contrast, pairs with low tag informative values contribute less (and the algorithm is similarly not missing out on using potentially available information). As with common interpretation, values near 0.0 indicate an uninformative correlation, and negative values indicate signals harming prediction quality.
Full experimental setup and results are described in App.~\ref{app:tagConsistencyAnalysis} and Tab.~\ref{tab:pearson_tags}, respectively.
The best-performing LLM (Claude 4.6) has an average $r=0.395$ (with a 95\%-confidence interval of $[0.25,0.52]$). Overall, correlations are moderate, but far from perfect. While LLama 3.3 has a stronger correlation at $r=0.572$, Claude 4.6 has a much larger number of tag pairs to leverage, which might indicate that the amount of evidence matters slightly more than the tag-pair ``quality''.

\textbf{Influence of LLM.}
We now take a closer look at the impact of tagging annotations produced by individual LLMs on the resulting performance. In a first evaluation, we select the best configurations of hyperparameters measured by $F_1$ score \textit{per LLM}. This allows every LLM to choose its optimal parameter setting, which might not align with the Claude 4.6 configuration, which was determined to be the globally best configuration. The selected parameter configurations per LLM and their $F_1$ scores are shown in Tab.~\ref{tab:best_f1_configs}. We observe no big qualitative differences between models. The resulting $F_1$ scores lie closely together, demonstrating that our algorithm is robust to changes in tagging sets. (The impact of other parameters is also relatively small; see further analysis of this in App.~\ref{sec:implementationDetails}.) In a second evaluation, we consider the effects of varying LLM within a model family (c.f. App.~\ref{sec:llm_different_versions}). We find that not a single LLM version is substantially better or worse than other versions of the same family on PC or GES. Lastly, we consider the influence of different prompt formulations and requesting fixed numbers of tags on the algorithm performance (Apps.~\ref{sec:llmPromptsAppx} and \ref{sec:llm_prompting}). The results show that, while our prompting strategy performs best, the other prompting strategies still perform well, remaining to be ranked similarly to the default prompt.

\subsection{Robustness Analysis}
\label{sec:faults}

\begin{figure*}[t]
    \centering
    \includegraphics[width=0.93\linewidth]{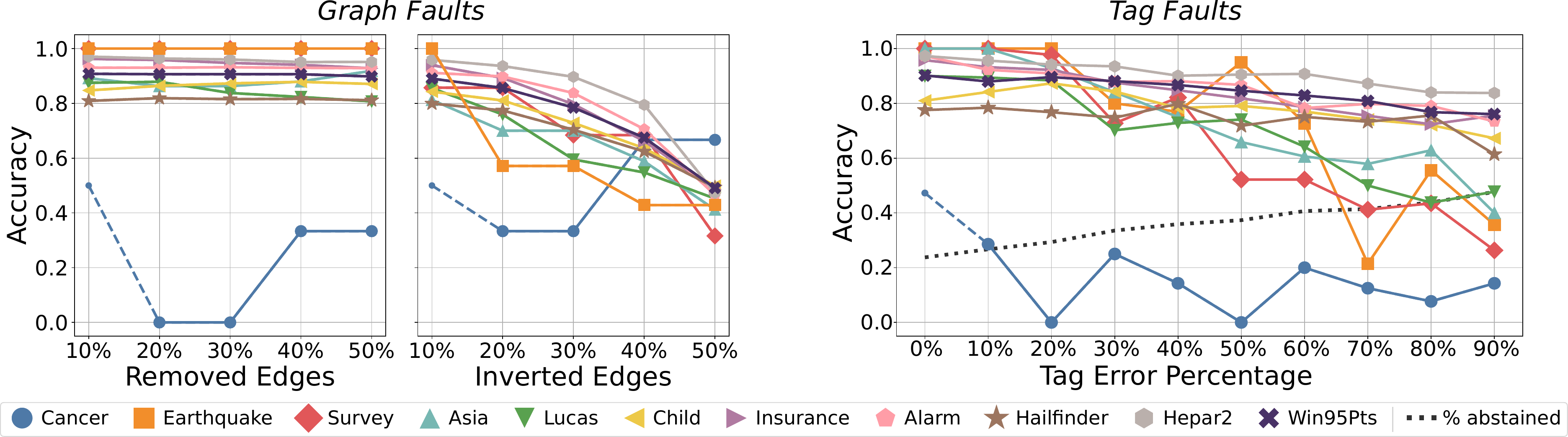}
    \caption{\textbf{Robustness to Faults.} Tags generated by Claude 4.6.
    Smaller point and dashed lines indicate 0 edge directions (defaulted to 50\%).
    \textbf{Graph Faults (left):} Removing edges has little effect on performance, whereas inverting edges has a stronger negative impact on accuracy. \textbf{Tag Faults (right):} Accuracy decreases with increasing noise, but remains substantially better than random chance. The dotted line shows an increasing abstention rate.
    Discussion on outliers (Cancer dataset) and other LLMs in Apps.~\ref{sec:all_faults} and \ref{app:tag_noise}.}
    \label{fig:ablationX}
\end{figure*}

In Sec.~\ref{sec:strategies}, we proposed multiple strategies for dealing with noisy settings.
These are put in place to make the algorithm direct edges when evidence is strong and abstain when evidence is weak.
To validate this behavior, we stress-test how well tagging performs under settings of increasingly noisy/faulty graphs.
We present the results for a select few experiments here and include full experiments in App.~\ref{sec:all_results}.

Our tagging heuristic can possibly serve to complement any CD algorithm that produces partially directed graphs.
While our experiments indicate that it performs well on error-free graphs (Apps.~\ref{sec:eval_cpdag} and \ref{sec:undirected}), we now examine the impact of different fault types.
To this end, we gradually introduce faults into the ground truth graph by removing or flipping several edges.
Subsequently, we undirect one of the remaining edges, disabling Meek rules to focus fully on the tag statistics.
We either evaluate all possible problem instances or sample 20,000 random instances for larger graphs.
Results for the removal and flipping of edges are shown in Fig.~\ref{fig:ablationX} (left). We find that tagging remains stable under the mild induction of faults on larger datasets if these do not insert incorrect edge directions.
Performance is observed to suffer most from flipped edges, which introduce erroneous information and even affect larger datasets, whereas edge removal keeps performance stable, except for the very small Cancer dataset.
Note that at 50\% inverted edges, half the evidence is ``correct'' and half ``incorrect''.
The resulting drop in performance to around 50\% is, thus, unsurprising.

As LLM-generated causal knowledge can be unreliable, we also include two experiments that investigate performance under noisy tag conditions.
To this end, we add $X\%$ noise by removing $\frac{X\%}{2}$ existing tags, and adding $\frac{X\%}{2}$ tags to variables that were not assigned this tag by the LLM.
We then undirect one edge and measure the accuracy of our edge direction based on tags, following the setup from the graph faults experiment.
In Figure~\ref{fig:ablationX} (right), we see that noisy tags slightly reduce the accuracy, but performance still remains relatively high.
This can be explained by our aggregation of tag relations, which ensures that ``bad'' tags have effectively random accuracy, so that actually informative relations can tip the otherwise uncertain decision in the right direction.
We also repeated our main evaluation from Sec.~\ref{sec:main_eval} with noisy tags of $50\%$, following the same noise setup as in the ablation experiment.
Even in this realistic setting, performance only drops slightly compared to the noise-free tags (Tab.~\ref{tab:noisy_eval}).
We thus conclude that tagging further improves graph quality and achieves robust performance, even under the introduction of faults in the initial graph and when noisy tags are used, thereby answering \textbf{Q2} positively.

\subsection{Mining Abstract Causal Relations}
\label{sec:homogeneity}

In the previous sections, we considered how the informative value $I^{\sampleSpace}(t^a, t^b)$ (Def.~\ref{def:tagInformativeValue}) of tags can help direct edges within graphs.
Now, we examine the structure of this knowledge and focus on the abstract causal relations.
We consider the ground-truth graphs as a data source for systematically discovering relations between abstract causal concepts, as realized by the annotated tags. Not all tags represent abstract concepts that are suited to infer causal relations. In Sec.~\ref{sec:tagsInformative}, we discussed the Tag Distribution Consistency assumption (Assm.~\ref{assm:tagDistrConsist}) that relates the tag informative value of particular tagging pairs to their predictive performance. In the following, we identify tag pairs that demonstrate high consistency within the dataset in order to discover common relations. As the tag informative value $I^{\sampleSpace}(t^a, t^b)$ was previously used to infer edge directions from tag pairs, we now refer to it as (tag pair) \textit{homogeneity} to indicate the process of identifying stable tag pairs from known edge directions. In our previous experiments, we implicitly showed that this assumption holds for our selected datasets, as without it, edge recovery would not have been as successful. We take a closer look at the actual tagging homogeneity within the ground-truth graphs and invert the direction of inference by mining abstract causal relations from these observations.

\begin{figure*}[t]
    \centering
\includegraphics[width=0.9\linewidth]{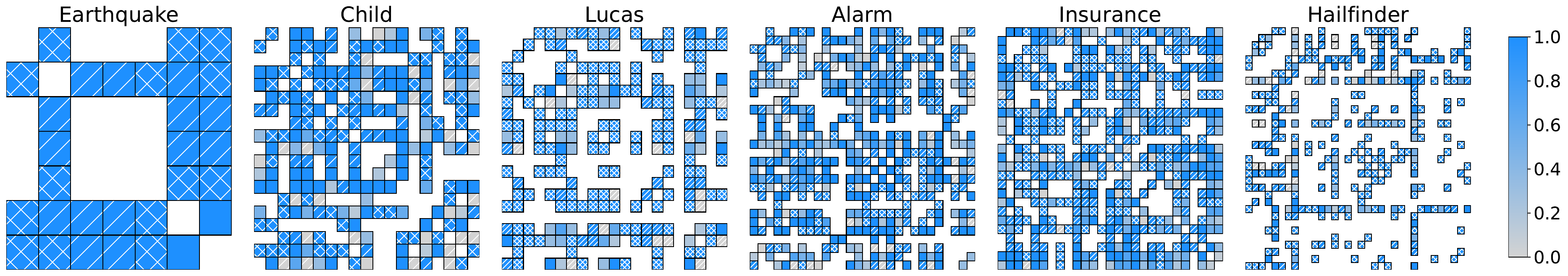}
\vspace{1.5mm}
\hrule
\vspace*{-2.5mm}
\resizebox{\textwidth}{!}{
\begin{minipage}[t]{\textwidth}
\begin{align*}
            \text{``Security Related''} &~\rightarrow~ \text{``Notification''} && (100\% ~/~ 3~~) &
            \text{``Heritable''} &~\rightarrow~ \text{``Symptom''} && (100\% ~/~ 3~~) \\
            \text{``Context''} &~\rightarrow~ \text{``Scenario Related''} && (100\% ~/~ 5~~) &
            \text{``Physical Hardware''} &~\rightarrow~ \text{``Intermediate State''} && (100\% ~/~ 12~~)
        \end{align*}
\end{minipage}
}
    \caption{\textbf{Top: Tag Homogeneity for Claude 4.6.} A tag relation has high homogeneity (blue/1) if variables of one tag consistently cause the other and low homogeneity (gray/0) if the cause-and-effect relationship between these variables is random. Blank fields have no evidence; low-evidence fields are hatched. Hatched areas might appear white on screen due to anti-aliasing. We observe high homogeneity throughout all datasets. See App.~\ref{sec:homogeneity_all} for high-resolution plots.
    \textbf{Bottom: Tag Relations.} Considering the most prominent tag pairs per dataset reveals high-level mechanisms present in the domain. We list tag pairs, with accuracy and support in brackets. Full list in App.~\ref{sec:homogeneity_all}.}
    \label{fig:combined_figure_table}
\end{figure*}

Tagging homogeneity of Claude 4.6 tags across different datasets is shown in Fig.~\ref{fig:combined_figure_table} (top).
We find that tagging pairs, when they exist, generally yield high homogeneity with reasonable support across all considered datasets. As we initially instructed the LLMs to produce high-level tags, finding tag pairs with high homogeneity can, therefore, be thought of as extracting relations between high-level concepts.
In Fig~\ref{fig:combined_figure_table} (bottom), we list some of the most exemplary tag pairs with the highest homogeneity and support among their respective datasets (further examples are presented in App.~\ref{sec:homogeneity_all}). Among the identified relations, top-ranking examples align with commonly held knowledge. Overall, we find that tagging provides robust statistics that offer useful insights into higher-level relations within the data. This answers \textbf{Q3} affirmatively.

\textbf{Transportability of Extracted Causal Relations.} While the found relations can give insights into their respective datasets, the question arises whether these relations can be transported across datasets to possibly build up broader domain knowledge or inform further causal discoveries. To show the possible generality of the identified abstract tag relations, we analyzed relations across all 11 datasets that were (pairwise) shared between any two of them. Despite spanning different domains and featuring diverse tags due to the open annotation scheme, scanning all of them yielded a small set of six shared tag relations (with sufficient support of 3+ edges under exact tag name matches). Setup and results are presented in App.~\ref{app:shared-tag-relations} and Tab.~\ref{tab:shared-tag-relations}.

Of the six relations found, five agreed on their direction across datasets, with the only exception being the ``respiratory -- disease'' relation, shared between the 'Cancer' and 'Lucas' datasets, which points in different directions depending on the dataset (Cancer: respiratory $\to$ disease; Lucas: disease$\to$respiratory). This also illustrates the primary obstacle to full generalization: depending on whether a domain models the direction of a disease-causing respiration, or the inference direction of (the lack of) respiration indicating a disease to a medical doctor, both directions may be valid within their respective dataset.

This conflict alludes to the general discussion about causal and anti-causal inference \citep{scholkopf2012causal}. For example, do the causal relations model the physical 'causal' process, or do they model decision processes that infer information in the anti-causal direction in order to act proactively? Deciding on the causal-versus-anti-causal question when transferring knowledge between datasets (e.g., whether direction needs to be flipped when switching from causal to anti-causal modeling) is not possible based solely on tags, but requires further information that goes beyond the considered tag information as discussed in this paper.

\section{Conclusion}
\label{sec:conclusion}
While both data and semantic knowledge are valuable sources of information for identifying causal relations, hybrid approaches that integrate both sources of information appear to be a valuable avenue for further refining causal graphs.
Our tagging approach advances this direction by utilizing semantic tags as light-weight meta-information, thereby allowing existing data to be leveraged more effectively.
We showed that the employed probabilistic approach for further refining edge directions is particularly robust under tagging noise and graph faults, achieving noticeable improvements across all experiments.

One observed limitation of our approach might be the possible trade-off between directing as many edges as possible and abstaining from a decision, particularly in smaller graphs with insufficient grounding.
Extending our method to account for confounders or unknown latent variables is an important future direction.

\vfill

\subsubsection*{Acknowledgments}
This work is supported by the Hessian Ministry of Higher Education, Research, Science and the Arts (HMWK).
The authors acknowledge the support of the German Science Foundation (DFG) research grant ``Tractable Neuro-Causal Models'' (KE 1686/8-1).
The work also benefited from the Cluster of Excellence "Reasonable AI" funded by the German Research Foundation (DFG) under Germany’s Excellence Strategy—EXC-3057.
The Eindhoven University of Technology authors received support from their Department of Mathematics and Computer Science and the Eindhoven Artificial Intelligence Systems Institute.

\bibliography{./refs}
\bibliographystyle{./tmlr}

\newpage

\appendix
\section*{Appendix ``Tags for DAGs: Graph Refinement with Meta-Informed Relations''}

This appendix is structured as follows.
First, we go into a theoretical analysis of our tagging approach in App.~\ref{app:theory}. Then, we describe a possible extension of the approach with a specificity prior in App.~\ref{appx:specificityPrior}. Next, we present the pseudo-code for our approach in App.~\ref{sec:pseudo_code}.
App.~\ref{sec:llmPromptsAppx} provides detailed information on the LLM prompts used to assign types and tags in this paper.
Finally, we present extensive experimental details and results in App.~\ref{sec:all_results}.

\section{Theoretical Analysis}
\label{app:theory}
Let us say that two variables $X_i$ and $X_j$ are only tagged with a single tag each, i.e., $(X_i,X_j) \in \Edges^{\sampleSpace,\tg^a,\tg^b}$ with $\tagFunc(X_i)=\mathbf{t}_i = \{\tg^a\}$ and $\tagFunc(X_j)=\mathbf{t}_j =\{\tg^b\}$.
Consider the case that $X_i \rightarrow X_j$ is true, but unknown.
The tag informative value $I^{\sampleSpace}(\tg^a, \tg^b)$ describes the probability provided by the tag pair $(\tg^a, \tg^b)$ according to which $\tg^a$ is a cause of $\tg^b$.
Thus, correctly predicting the direction of an undirected edge follows a simple Bernoulli distribution and is correct with probability $d'^{\sampleSpace}_{ij} = I^{\sampleSpace}(\tg^a, \tg^b)$ (see Eq.~\ref{eq:direction}).
For ease of notation, we assume that all considered tag pairs have evidence ($o^{\sampleSpace}(t^a, t^b) = 1,\ \forall\, t^a, t^b$) and that the edge-specific noise $N_{ij}$ does not influence the prediction.

For assigning tags, we assume a noisy oracle that assigns tags that result in a particular accuracy of $\hat{A} := \mathbb{E}(I^{\sampleSpace}(\tg^a, \tg^b))$. Generally, we assume that the tag informative values follow a Beta distribution, the conjugate prior of the Bernoulli distribution:
\begin{equation}
    I^{\sampleSpace}(\tg^a, \tg^b) \sim \text{Beta}(\alpha, \beta) \text{~with~} \hat{A} = \frac{\alpha}{\alpha + \beta}
\end{equation} for some $\alpha, \beta \in \mathbb{R}$.
Here, $\alpha$ and $\beta$ can be seen as some underlying evidence for directing the edge $X_i - X_j$ in either direction.
Following Assumption~\ref{assm:tagDistrConsist}, according to which the tag informative values of different datasets converge towards each other (i.e., tag relations observed in some data help with correctly predicting the edges of new data), we can assume that, on average, $\alpha > \beta$ if $X_i \rightarrow X_j$ and vice versa.

Under this setting, we are now interested in the probability of correctly predicting the edge direction using our tagging approach $\mathbb{P}[d'^{\sampleSpace}_{ij} > 0.5]$.
In the case $n= |\mathbf{t}_i \times \mathbf{t}_j| = 1$, we directly use the only tag informative value for our prediction:
\begin{equation}
    \mathbb{P}[\hat{p}_1(X_i \rightarrow X_j) > 0.5] = \mathbb{P}[I^{\sampleSpace}(\tg^a, \tg^b) > 0.5] = 1 - \mathcal{I}_{0.5}(\alpha, \beta),
\end{equation}
where $\mathcal{I}_{0.5}(\alpha, \beta) = \frac{\int_0^{0.5} u^{\alpha-1}(1 - u)^{\beta-1} \, du}{\int_0^1 u^{\alpha-1}(1 - u)^{\beta-1} \, du}
$ is the regularized incomplete beta function\footnote{We use $\mathcal{I}$ instead of the commonly used $I$ to separate it from the tag informative value $I$.}, i.e., the cumulative distribution function (CDF) of the beta distribution.
Intuitively, for a single tag pair, the probability of correctly predicting the edge $X_i \rightarrow X_j$ when $\alpha > \beta$ is the same as the probability of sampling a probability (tag informative value) higher than $0.5$ from the underlying beta distribution.

The previous case is similar to the typing scenario, since we only considered a single tag pair influencing the decision.
Let us now consider the more general scenario with multiple tags: $\mathbf{t}_i =\{\tg^a_1, \tg^a_2, \dots, \tg^a_k\}$ and $\mathbf{t}_j=\{\tg^b_1, \tg^b_2, \dots, \tg^b_l\}$, resulting in $n = |\mathbf{t}_i \times \mathbf{t}_j|$ tag pairs and tag informative values $\mathbf{I}^{\sampleSpace} = \{I^{\sampleSpace}_1, I^{\sampleSpace}_2, \dots, I^{\sampleSpace}_n\}$.
To allow for a simple formal analysis, we assume these $n$ tag informative values to be independent and identically distributed (i.i.d.).
Our tagging approach makes the following prediction:
\begin{equation}
    d'^{\sampleSpace}_{ij} = \frac{1}{n} \sum_{m=1}^n I^{\sampleSpace}_m.
\end{equation}
We can see that the mean is the same as the mean of the underlying beta distribution
\begin{equation}
    \mathbb{E}[d'^{\sampleSpace}_{ij}]= \mathbb{E}[\frac{1}{n} \sum_{m=1}^n \text{Beta}(\alpha, \beta)] = \frac{1}{n} \sum_{m=1}^n \mathbb{E}[\text{Beta}(\alpha, \beta)] = \frac{1}{n} \sum_{m=1}^n \frac{\alpha}{\alpha + \beta} = \frac{\alpha}{\alpha + \beta},
\end{equation}
which then tells us that we are making the correct predictions in the mean as $\mathbb{P}[\frac{\alpha}{\alpha + \beta} > 0.5]$ is true following our assumption that $\alpha > \beta$.
More generally, we know that the probability of correctly predicting an edge using tagging is
\begin{equation}
    \mathbb{P}[d'^{\sampleSpace}_{ij} > 0.5] = \mathbb{P}[\frac{1}{n} \sum_{m=1}^n I^{\sampleSpace}_m > 0.5] = 1 - F^n_{0.5}(\alpha, \beta),
\end{equation}
where $F^n_{0.5}(\alpha, \beta)$ is the CDF of the distribution of the average of beta-distributed random variables ($I^{\sampleSpace}$) to which there exists no closed-form expression.
However, given the variance of the beta distribution $\text{Var}[I^{\sampleSpace}] = \frac{\alpha \beta}{(\alpha + \beta)^2 (\alpha + \beta + 1)}$, we can make use of the Central Limit Theorem (CLT), according to which our prediction is approximately normally distributed with variance
\begin{equation}
    \text{Var}[d'^{\sampleSpace}_{ij}] = \text{Var}[\frac{1}{n} \sum_{m=1}^n I^{\sampleSpace}_m] = \frac{\alpha \beta}{(\alpha + \beta)^2 (\alpha + \beta + 1) n}.
\end{equation}
Since $\text{Var}[d'^{\sampleSpace}_{ij}]$ becomes smaller for larger $n$, we know that the CDF $F^n(\alpha, \beta)$ will continually amass more density around the mean (as it is normally distributed following the CLT).
Therefore, $F^n_{0.5}(\alpha, \beta)$ will get smaller, the more $n$ increases, since the mean is not in the interval $[0, 0.5]$.
In other words, the more tag pairs we utilize for our prediction, the more likely it is that our prediction is correct, exceeding the probability of only using a single tag pair or type.
In the theoretical setting of $n \rightarrow \infty$, it even follows that $\text{Var}[d'^{\sampleSpace}_{ij}]_{n \rightarrow \infty} = 0$ and, therefore, $\mathbb{P}[d'^{\sampleSpace}_{ij} > 0.5] = 1$.

The opposite case of $X_i \leftarrow X_j$ follows by analogy.

\textbf{Limitations.} Our proof requires a strict i.i.d. assumption on the tag distribution, which might be violated in practice, but could be relaxed by adopting a more general notion of non-systematicity in the proof. Empirical evaluation indicates that the most severe errors usually stem from limited data artifacts (incorrect discovery of the CPDAG, insufficient tagging evidence due to a limited graph size, \dots), rather than obvious systematic tagging violations. We therefore expect the proof to `approximately' hold in practice under mild conditions. Further generalization might require a more careful reframing of the assumed (non-)systematicity of tagging and is left for future work.

\section{Specificity Prior}
\label{appx:specificityPrior}

Upon considering the directionality of edges, one might want to weigh in tagging pairs $(\tg^a,\tg^b)$ more strongly, which feature a higher \textit{specificity}. In general, specific tags might be used to indicate variables with deviating behavior, e.g., variables featuring different causal dynamics in contrast to the overall system. Such specific tags might indicate local environments where global assumptions might not hold, and upon encountering such tags, we want to prioritize them more strongly in our decision. We base the prior distribution on the Geometric series, counting the number of nodes assigned to any of the tags:
\begin{equation}
\label{eq:specificity}
    \hat{p}^{\text{spec}}_{ab}  = \frac{1}{2^{|\{ X_i \in \CVars | \tg^a \in \taggedFunc(X_i) \}| + \{ X_i \in \CVars | \tg^b \in \taggedFunc(X_i) \}| - 1}}.
\end{equation}
Generally, the prior might or might not be used based on whether the specificity assumption is assumed to hold for the specific application. Thus, it is a hyperparameter for our algorithm. In case the prior is used, it is multiplied to the respective probability; otherwise, $\hat{p}^{\text{spec}}_{ab}$ is set to one:
\begin{equation}
    \hat{p}'_{ab} = p^{\text{spec}}_{ab} \cdot \hat{p}_{ab}.
\end{equation}

Note that Eq.~\ref{eq:edgeQ} now requires a normalization term, since $\hat{p}'_{ab} + \hat{p}'_{ba}$ might no longer add up to one. We repeat the corresponding equation with normalization in the following:
\begin{equation}
\label{eq:edgeQNorm}
    Q(\Edge_{ij}) = %
    \frac{\sum\nolimits_{\tg^a\in \tagset_i, \tg^b\in \tagset_j} \hat{p}'_{ab} \cdot o(t^a,t^b)}{\sum\nolimits_{\tg^a\in \tagset_i, \tg^b\in \tagset_j} \hat{p}'_{ab} + \hat{p}'_{ba}}.
\end{equation}

\section{Algorithm Pseudo Code}
\label{sec:pseudo_code}
We introduce some methods required in our approach in Alg.~\ref{alg:tagging_helper} and then provide the pseudo-code for the actual tagging informed edge direction in Alg.~\ref{alg:tagging}.

\begin{algorithm}
\caption{Tagging Helper Procedures}
\begin{algorithmic}[1]
    \Procedure{Collect Evidence}{$\mathcal{G}$, $\mathbf{t}$}
        \State $\mathbf{E} \gets$ zero matrix of size $|\mathbf{t}| \times |\mathbf{t}|$
        \For{each directed edge $(u, v) \in \mathcal{G}$}
            \For{each tag $t_u \in \mathbf{t}[u]$}
                \For{each tag $t_v \in \mathbf{t}[v]$}
                    \State $\mathbf{E}[t_u, t_v] \gets \mathbf{E}[t_u, t_v] + 1$
                \EndFor
            \EndFor
        \EndFor
        \State \Return $\mathbf{E}$
    \EndProcedure

    \Procedure{Find Most Promising Edge}{$\text{Edges}$, $\mathbf{t}$, $\mathbf{E}$, $\text{OnlyBackward}$}
        \State $\text{BestEdge} \gets \text{None}$
        \State $\text{BestProb} \gets 0$
        \For{each edge $(u, v) \in \text{Edges}$}
            \State $\text{ForwardScore}, \text{BackwardScore}, \text{ForwardCount}, \text{BackwardCount} \gets 0$
            \For{each tag $t_u \in \mathbf{t}[u]$}
                \For{each tag $t_v \in \mathbf{t}[v]$}
                    \If{$t_u \neq t_v$} \Comment{Skip relations between identical tags}
                        \State $\text{Forward} \gets \mathbf{E}[t_u, t_v]$
                        \If{$\text{not}(\text{IncludeCurrentEdgeAsEvidence})$} \Comment{Parameter for redirecting edges}
                            \State $\text{Forward} \gets \text{Forward} - 1$ \Comment{Exclude the current edge}
                        \EndIf
                        \State $\text{Backward} \gets \mathbf{E}[t_v, t_u]$
                        \If{$\text{SpecificityPrior}$} \Comment{Parameter for applying the specificity prior}
                            \State $p^\text{spec} \gets \Call{\text{CalcSpecificityPrior}}{t_u, t_v}$ \Comment{Eq.~\ref{eq:specificity}}
                        \Else
                            \State $p^\text{spec} \gets 1$
                        \EndIf
                        \State $\text{ForwardScore} \gets \text{Forward} \cdot p^\text{spec}$
                        \State $\text{BackwardScore} \gets \text{Backward} \cdot 
                        p^\text{spec}$
                        \State $\text{ForwardCount} \gets \text{ForwardCount} + \text{ForwardScore}$
                        \State $\text{BackwardCount} \gets \text{BackwardCount} + \text{BackwardScore}$
                    \EndIf
                \EndFor
            \EndFor
            \If{$\text{ForwardCount} + \text{BackwardCount} < \text{MinSamples}$} \Comment{Minimum\,tag\,pairs,\,we\,use\,either\,1\,or\,2}
                \State $\text{ForwardProb} \gets 0.5$
            \Else
                \State $\text{ForwardProb} \gets \frac{\text{ForwardScore}}{\text{ForwardScore}+\text{BackwardScore}}$
            \EndIf
            \State $\text{BackwardProb} \gets 1 - \text{ForwardProb}$
            \If{$\text{not}(\text{BackwardOnly})~\textbf{and}~ \text{ForwardProb} > \text{BestProb}$} \Comment{Ignore forward direction for redirections}
                \State $\text{BestEdge} \gets (u, v)$
                \State $\text{BestProb} \gets \text{ForwardProb}$ 
            \EndIf
            \If{$\text{BackwardProb} > \text{BestProb}$}
                \State $\text{BestEdge} \gets (v, u)$
                \State $\text{BestProb} \gets \text{BackwardProb}$ 
            \EndIf
        \EndFor
        \State \Return $\text{BestEdge}$, $\text{BestProb}$
    \EndProcedure
\end{algorithmic}
\label{alg:tagging_helper}
\end{algorithm}

\begin{algorithm}
\caption{Tagging Informed Edge Direction}
\begin{algorithmic}[1]

    \Procedure{Tagging Informed Edge Directioning}{CPDAG $\mathcal{G}$, Tags $\mathbf{t}$}
    \State $\mathbf{E} \gets$ \Call{Collect Evidence}{$\mathcal{G}$, $\mathbf{t}$}
    \If{$\text{RedirectEdges}$} \Comment{Parameter: consider redirectioning of edges?}
        \State $\text{Loop} \gets \text{True}$
        \State $\text{DirEdges} \gets \text{directed\_edges}(\mathcal{G})$
        \State $\text{RedirEdges} \gets []$
        \State $\text{IterationThreshold} \gets 0$
        \While{$\text{Loop}$ \textbf{and} $\text{IterationThreshold} < 100$} \Comment{Stop after 100 iterations to avoid infinite loops}
            \State $\text{Loop} \gets \text{False}$
            \State $(u, v), p \gets$ \Call{Find Most Promising Edge}{$\text{DirEdges}$, $\mathbf{t}$, $\mathbf{E}$, $\text{True}$}
            \If{$p > 0.6$}
                \State $\mathcal{G}^* \gets \mathcal{G}.\text{direct}((u, v))$ \Comment{Redirect edge}
                \If{$\text{is\_acyclic}(\mathcal{G}^*)$}
                    \State $\mathcal{G} \gets \text{copy}(\mathcal{G}^*)$ \Comment{Accept redirection}
                    \State $\text{DirEdges} \gets \text{directed\_edges}(\mathcal{G})$
                    \If{$\text{RedirectStrategy} == 0$} \Comment{Update evidence after each step}
                        \State $\mathbf{E} \gets$ \Call{Collect Evidence}{$\mathcal{G}$, $\mathbf{t}$}
                    \Else \Comment{$\text{RedirectStrategy} == 1$}
                        \State $\text{RedirEdges} \gets \text{RedirEdges}.\text{append}((u, v))$
                        \State $\text{DirEdges} \gets \text{DirEdges}.\text{remove\_all}(\text{RedirEdges})$ \Comment{Remove already redirected edges}
                    \EndIf
                \Else \Comment{Do not redirect}
                    \State $\text{DirEdges} \gets \text{DirEdges}.\text{remove}((u, v))$ \Comment{Do not consider same edge again next}
                \EndIf
                \If{$\text{RedirectStrategy} == 0$}
                    \State $\text{IterationThreshold} \gets \text{IterationThreshold} + 1$
                \EndIf
                \State $\text{Loop} \gets \text{True}$
            \EndIf
        \EndWhile
        \State $\mathbf{E} \gets$ \Call{Collect Evidence}{$\mathcal{G}$, $\mathbf{t}$}
    \EndIf
    \While{$\text{len}(\text{undirected\_edges}(\mathcal{G})) > 0$}  \Comment{As long as there are undirected edges}
        \State $(u, v), p \gets$ \Call{Find Most Promising Edge}{$\text{undirected\_edges}(\mathcal{G})$, $\mathbf{t}$, $\mathbf{E}$, $\text{False}$}
        \If{$p > 0.5$} \Comment{Stricter thresholds are also possible (hyperparameter)}
            \State $\mathcal{G}^* \gets \mathcal{G}.\text{direct}((u, v))$
                \If{$\mathcal{G}^*.\text{is\_acyclic}()$} \Comment{If the direction would introduce cycles...}
                    \State $\mathcal{G} \gets \mathcal{G}.\text{direct}((v, u))$ \Comment{...instead add the edge in the opposite direction}
                \Else
                    \State $\mathcal{G} \gets \mathcal{G}^*$ \Comment{Otherwise, accept the directed edge}
                \EndIf
                \If{$\text{AlwaysMeek}$} \Comment{Parameter, to apply Meek Rules after every edge direction step}
                    \State $\mathcal{G} \gets \Call{Meek}{\mathcal{G}}$ \Comment{Meek Rules}
                \EndIf
        \Else \Comment{If $p=0.5$ (can not be smaller), keep remaining edges undirected}
            \State $\textbf{break}$
        \EndIf
    \EndWhile
    \State $\mathcal{G} \gets \Call{Meek}{\mathcal{G}}$ \Comment{Always apply one final Meek Rules}
    \State \Return $\mathcal{G}$
    \EndProcedure
\end{algorithmic}
\label{alg:tagging}
\end{algorithm}

\newpage

\section{LLM Prompts}
\label{sec:llmPromptsAppx}

In the following, we describe the details of LLM querying.
All answers are generated using the same system prompt and either the following tagging or typing instructions. Model temperature was set to 1.0.
All LLM answers are included within the code repository. Within every prompt, the \texttt{\{variables\}} fields were replaced by a comma-separated list of the individual dataset's variable names.

\vspace{5mm}

\begin{promptbox}[System Prompt]
You are an expert in annotating variables to provide additional information that helps to support a causal discovery algorithm.
\end{promptbox}
\begin{promptbox}[Tagging Instruction Prompt]
A tag is a single word or short phrase that describes a variable. Tags should be general enough to be applicable to multiple variables but specific enough to identify differences between similar variables. Tags will be used to identify causal directions between variables. Therefore, the individual sets of tags per variable should be discriminative enough to inform the algorithm. Variables can have multiple tags.\texttt{{\textbackslash}n}

Consider the following variables: \texttt{\{variables\}}.\texttt{{\textbackslash}n}\texttt{{\textbackslash}n}\\

Please generate a list of tags that can be assigned to one or multiple variables. Generate the number of tags necessary to strike a good balance between expressivity and specificity. Avoid duplicate tags that contain the same set of variables. Reply with one line per tag, where each line starts with the name of the tag, followed by a colon, and then a comma-separated list of variables that have that tag. The output should be machine parsable. For that reason, do not include any explanations or additional comments.
\end{promptbox}

\begin{promptbox}[Typing Instruction Prompt]
A type is a single word or short phrase that describes a variable. Types should be general enough to be applicable to multiple variables but specific enough to identify differences between similar variables. Types will be used to identify causal directions between variables. Therefore, the individual types should be discriminative enough to inform the algorithm. Variables are assigned to a single type only.\texttt{{\textbackslash}n}

Consider the following variables: \texttt{\{variables\}}.\texttt{{\textbackslash}n}\texttt{{\textbackslash}n}\\
        
Please generate a list of types that can be assigned to one or multiple variables. Generate the number of types necessary to strike a good balance between expressivity and specificity. Reply with one line per type, where each line starts with the name of the type, followed by a colon, and then a comma-separated list of variables that belong to that type. Make sure that no variable appears in more than one the lists. The output should be machine parsable. For that reason, do not include any explanations or additional comments.\\
\end{promptbox}

\newpage

\subsection{Prompt Variations}
\label{sec:prompt_variations}

The following subsection contains prompts for the different prompt variation experiments that test robustness of the tagging approach under varying LLM instructions. Results are discussed in Sec.~\ref{sec:appxPromptAblation} (Tabs.~\ref{tab:llm_versions_pc} and~\ref{tab:llm_versions_ges}).

\newcommand\myxrule{\bgroup \markoverwith{\lower3.5pt\hbox{\tiny x}}\ULon}
\newcommand\myorule{\bgroup \markoverwith{\lower3pt\hbox{\tiny '}}\ULon}
\definecolor{myblue1}{HTML}{7ee8ec}
\definecolor{myblue2}{HTML}{4cc6da}
\definecolor{myblue3}{HTML}{02a4c8}
\definecolor{myblue4}{HTML}{0081b6}
\definecolor{myblue5}{HTML}{005ea0}
\definecolor{myblue6}{HTML}{00388c}
\definecolor{myblue7}{HTML}{020873}
\textbf{Prompt Decomposition.} This setup decomposes and individually holds out different parts of the standard prompt. The parts in the following prompt are marked by individual colors and underscore styles:
\begin{multicols}{3}
\begin{itemize}
    \item {\color{myblue1}\uline{tag definition}}
    \item {\color{myblue2}\dashuline{specificity tradeoff}}
    \item {\color{myblue3}\dotuline{downstream purpose}}
    \item {\color{myblue4}\uwave{tag discriminativeness}}
    \item {\color{myblue5}\myorule{multiple tags}}
    \item {\color{myblue6}\myxrule{tag balance}}
    \item {\color{myblue7}\uuline{no duplicates}}
\end{itemize}
\end{multicols}
Unmarked text provides required instructions, variable inputs or describes the output format and are always included.

\begin{promptbox}[Prompt Decomposition]
{\color{myblue1}\uline{A tag is a single word or short phrase that describes a variable.}} {\color{myblue2}\dashuline{Tags should be general enough to be applicable to multiple variables but specific enough to identify differences between similar variables.}} {\color{myblue3}\dotuline{Tags will be used to identify causal directions between variables.}} {\color{myblue4}\uwave{Therefore, the individual sets of tags per variable should be discriminative enough to inform the algorithm.}} {\color{myblue5}\myorule{Variables can have multiple tags.}}\texttt{{\textbackslash}n}

Consider the following variables: \texttt{\{variables\}}.\texttt{{\textbackslash}n}\texttt{{\textbackslash}n}\\

Please generate a list of tags that can be assigned to one or multiple variables. {\color{myblue6}\myxrule{Generate the number of tags necessary to strike a good balance between expressivity and specificity.}} {\color{myblue7}\uuline{Avoid duplicate tags that contain the same set of variables.}} Reply with one line per tag, where each line starts with the name of the tag, followed by a colon, and then a comma-separated list of variables that have that tag. The output should be machine parsable. For that reason, do not include any explanations or additional comments.
\end{promptbox}

\textbf{Paraphrasing.} The following prompts rewrite instructions of the original prompt as a bullet point list (see `Paraphrasing - Bullets') of varying verbosity: `Paraphrasing - Terse' additionally shortens the prompt to the strictly relevant instructions, while `Paraphrasing - Verbose' lengthens the sentences and introduces repetition of statements.

\begin{promptbox}[Paraphrasing - Bullets]
\begin{itemize}[leftmargin=*]
    \item[-] A tag is a single word or short phrase that describes a variable.\vspace{-2mm}
    \item[-] Tags should be general enough to be applicable to multiple variables, but specific enough to identify differences between similar variables.\vspace{-2mm}
    \item[-] Tags will be used to identify causal directions between variables. The individual sets of tags per variable should therefore be discriminative enough to inform the algorithm.\vspace{-2mm}
    \item[-] Variables can have multiple tags.\vspace{-2mm}
    \item[-] The variables are: \texttt{\{variables\}}.\vspace{-2mm}
    \item[-] Generate a list of tags that can be assigned to one or multiple variables.\vspace{-2mm}
    \item[-] Generate the number of tags necessary to strike a good balance between expressivity and specificity.\vspace{-2mm}
    \item[-] Avoid duplicate tags that contain the same set of variables.\vspace{-2mm}
    \item[-] Reply with one line per tag: the name of the tag, a colon, then a comma-separated list of variables that have that tag.\vspace{-2mm}
    \item[-] The output should be machine parsable. Do not include any explanations or additional comments.\vspace{2mm}
\end{itemize}
\end{promptbox}

\newpage
\textit{Paraphrasing prompts continued \dots}
\begin{promptbox}[Paraphrasing - Terse]
\begin{itemize}[leftmargin=*]
    \item[-] Tag = word or short phrase describing a variable.\vspace{-2mm}
    \item[-] General enough for several variables, specific enough to separate similar ones.\vspace{-2mm}
    \item[-] Purpose: identifying causal directions. Tag sets per variable must be discriminative.\vspace{-2mm}
    \item[-] Multiple tags per variable allowed.\vspace{-2mm}
    \item[-] Variables: \texttt{\{variables\}}.\vspace{-2mm}
    \item[-] Balance expressivity and specificity. No two tags with the same variable set.\vspace{-2mm}
    \item[-] Format: one line per tag, `tag: var1, var2, ...`. Machine parsable. No other text.\vspace{2mm}
\end{itemize}
\end{promptbox}

\begin{promptbox}[Paraphrasing - Verbose]
\begin{itemize}[leftmargin=*]
    \item[-] A tag is a single word or a short phrase that describes a variable. That is, each tag is a brief textual label, consisting of one word or at most a few words, which characterizes some property of the variable it is assigned to.\vspace{-2mm}
    \item[-] Tags should be general enough to be applicable to multiple variables. At the same time, tags should be specific enough to identify differences between similar variables. In other words, a tag should neither be so broad that it applies indiscriminately to every variable, nor so narrow that it applies to only one variable and captures nothing shared.\vspace{-2mm}
    \item[-] Tags will be used to identify causal directions between variables. Because the tags serve this downstream purpose, the individual set of tags assigned to each variable should be discriminative enough to inform the algorithm. That is, two different variables should, wherever possible, receive different sets of tags, so that the algorithm can tell them apart.\vspace{-2mm}
    \item[-] Variables can have multiple tags. There is no requirement that a variable be assigned only a single tag, and there is no requirement that the tags form a partition of the variables.\vspace{-2mm}
    \item[-] Consider the following variables, which are the variables to be annotated: \texttt{\{variables\}}.\vspace{-2mm}
    \item[-] Please generate a list of tags that can be assigned to one or multiple variables. Each tag in the list should be accompanied by the variables to which it applies.\vspace{-2mm}
    \item[-] Generate the number of tags necessary to strike a good balance between expressivity and specificity. Do not generate so few tags that the annotation becomes uninformative, and do not generate so many tags that the annotation becomes redundant.\vspace{-2mm}
    \item[-] Avoid duplicate tags that contain the same set of variables. If two candidate tags would apply to exactly the same set of variables, they are redundant with one another, and only one of them should be kept.\vspace{-2mm}
    \item[-] Reply with one line per tag, where each line starts with the name of the tag, followed by a colon, and then a comma-separated list of variables that have that tag. Each line therefore has the form `tag: variable, variable, ...`.\vspace{-2mm}
    \item[-] The output should be machine parsable. For that reason, do not include any explanations or additional comments, and do not include any introductory or concluding text, headers, numbering, or formatting beyond the specified line format.
\end{itemize}
\end{promptbox}

\newpage
\textbf{Politeness.} The following prompt alter the level of politeness of the given instructions. The prompt `Politeness - Polite' is thought to be more polite than the original phrasing, while `Politeness - Demanding' and `Politeness - Imperative' are phrased increasingly more demanding.

\begin{promptbox}[Politeness - Polite]
A tag is a single word or short phrase that describes a variable. Ideally, tags would be general enough to be applicable to multiple variables but specific enough to identify differences between similar variables. The tags will be used to identify causal directions between variables, so it would be very helpful if the individual sets of tags per variable were discriminative enough to inform the algorithm. Variables may of course have multiple tags.\texttt{{\textbackslash}n}\\
Consider the following variables: \texttt{\{variables\}}.\texttt{{\textbackslash}n}\texttt{{\textbackslash}n}\\

If you would be so kind, could you please generate a list of tags that can be assigned to one or multiple variables? Please feel free to generate whatever number of tags you consider necessary to strike a good balance between expressivity and specificity. It would be much appreciated if duplicate tags containing the same set of variables could be avoided. When you are ready, please reply with one line per tag, where each line starts with the name of the tag, followed by a colon, and then a comma-separated list of variables that have that tag. Since the output needs to be machine parsable, we would kindly ask you to omit any explanations or additional comments. Thank you very much for your help.
\end{promptbox}

\begin{promptbox}[Politeness - Demanding]
A tag is a single word or short phrase that describes a variable. Tags must be general enough to be applicable to multiple variables but specific enough to identify differences between similar variables. Tags will be used to identify causal directions between variables. The individual sets of tags per variable must therefore be discriminative enough to inform the algorithm. Variables can have multiple tags.\texttt{{\textbackslash}n}\\
Consider the following variables: \texttt{\{variables\}}.\texttt{{\textbackslash}n}\texttt{{\textbackslash}n}\\

Generate a list of tags that can be assigned to one or multiple variables. You need to generate the number of tags required to strike a good balance between expressivity and specificity. Duplicate tags that contain the same set of variables are not acceptable. Your reply must contain one line per tag, where each line starts with the name of the tag, followed by a colon, and then a comma-separated list of variables that have that tag. The output has to be machine parsable, so it must not contain explanations or additional comments.
\end{promptbox}

\begin{promptbox}[Politeness - Imperative]
A tag is a single word or short phrase describing a variable. Make tags general enough to apply to multiple variables and specific enough to separate similar variables. Tags will be used to identify causal directions between variables: make the set of tags per variable discriminative. Assign multiple tags per variable where appropriate.\texttt{{\textbackslash}n}\\
Variables: \texttt{\{variables\}}.\texttt{{\textbackslash}n}\texttt{{\textbackslash}n}\\

Generate the tag list now. Choose the tag count that balances expressivity and specificity. Never emit two tags covering the same set of variables. Output one line per tag: tag name, colon, comma-separated variables. Output nothing else. No explanations. No comments. No preamble. Machine-parsable output only.
\end{promptbox}

\newpage
\textbf{Number of Tags.} In order to analyze effects of differing tag counts on the tagging approach, the following prompt requires the LLM to assign a fixed number of tags ($\texttt{numTags} \in \{2, 3, 4, 6, 8, 10\}$) to each variable, instead of letting the LLM determine a reasonable number of LLM individually.

\begin{promptbox}[Number of Tags]
A tag is a single word or short phrase that describes a variable. Tags should be general enough to be applicable to multiple variables but specific enough to identify differences between similar variables. Tags will be used to identify causal directions between variables. Therefore, the individual sets of tags per variable should be discriminative enough to inform the algorithm. Every variables MUST be assigned exactly \texttt{\{numTags\}} tags.\texttt{{\textbackslash}n}\\
Consider the following variables: \texttt{\{variables\}}.\texttt{{\textbackslash}n}\texttt{{\textbackslash}n}\\

Please generate a list of tags that can be assigned to one or multiple variables. Generate the number of tags necessary to strike a good balance between expressivity and specificity. Avoid duplicate tags that contain the same set of variables. Reply with one line per tag, where each line starts with the name of the tag, followed by a colon, and then a comma-separated list of variables that have that tag. The output should be machine parsable. For that reason, do not include any explanations or additional comments.
\end{promptbox}

\newpage

\section{Further Experimental Results}
\label{sec:all_results}
In this section, we provide additional details on the experiments and further experimental results.
Starting off, we take a look at how well the tag consistency assumption holds empirically (Sec.~\ref{app:tagConsistencyAnalysis}).
In the following subsections, we take a closer look at our main evaluation from Tab.~\ref{tab:avgBestRankedResults}.
First, we describe two additional parameters of the actual algorithm's implementation (Sec.~\ref{sec:implementationDetails}) regarding the preprocessing of tagging sets. Second, we give more information on the best parameter configurations of our main evaluation and the difference between these with respect to different LLMs (Sec.~\ref{sec:best_configs}), and we conduct a leave-one-out cross-validation to check whether our analysis overfits to a specific parameter configuration (Sec.~\ref{sec:loocv}).
Then, we show the results on all datasets, supplementing the purely ranking-based results of the main paper (Sec.~\ref{sec:all_evals}).
We consider an optimal causal discovery method by investigating how well our tagging-based method performs on the ground-truth CPDAG (Sec.~\ref{sec:eval_cpdag}).
Two further experiments focus on the amount of evidence necessary for making a decision, one varying the threshold for making predictions in the first place (Sec.~\ref{sec:vary_thresholds}), the other repeating our main evaluation using Laplace correction to account for epistemic uncertainty (Sec.~\ref{sec:laplace_correction}).
The last section to considers different variations to the main evaluation investigates tags obtained using different LLM versions or prompts (Sec.~\ref{sec:appxPromptAblation}).

After that, we move on to experiments that employ different experimental setups from our main evluation.
First, we undirect some edges in an ablation study, starting from the ground-truth graph (Sec.~\ref{sec:undirected}).
Investigating the robustness of our approach, we first consider settings where edges have been removed or inverted (Sec.~\ref{sec:all_faults}), followed by different experiments where we add noise to the tag assignment and investigate the effect on the tagging accuracy (Sec.~\ref{app:tag_noise}).
Lastly, we complement our evaluations of the main paper investigation into homogeneity and tag relations by providing a larger amount of experimental results in  Sec.~\ref{sec:homogeneity_all} and studying the overlap of discovered tag relations (Sec.~\ref{app:shared-tag-relations}).

\subsection{Tag Consistency}
\label{app:tagConsistencyAnalysis}
Our approach hinges on the tag distribution consistency assumption, i.e., tag information gathered from directed edges helps inform the direction of undirected edges.
To check this assumption empirically, we run an experiment on the same datasets used for the other experiments and consider the true CPDAG and the true DAG.
Precisely, we gather tag evidence on the CPDAG to determine the tag informative value for all tag pairs (i.e., 0 or 1 if the evidence is uniquely in favor of one direction, and something in between otherwise).
We then do the same for the edges in the true DAG that were undirected in the CPDAG.
Based on this data, we compare all tag pairs that have some evidence in both and determine the Pearson correlation and the confidence interval using the Fisher transformation.
In this setup, the overall correlation would be 1 if the tag informative values are always identical, -1 if the tag informative values would flip across scenarios, 0 if both groups are uncorrelated, or something in between.

Results are shown in Tab~\ref{tab:pearson_tags}.
Across LLMs, we see positive point-estimate correlations between tag informative values collected as evidence for directing and those to which the learned evidence is applied.
Further, 4 out of 7 LLMs place the correlation coefficient $r$ in the positive regime with 95\% confidence.
Out of the other 3 LLMs, GPT 5.2 and Qwen 3.5 contain a relatively small number of tag pairs, hence the confidence intervals are larger, and GLM shows the lowest correlation coefficient at $r=0.247$.
Interestingly, our best-performing model in the main experiments (Claude 4.6) does not have the strongest correlation, which is given for Llama 3.3.
However, Claude 4.6 has many more tag pairs available as evidence, which indicates that the amount of evidence might matter slightly more than the tag-pair ``quality''.
This, as well as the fact that all correlations are clearly far from perfect (which would be $r = 1$), provides further empirical evidence for our decision to move beyond types to tags, which allows the algorithm to ``overrule'' bad tags in the evidence aggregation process.
Hence, these results also fit our theoretical analysis (App.~\ref{app:theory}).

\begin{table}[]
    \centering
\resizebox{\textwidth}{!}{
\begin{tabular}{lccccccc}
    \toprule
     & GPT 5.2 & Claude 4.6 & Gemini 3 Pro & Llama 3.3 & Qwen 3.5 & GLM 5 & Minimax 2.5 \\
    \midrule
    $n$ & 10 & 153 & 95 & 29 & 18 & 37 & 46 \\
    Pearson $r$ & 0.393 & 0.395 & 0.387 & 0.572 & 0.404 & 0.247 & 0.329 \\
    95\% CI & [-0.31, 0.82] & [0.25, 0.52] & [0.20, 0.55] & [0.26, 0.78] & [-0.08, 0.73] & [-0.08, 0.53] & [0.04, 0.56] \\
    \bottomrule
\end{tabular}
}
    \caption{\textbf{Correlation between Tag-Pair Direction of Directed and Undirected Edges.} Pearson correlation coefficients for all LLMs are distinctly positive, with most 95\% confidence intervals staying within the positive correlation area. The fact that the correlation is not perfect further motivates our decision to move from typing to tagging, as single tag pairs (types) can easily result in wrong predictions, while, taken together, the set of tag pairs is informative.}
    \label{tab:pearson_tags}
\end{table}

\subsection{Additional Algorithm Parameters}
\label{sec:implementationDetails}

In addition to the algorithm parameters discussed in the main text, the code realization of the algorithm considers two additional parameters: a `specificity prior' and `tag set deduplication'. Both parameters were dedicated to balancing the influence of certain tagging sets, but turned out not to play a major role in the eventual performance of the algorithm:

\textbf{Specificity Prior.} We furthermore considered the use of a `specificity prior' that was intended to reward in-group specificity by weighting tags occurring in specific sub-parts of the causal graphs more strongly. Experiments, however, found no empirical improvements from the introduction of this weighting term. A full description is given in App.~\ref{appx:specificityPrior} for completeness.

\textbf{Tag Set Deduplication.} As the same concept might be described by similar tags, we always deduplicate tags covering the same set of variables to avoid stronger weighting of certain concepts due to duplicate tags. To further differentiate from a typing approach, we optionally exclude singleton tagging sets. In our experiments, we measure the performance of both variants. We observed no major qualitative differences but found that the best-performing configuration included singular tagging sets.

\subsection{Best Parameter Configurations}
\label{sec:best_configs}
There are a total of 560 possible parameter configurations per dataset and seed.
In Tab.~\ref{tab:best_configs}, we show the 10 best parameter configurations overall.
As in the main paper, these are determined by their average F\textsubscript{1} score across all datasets and seeds to enable easy comparability between datasets (since SHD and SID values tend to be much larger for bigger datasets).
It can be seen that the top ten scores are not far from each other, indicating a certain amount of robustness of our results with respect to the set of all possible parameters.

Tagging on top of the PC performs worse than the GES variant, and is achieved using tags from Gemini 3 Pro, requiring only 1 sample per tag pair, without removing single tags, using the specificity prior, without applying Meek rules after every step, and by redirecting edges with iterative updates and considering the edge under consideration of being redirected as part of the evidence.
For the typing approaches, the best performing tags are generated by the following LLMs (consider App.~\ref{sec:all_evals} for more information on t-propagation):
\begin{itemize}
    \item Typed-PC (Naive): Claude 4.6,
    \item Typed-PC (Maj): Gemini 3 Pro,
    \item PC + t-Propagation: GLM 5,
    \item GES + t-Propagation: GLM 5.
\end{itemize}

In Tab.~\ref{tab:best_f1_configs}, we consider the best parameter configuration for each LLM.
We observe that the choice of LLM used for tagging does not appear to play a crucial role in obtaining good results.

Tab.~\ref{tab:tagged_ges_sensitivity_non_llm} shows the difference in F\textsubscript{1} score when changing one parameter compared to the optimal configuration (Tagged-GES).
Although the differences between LLMs are also relatively small, the choice of the LLM (i.e., the set of tags) is clearly the most dominant hyperparameter.
The second most impactful parameter reduces the average F\textsubscript{1} score by less than $0.01$, and the remaining ones make an even smaller difference.

\begin{table}[h]
    \centering
    \resizebox{\textwidth}{!}{
\begin{tabular}{cccccccc|c}
    \toprule
    LLM & Min-Samples & Fewer Tags & S. Prior & Always Meek & Redirect & Strategy & Include Edge & F\textsubscript{1} \\
    \midrule
    Claude 4.6 & 2 & False & False & False & True & 0 & True & 0.7954 \\
    Claude 4.6 & 2 & True & False & False & True & 0 & True & 0.7953 \\
    Claude 4.6 & 2 & False & False & False & True & 1 & True & 0.7948 \\
    Claude 4.6 & 2 & True & False & False & True & 1 & True & 0.7947 \\
    Claude 4.6 & 2 & True & False & True & True & 0 & True & 0.7946 \\
    Claude 4.6 & 2 & False & False & True & True & 0 & True & 0.7945 \\
    Claude 4.6 & 2 & True & False & True & True & 1 & True & 0.7940 \\
    Claude 4.6 & 2 & False & False & True & True & 1 & True & 0.7939 \\
    Claude 4.6 & 1 & False & False & False & True & 0 & True & 0.7935 \\
    Claude 4.6 & 1 & True & False & False & True & 0 & True & 0.7934 \\
    \bottomrule
\end{tabular}
    }
    \caption{\textbf{10 Best Configurations by F\textsubscript{1} score.} The parameters (columns) are: which LLM is used for determining tags (LLM), the number of tag combinations that must have been observed in order to allow for directing an edge (Min-Samples), whether to include all tags or remove those that are only assigned to single variables (Fewer Tags), whether to apply the specificity prior (S. Prior, see Eq.~\ref{eq:specificity}), whether to apply Meek rules after every direction or only at the end (Always Meek), whether to redirect edges at all (Redirect), which strategy to use for updating the evidence during edge redirection (Strategy; with 0 indicating iterative updates and 1 no updates), and whether to include the edge that is under consideration for being redirected as part of the evidence to decide on redirection (Include Edge). The score in the final column is the average F\textsubscript{1} score across all seeds and datasets. All results are achieved on the CPDAG from GES (results on the CPDAG output by PC performed worse).}
    \label{tab:best_configs}
\end{table}

\begin{table}[h]
    \centering
    \resizebox{\textwidth}{!}{
\begin{tabular}{cccccccc|c}
    \toprule
    LLM & Min-Samples & Fewer Tags & S. Prior & Always Meek & Redirect & Strategy & Include Edge & F\textsubscript{1} \\
    \midrule
    Claude 4.6 & 2 & False & False & False & True & 0 & True & 0.7954 \\
    Gemini 3 Pro & 2 & True & False & True & True & 0 & True & 0.7893 \\
    Qwen 3.5 & 1 & False & True & True & True & 1 & True & 0.7780 \\
    Llama 3.3 & 1 & False & False & True & False & - & - & 0.7754 \\
    GPT 5.2 & 1 & False & True & True & True & 1 & True & 0.7710 \\
    Minimax 2.5 & 2 & False & False & False & True & 1 & True & 0.7678 \\
    GLM 5 & 2 & False & False & False & True & 1 & True & 0.7656 \\
    \bottomrule
\end{tabular}
    }
    \caption{\textbf{Best Configurations by F\textsubscript{1} score per LLM.} The parameters (columns) are: which LLM is used for determining tags (LLM), the number of tag combinations that must have been observed in order to allow for directing an edge (Min-Samples), whether to include all tags or remove those that are only assigned to single variables (Fewer Tags), whether to apply the specificity prior (S. Prior, see Eq.~\ref{eq:specificity}), whether to apply Meek rules after every direction or only at the end (Always Meek), whether to redirect edges at all (Redirect), which strategy to use for updating the evidence during edge redirection (Strategy; with 0 indicating iterative updates and 1 no updates), and whether to include the edge that is under consideration for being redirected as part of the evidence to decide on redirection (Include Edge). The score in the final column is the average F\textsubscript{1} score across all seeds and datasets. All results are achieved on the CPDAG from GES (results on the CPDAG output by PC performed worse). The performance of all LLMs is relatively close to each other, with some deviations in the optimal parameter for their respective configuration.}
    \label{tab:best_f1_configs}
\end{table}

\begin{table}[t]
\centering
\begin{tabular}{l|l|r}
\textbf{Parameter} & \textbf{Value} & \textbf{$\Delta F_1$} \\
\hline
LLM & Qwen 3.5 & -0.039881 \\
LLM & GPT 5.2 & -0.036041 \\
LLM & Llama 3.3 & -0.035907 \\
LLM & GLM 5 & -0.032763 \\
LLM & Minimax 2.5 & -0.027631 \\
LLM & Gemini 3 Pro & -0.018575 \\
\hline
Redirect & False & -0.007511 \\
\hline
S. Prior & True & -0.005633 \\
\hline
Include Edge & False & -0.005052 \\
\hline
Min-Samples & 1 & -0.001902 \\
\hline
Always Meek & True & -0.000924 \\
\hline
Strategy & 1 & -0.000574 \\
\hline
Fewer Tags & True & -0.000093 \\
\end{tabular}
\caption{\textbf{One-At-A-Time Sensitivity of Tagged-GES.} Each row shows the change in $F_1$ after varying a single hyperparameter while keeping all others fixed at the best setting. The difference in choosing the best tag set (i.e., LLM) is more important than any other hyperparameter.}
\label{tab:tagged_ges_sensitivity_non_llm}
\end{table}

\subsection{Leave-One-Out Cross-Validation}
\label{sec:loocv}
In this experiment, we repeat our main evaluation, but instead of choosing the globally-best parameter configuration, we determine the best parameter configuration based on the other 10 datasets (leave-one-out cross-validation; LOOCV).
As shown in Tab.~\ref{tab:avgBestRankedResultsLOOCV}, tagging drops very slightly, but remains better than its competitors, resulting in no qualitative difference to the parameter selection without LOOCV.
As many of the further specific and ablation experiments in this paper are run under different setups than the main evaluation, we adopted the best-performing parameter configuration from the main evaluation for all subsequent experiments, unless otherwise specified.

\begin{table*}[t]
    \centering
\resizebox{\textwidth}{!}{
\begin{tabular}{l|ccccccc}
& SHD & SHD\textsubscript{double} & SID\textsubscript{min} & SID\textsubscript{max} & Precision & Recall & F\textsubscript{1} \\
 & Ranks & Ranks & Ranks & Ranks & Ranks & Ranks & Ranks \\
\hline
PC & $4.16 {\scriptstyle \pm 0.40}$ & $4.10 {\scriptstyle \pm 0.36}$ & $3.00 {\scriptstyle \pm 0.30}$ & $3.83 {\scriptstyle \pm 0.49}$ & $3.01 {\scriptstyle \pm 0.42}$ & $4.26 {\scriptstyle \pm 0.36}$ & $4.28 {\scriptstyle \pm 0.34}$ \\
GES & $3.54 {\scriptstyle \pm 0.51}$ & $3.06 {\scriptstyle \pm 0.46}$ & $\mathbf{1.27} {\scriptstyle \pm 0.13}$ & $3.17 {\scriptstyle \pm 0.51}$ & $2.69 {\scriptstyle \pm 0.51}$ & $3.17 {\scriptstyle \pm 0.42}$ & $3.13 {\scriptstyle \pm 0.51}$ \\
\hline
Typed-PC (Naive) & $3.30 {\scriptstyle \pm 0.34}$ & $3.45 {\scriptstyle \pm 0.35}$ & $2.94 {\scriptstyle \pm 0.28}$ & $3.05 {\scriptstyle \pm 0.32}$ & $2.91 {\scriptstyle \pm 0.46}$ & $3.35 {\scriptstyle \pm 0.35}$ & $3.44 {\scriptstyle \pm 0.41}$ \\
Typed-PC (Maj.) & $2.85 {\scriptstyle \pm 0.28}$ & $3.11 {\scriptstyle \pm 0.35}$ & $3.21 {\scriptstyle \pm 0.37}$ & $3.12 {\scriptstyle \pm 0.36}$ & $2.68 {\scriptstyle \pm 0.34}$ & $2.94 {\scriptstyle \pm 0.28}$ & $3.05 {\scriptstyle \pm 0.28}$ \\
\hline
Tagged-PC (AntiV) & $3.61 {\scriptstyle \pm 0.48}$ & $4.05 {\scriptstyle \pm 0.42}$ & $4.34 {\scriptstyle \pm 0.32}$ & $3.83 {\scriptstyle \pm 0.45}$ & $4.25 {\scriptstyle \pm 0.46}$ & $3.69 {\scriptstyle \pm 0.40}$ & $4.11 {\scriptstyle \pm 0.44}$ \\
Tagged-PC & $2.89 {\scriptstyle \pm 0.33}$ & $3.18 {\scriptstyle \pm 0.41}$ & $3.30 {\scriptstyle \pm 0.39}$ & $3.08 {\scriptstyle \pm 0.51}$ & $3.28 {\scriptstyle \pm 0.44}$ & $3.12 {\scriptstyle \pm 0.29}$ & $3.17 {\scriptstyle \pm 0.38}$ \\
Tagged-GES & $\mathbf{2.15} {\scriptstyle \pm 0.56}$ & $\mathbf{1.88} {\scriptstyle \pm 0.41}$ & $1.56 {\scriptstyle \pm 0.21}$ & $\mathbf{1.84} {\scriptstyle \pm 0.42}$ & $\mathbf{2.55} {\scriptstyle \pm 0.53}$ & $\mathbf{1.84} {\scriptstyle \pm 0.42}$ & $\mathbf{1.85} {\scriptstyle \pm 0.49}$ \\
\end{tabular}
}
\caption{\textbf{Average \underline{Ranks} for Best Configurations over all Datasets (lower is better) using LOOCV.} The table presents the results of the method as tuned by Leave-One-Out Cross-Validation. Instead of using the best configuration across all datasets, hyperparameters are determined for individual datasets by tuning parameters while leaving out the dataset under test. Results are observed to worsen slightly in contrast to the joint best configuration (cf. Tab.~\ref{tab:avgBestRankedResults}), Tagged-GES remains the best method, and Tagged-PC remains a major improvement over PC.  ${}_{\pm y}$ indicates standard deviation.}
\label{tab:avgBestRankedResultsLOOCV}
\end{table*}

\subsection{Evaluation Results on All Datasets}
\label{sec:all_evals}
In the main body, we show the rankings of our evaluation in Tab.~\ref{tab:avgBestRankedResults}.
We now include the detailed results in Tab.~\ref{tab:detailedOverallResultsAppx}, where we show the unranked evaluation results per dataset.
Here, we also include some steps of \textit{t-Propagation} as proposed in \citet{brouillard2022typing}.
However, we were unable to apply the last step of \textit{t-Propagation}, where all type-consistent graphs are enumerated and aggregated into a final graph, since type consistency is not always satisfied using our datasets and types.
The results for t-Propagation in the table are thus obtained by skipping this step (only propagating types and applying Meek rules), which turned out to perform rather badly on top of PC, while being more competitive than our method on GES.
However, note that in smaller datasets where only a few edges need to be directed, using tags instead of types might not be necessary, as the information given by types could be sufficient.
Thus, on the 4 smallest datasets, GES with t-Propagation sometimes achieves better scores than our tagging approach.
For the remaining datasets, the tagging approach matches the t-Propagation in 2 cases and exceeds it in the remaining 5 cases, indicating an increased benefit of tagging over typing on larger, more complex datasets.
This also aligns with our theoretical analysis in App.~\ref{app:theory}.

\textbf{LLM-based Prediction.} As a complementary approach to data-based causal discovery, we consider the common pure LLM-based approach of \citet{jiralerspong2024efficient}. LLM-based methods that query for the individual pairwise existence of edges \citep{kiciman2023causal,zevcevic2023causal} between all possible pairs of variables require a quadratic number of queries. These approaches quickly become computationally infeasible for larger graphs and lead to unreliable discovery, as no information about previous predictions is leveraged per query, and the acyclicity of the recovered graph can not be guaranteed.
To this, the work of \citet{jiralerspong2024efficient} employs a breadth-first search, which starts by identifying root nodes (nodes without parents) from the complete set of nodes, and then traverses individual nodes, directly querying for all children of a particular node, excluding already visited nodes. The algorithm is linear in the number of steps required, and the recovered graph is acyclic by construction. The approach is denoted as `\textit{LLM Root Recursive}' in all results tables.

Alternatively, we employ LLMs on top of causal discovery methods. Given a discovered MEC, we query LLMs to indicate the direction of all undirected edges. Results on top of GES and PC are reported as `\textit{LLM on PC}' and `\textit{LLM on GES}', respectively.
While this approach often meets or exceeds the tag-based prediction accuracy, there are two aspects to consider when comparing them. For one, the prior knowledge is vastly different; tagging merely requires that some semantic similarity can be identified between variables, while direct LLM orienting only works if some knowledge about the causal relations themselves is already encoded in the LLM. Secondly, these datasets are well known throughout the CD literature and are used frequently when benchmarking CD methods. Thus, it is highly likely that they have been included as part of the training datasets for the tested LLMs. As data contamination through LLM memorization can not be ruled out, LLM performance might vary for novel domains.

\begin{table}[p]
    \centering
    \resizebox{0.99\textwidth}{!}{
\begin{tabular}{l|ccccccc}
\textbf{Evaluation Results} & SHD & SHD\textsubscript{double} & SID\textsubscript{min} & SID\textsubscript{max} & Precision & Recall & F\textsubscript{1} \\
\hline \hline
Dataset Cancer & & & & & & & \\
\hline
PC & $2.00 {\scriptstyle \pm 0.45}$ & $3.90 {\scriptstyle \pm 0.70}$ & $9.60 {\scriptstyle \pm 1.56}$ & $9.60 {\scriptstyle \pm 1.56}$ & $0.51 {\scriptstyle \pm 0.09}$ & $0.50 {\scriptstyle \pm 0.11}$ & $0.50 {\scriptstyle \pm 0.10}$ \\
GES & $0.80 {\scriptstyle \pm 1.60}$ & $0.80 {\scriptstyle \pm 1.60}$ & $\mathbf{0.20} {\scriptstyle \pm 0.40}$ & $2.20 {\scriptstyle \pm 4.40}$ & $0.80 {\scriptstyle \pm 0.40}$ & $0.80 {\scriptstyle \pm 0.40}$ & $0.80 {\scriptstyle \pm 0.40}$ \\
Typed-PC (Naive) & $2.00 {\scriptstyle \pm 0.45}$ & $3.90 {\scriptstyle \pm 0.70}$ & $9.60 {\scriptstyle \pm 1.56}$ & $9.60 {\scriptstyle \pm 1.56}$ & $0.51 {\scriptstyle \pm 0.09}$ & $0.50 {\scriptstyle \pm 0.11}$ & $0.50 {\scriptstyle \pm 0.10}$ \\
Typed-PC (Maj.) & $2.00 {\scriptstyle \pm 0.45}$ & $3.90 {\scriptstyle \pm 0.70}$ & $9.60 {\scriptstyle \pm 1.56}$ & $9.60 {\scriptstyle \pm 1.56}$ & $0.51 {\scriptstyle \pm 0.09}$ & $0.50 {\scriptstyle \pm 0.11}$ & $0.50 {\scriptstyle \pm 0.10}$ \\
PC + t-Propagation & $2.00 {\scriptstyle \pm 0.00}$ & $3.80 {\scriptstyle \pm 0.60}$ & $9.10 {\scriptstyle \pm 2.70}$ & $9.60 {\scriptstyle \pm 1.20}$ & $0.55 {\scriptstyle \pm 0.15}$ & $0.50 {\scriptstyle \pm 0.00}$ & $0.52 {\scriptstyle \pm 0.05}$ \\
GES + t-Propagation & $0.40 {\scriptstyle \pm 0.80}$ & $0.40 {\scriptstyle \pm 0.80}$ & $\mathbf{0.20} {\scriptstyle \pm 0.40}$ & $1.20 {\scriptstyle \pm 2.40}$ & $\mathbf{1.00} {\scriptstyle \pm 0.00}$ & $0.90 {\scriptstyle \pm 0.20}$ & $0.93 {\scriptstyle \pm 0.13}$ \\
Tagged-PC (AntiV) & $2.00 {\scriptstyle \pm 0.45}$ & $3.90 {\scriptstyle \pm 0.70}$ & $9.60 {\scriptstyle \pm 1.56}$ & $9.60 {\scriptstyle \pm 1.56}$ & $0.51 {\scriptstyle \pm 0.09}$ & $0.50 {\scriptstyle \pm 0.11}$ & $0.50 {\scriptstyle \pm 0.10}$ \\
Tagged-PC & $2.00 {\scriptstyle \pm 0.45}$ & $3.90 {\scriptstyle \pm 0.70}$ & $9.60 {\scriptstyle \pm 1.56}$ & $9.60 {\scriptstyle \pm 1.56}$ & $0.51 {\scriptstyle \pm 0.09}$ & $0.50 {\scriptstyle \pm 0.11}$ & $0.50 {\scriptstyle \pm 0.10}$ \\
Tagged-GES & $0.80 {\scriptstyle \pm 1.60}$ & $0.80 {\scriptstyle \pm 1.60}$ & $\mathbf{0.20} {\scriptstyle \pm 0.40}$ & $2.20 {\scriptstyle \pm 4.40}$ & $0.80 {\scriptstyle \pm 0.40}$ & $0.80 {\scriptstyle \pm 0.40}$ & $0.80 {\scriptstyle \pm 0.40}$ \\
\hline
LLM Root Recursive & $3.00 {\scriptstyle \pm 0.00}$ & $3.00 {\scriptstyle \pm 0.00}$ & $1.00 {\scriptstyle \pm 0.00}$ & $1.00 {\scriptstyle \pm 0.00}$ & $0.60 {\scriptstyle \pm 0.00}$ & $0.75 {\scriptstyle \pm 0.00}$ & $0.67 {\scriptstyle \pm 0.00}$ \\
LLM on PC & $2.00 {\scriptstyle \pm 0.45}$ & $3.90 {\scriptstyle \pm 0.70}$ & $9.60 {\scriptstyle \pm 1.56}$ & $9.60 {\scriptstyle \pm 1.56}$ & $0.51 {\scriptstyle \pm 0.09}$ & $0.50 {\scriptstyle \pm 0.11}$ & $0.50 {\scriptstyle \pm 0.10}$ \\
LLM on GES & $\mathbf{0.20} {\scriptstyle \pm 0.40}$ & $\mathbf{0.20} {\scriptstyle \pm 0.40}$ & $\mathbf{0.20} {\scriptstyle \pm 0.40}$ & $\mathbf{0.20} {\scriptstyle \pm 0.40}$ & $\mathbf{1.00} {\scriptstyle \pm 0.00}$ & $\mathbf{0.95} {\scriptstyle \pm 0.10}$ & $\mathbf{0.97} {\scriptstyle \pm 0.06}$ \\
\hline \hline
Dataset Earthquake & & & & & & & \\
\hline
PC & $0.60 {\scriptstyle \pm 1.50}$ & $0.80 {\scriptstyle \pm 1.83}$ & $\mathbf{0.00} {\scriptstyle \pm 0.00}$ & $1.40 {\scriptstyle \pm 4.20}$ & $0.90 {\scriptstyle \pm 0.30}$ & $0.90 {\scriptstyle \pm 0.30}$ & $0.90 {\scriptstyle \pm 0.30}$ \\
GES & $\mathbf{0.00} {\scriptstyle \pm 0.00}$ & $\mathbf{0.00} {\scriptstyle \pm 0.00}$ & $\mathbf{0.00} {\scriptstyle \pm 0.00}$ & $\mathbf{0.00} {\scriptstyle \pm 0.00}$ & $\mathbf{1.00} {\scriptstyle \pm 0.00}$ & $\mathbf{1.00} {\scriptstyle \pm 0.00}$ & $\mathbf{1.00} {\scriptstyle \pm 0.00}$ \\
Typed-PC (Naive) & $0.40 {\scriptstyle \pm 0.92}$ & $0.60 {\scriptstyle \pm 1.28}$ & $\mathbf{0.00} {\scriptstyle \pm 0.00}$ & $0.90 {\scriptstyle \pm 2.70}$ & $\mathbf{1.00} {\scriptstyle \pm 0.00}$ & $0.95 {\scriptstyle \pm 0.15}$ & $0.97 {\scriptstyle \pm 0.10}$ \\
Typed-PC (Maj.) & $0.40 {\scriptstyle \pm 0.92}$ & $0.60 {\scriptstyle \pm 1.28}$ & $\mathbf{0.00} {\scriptstyle \pm 0.00}$ & $0.90 {\scriptstyle \pm 2.70}$ & $\mathbf{1.00} {\scriptstyle \pm 0.00}$ & $0.95 {\scriptstyle \pm 0.15}$ & $0.97 {\scriptstyle \pm 0.10}$ \\
PC + t-Propagation & $2.40 {\scriptstyle \pm 0.92}$ & $4.40 {\scriptstyle \pm 1.20}$ & $9.00 {\scriptstyle \pm 3.00}$ & $10.80 {\scriptstyle \pm 2.75}$ & $0.50 {\scriptstyle \pm 0.22}$ & $0.45 {\scriptstyle \pm 0.15}$ & $0.47 {\scriptstyle \pm 0.16}$ \\
GES + t-Propagation & $\mathbf{0.00} {\scriptstyle \pm 0.00}$ & $\mathbf{0.00} {\scriptstyle \pm 0.00}$ & $\mathbf{0.00} {\scriptstyle \pm 0.00}$ & $\mathbf{0.00} {\scriptstyle \pm 0.00}$ & $\mathbf{1.00} {\scriptstyle \pm 0.00}$ & $\mathbf{1.00} {\scriptstyle \pm 0.00}$ & $\mathbf{1.00} {\scriptstyle \pm 0.00}$ \\
Tagged-PC (AntiV) & $0.40 {\scriptstyle \pm 0.92}$ & $0.60 {\scriptstyle \pm 1.28}$ & $\mathbf{0.00} {\scriptstyle \pm 0.00}$ & $0.90 {\scriptstyle \pm 2.70}$ & $\mathbf{1.00} {\scriptstyle \pm 0.00}$ & $0.95 {\scriptstyle \pm 0.15}$ & $0.97 {\scriptstyle \pm 0.10}$ \\
Tagged-PC & $0.60 {\scriptstyle \pm 1.50}$ & $0.80 {\scriptstyle \pm 1.83}$ & $\mathbf{0.00} {\scriptstyle \pm 0.00}$ & $1.40 {\scriptstyle \pm 4.20}$ & $0.90 {\scriptstyle \pm 0.30}$ & $0.90 {\scriptstyle \pm 0.30}$ & $0.90 {\scriptstyle \pm 0.30}$ \\
Tagged-GES & $\mathbf{0.00} {\scriptstyle \pm 0.00}$ & $\mathbf{0.00} {\scriptstyle \pm 0.00}$ & $\mathbf{0.00} {\scriptstyle \pm 0.00}$ & $\mathbf{0.00} {\scriptstyle \pm 0.00}$ & $\mathbf{1.00} {\scriptstyle \pm 0.00}$ & $\mathbf{1.00} {\scriptstyle \pm 0.00}$ & $\mathbf{1.00} {\scriptstyle \pm 0.00}$ \\
\hline
LLM Root Recursive & $4.00 {\scriptstyle \pm 0.00}$ & $4.00 {\scriptstyle \pm 0.00}$ & $10.00 {\scriptstyle \pm 0.00}$ & $10.00 {\scriptstyle \pm 0.00}$ & $0.00 {\scriptstyle \pm 0.00}$ & $0.00 {\scriptstyle \pm 0.00}$ & $0.00 {\scriptstyle \pm 0.00}$ \\
LLM on PC & $0.20 {\scriptstyle \pm 0.40}$ & $0.20 {\scriptstyle \pm 0.40}$ & $\mathbf{0.00} {\scriptstyle \pm 0.00}$ & $\mathbf{0.00} {\scriptstyle \pm 0.00}$ & $0.96 {\scriptstyle \pm 0.08}$ & $\mathbf{1.00} {\scriptstyle \pm 0.00}$ & $0.98 {\scriptstyle \pm 0.04}$ \\
LLM on GES & $\mathbf{0.00} {\scriptstyle \pm 0.00}$ & $\mathbf{0.00} {\scriptstyle \pm 0.00}$ & $\mathbf{0.00} {\scriptstyle \pm 0.00}$ & $\mathbf{0.00} {\scriptstyle \pm 0.00}$ & $\mathbf{1.00} {\scriptstyle \pm 0.00}$ & $\mathbf{1.00} {\scriptstyle \pm 0.00}$ & $\mathbf{1.00} {\scriptstyle \pm 0.00}$ \\
\hline \hline
Dataset Survey & & & & & & & \\
\hline
PC & $2.30 {\scriptstyle \pm 0.64}$ & $4.30 {\scriptstyle \pm 0.64}$ & $11.10 {\scriptstyle \pm 1.92}$ & $11.80 {\scriptstyle \pm 2.44}$ & $0.64 {\scriptstyle \pm 0.05}$ & $0.62 {\scriptstyle \pm 0.11}$ & $0.63 {\scriptstyle \pm 0.08}$ \\
GES & $4.20 {\scriptstyle \pm 1.89}$ & $4.30 {\scriptstyle \pm 1.79}$ & $8.50 {\scriptstyle \pm 2.06}$ & $15.80 {\scriptstyle \pm 7.52}$ & $0.48 {\scriptstyle \pm 0.48}$ & $0.30 {\scriptstyle \pm 0.31}$ & $0.36 {\scriptstyle \pm 0.37}$ \\
Typed-PC (Naive) & $2.40 {\scriptstyle \pm 0.66}$ & $4.40 {\scriptstyle \pm 0.66}$ & $11.50 {\scriptstyle \pm 1.86}$ & $12.30 {\scriptstyle \pm 2.33}$ & $0.64 {\scriptstyle \pm 0.05}$ & $0.60 {\scriptstyle \pm 0.11}$ & $0.62 {\scriptstyle \pm 0.09}$ \\
Typed-PC (Maj.) & $2.20 {\scriptstyle \pm 0.60}$ & $4.30 {\scriptstyle \pm 0.90}$ & $13.30 {\scriptstyle \pm 0.90}$ & $13.30 {\scriptstyle \pm 0.90}$ & $0.64 {\scriptstyle \pm 0.08}$ & $0.63 {\scriptstyle \pm 0.10}$ & $0.64 {\scriptstyle \pm 0.09}$ \\
PC + t-Propagation & $5.40 {\scriptstyle \pm 0.49}$ & $10.70 {\scriptstyle \pm 0.90}$ & $27.10 {\scriptstyle \pm 1.92}$ & $27.10 {\scriptstyle \pm 1.92}$ & $0.10 {\scriptstyle \pm 0.08}$ & $0.10 {\scriptstyle \pm 0.08}$ & $0.10 {\scriptstyle \pm 0.08}$ \\
GES + t-Propagation & $4.20 {\scriptstyle \pm 1.89}$ & $5.50 {\scriptstyle \pm 3.26}$ & $12.70 {\scriptstyle \pm 7.62}$ & $15.80 {\scriptstyle \pm 7.52}$ & $0.48 {\scriptstyle \pm 0.48}$ & $0.30 {\scriptstyle \pm 0.31}$ & $0.36 {\scriptstyle \pm 0.37}$ \\
Tagged-PC (AntiV) & $2.60 {\scriptstyle \pm 0.49}$ & $4.60 {\scriptstyle \pm 0.49}$ & $13.10 {\scriptstyle \pm 0.30}$ & $13.60 {\scriptstyle \pm 0.49}$ & $0.63 {\scriptstyle \pm 0.03}$ & $0.57 {\scriptstyle \pm 0.08}$ & $0.59 {\scriptstyle \pm 0.06}$ \\
Tagged-PC & $2.20 {\scriptstyle \pm 0.40}$ & $4.30 {\scriptstyle \pm 0.64}$ & $11.20 {\scriptstyle \pm 2.04}$ & $11.20 {\scriptstyle \pm 2.04}$ & $0.64 {\scriptstyle \pm 0.05}$ & $0.63 {\scriptstyle \pm 0.07}$ & $0.64 {\scriptstyle \pm 0.06}$ \\
Tagged-GES & $4.20 {\scriptstyle \pm 1.89}$ & $4.30 {\scriptstyle \pm 1.79}$ & $8.50 {\scriptstyle \pm 2.06}$ & $15.80 {\scriptstyle \pm 7.52}$ & $0.48 {\scriptstyle \pm 0.48}$ & $0.30 {\scriptstyle \pm 0.31}$ & $0.36 {\scriptstyle \pm 0.37}$ \\
\hline
LLM Root Recursive & $6.00 {\scriptstyle \pm 0.00}$ & $6.00 {\scriptstyle \pm 0.00}$ & $17.00 {\scriptstyle \pm 0.00}$ & $17.00 {\scriptstyle \pm 0.00}$ & $0.00 {\scriptstyle \pm 0.00}$ & $0.00 {\scriptstyle \pm 0.00}$ & $0.00 {\scriptstyle \pm 0.00}$ \\
LLM on PC & $2.10 {\scriptstyle \pm 0.30}$ & $4.10 {\scriptstyle \pm 0.30}$ & $11.10 {\scriptstyle \pm 1.92}$ & $11.10 {\scriptstyle \pm 1.92}$ & $0.66 {\scriptstyle \pm 0.02}$ & $0.65 {\scriptstyle \pm 0.05}$ & $0.65 {\scriptstyle \pm 0.04}$ \\
LLM on GES & $\mathbf{2.00} {\scriptstyle \pm 0.45}$ & $\mathbf{2.10} {\scriptstyle \pm 0.54}$ & $\mathbf{7.30} {\scriptstyle \pm 2.00}$ & $\mathbf{7.30} {\scriptstyle \pm 2.00}$ & $\mathbf{0.98} {\scriptstyle \pm 0.06}$ & $\mathbf{0.67} {\scriptstyle \pm 0.07}$ & $\mathbf{0.79} {\scriptstyle \pm 0.06}$ \\
\hline \hline
Dataset Asia & & & & & & & \\
\hline
PC & $6.10 {\scriptstyle \pm 0.30}$ & $6.40 {\scriptstyle \pm 0.80}$ & $14.40 {\scriptstyle \pm 3.90}$ & $28.00 {\scriptstyle \pm 2.68}$ & $0.95 {\scriptstyle \pm 0.15}$ & $0.25 {\scriptstyle \pm 0.00}$ & $0.39 {\scriptstyle \pm 0.02}$ \\
GES & $3.00 {\scriptstyle \pm 0.00}$ & $3.00 {\scriptstyle \pm 0.00}$ & $0.80 {\scriptstyle \pm 0.40}$ & $8.80 {\scriptstyle \pm 1.60}$ & $\mathbf{1.00} {\scriptstyle \pm 0.00}$ & $0.62 {\scriptstyle \pm 0.00}$ & $0.77 {\scriptstyle \pm 0.00}$ \\
Typed-PC (Naive) & $3.00 {\scriptstyle \pm 1.26}$ & $3.60 {\scriptstyle \pm 2.24}$ & $16.60 {\scriptstyle \pm 6.97}$ & $17.70 {\scriptstyle \pm 7.48}$ & $0.91 {\scriptstyle \pm 0.20}$ & $0.64 {\scriptstyle \pm 0.15}$ & $0.75 {\scriptstyle \pm 0.17}$ \\
Typed-PC (Maj.) & $3.40 {\scriptstyle \pm 1.20}$ & $4.40 {\scriptstyle \pm 2.24}$ & $20.00 {\scriptstyle \pm 8.39}$ & $21.10 {\scriptstyle \pm 8.38}$ & $0.84 {\scriptstyle \pm 0.21}$ & $0.59 {\scriptstyle \pm 0.15}$ & $0.69 {\scriptstyle \pm 0.17}$ \\
PC + t-Propagation & $3.40 {\scriptstyle \pm 0.66}$ & $4.40 {\scriptstyle \pm 0.66}$ & $21.10 {\scriptstyle \pm 0.54}$ & $21.10 {\scriptstyle \pm 0.54}$ & $0.81 {\scriptstyle \pm 0.05}$ & $0.59 {\scriptstyle \pm 0.06}$ & $0.68 {\scriptstyle \pm 0.05}$ \\
GES + t-Propagation & $\mathbf{0.80} {\scriptstyle \pm 0.40}$ & $\mathbf{0.80} {\scriptstyle \pm 0.40}$ & $0.80 {\scriptstyle \pm 0.40}$ & $0.80 {\scriptstyle \pm 0.40}$ & $\mathbf{1.00} {\scriptstyle \pm 0.00}$ & $0.90 {\scriptstyle \pm 0.05}$ & $\mathbf{0.95} {\scriptstyle \pm 0.03}$ \\
Tagged-PC (AntiV) & $2.60 {\scriptstyle \pm 0.80}$ & $2.80 {\scriptstyle \pm 1.25}$ & $14.40 {\scriptstyle \pm 3.90}$ & $14.40 {\scriptstyle \pm 3.90}$ & $0.95 {\scriptstyle \pm 0.11}$ & $0.69 {\scriptstyle \pm 0.08}$ & $0.80 {\scriptstyle \pm 0.09}$ \\
Tagged-PC & $5.10 {\scriptstyle \pm 0.30}$ & $5.40 {\scriptstyle \pm 0.92}$ & $14.90 {\scriptstyle \pm 5.39}$ & $24.20 {\scriptstyle \pm 3.12}$ & $0.93 {\scriptstyle \pm 0.16}$ & $0.38 {\scriptstyle \pm 0.00}$ & $0.53 {\scriptstyle \pm 0.04}$ \\
Tagged-GES & $3.00 {\scriptstyle \pm 0.00}$ & $3.00 {\scriptstyle \pm 0.00}$ & $0.80 {\scriptstyle \pm 0.40}$ & $8.80 {\scriptstyle \pm 1.60}$ & $\mathbf{1.00} {\scriptstyle \pm 0.00}$ & $0.62 {\scriptstyle \pm 0.00}$ & $0.77 {\scriptstyle \pm 0.00}$ \\
\hline
LLM Root Recursive & $4.00 {\scriptstyle \pm 0.00}$ & $4.00 {\scriptstyle \pm 0.00}$ & $\mathbf{0.00} {\scriptstyle \pm 0.00}$ & $\mathbf{0.00} {\scriptstyle \pm 0.00}$ & $0.67 {\scriptstyle \pm 0.00}$ & $\mathbf{1.00} {\scriptstyle \pm 0.00}$ & $0.80 {\scriptstyle \pm 0.00}$ \\
LLM on PC & $2.60 {\scriptstyle \pm 0.80}$ & $2.80 {\scriptstyle \pm 1.25}$ & $14.40 {\scriptstyle \pm 3.90}$ & $14.40 {\scriptstyle \pm 3.90}$ & $0.95 {\scriptstyle \pm 0.11}$ & $0.69 {\scriptstyle \pm 0.08}$ & $0.80 {\scriptstyle \pm 0.09}$ \\
LLM on GES & $\mathbf{0.80} {\scriptstyle \pm 0.40}$ & $\mathbf{0.80} {\scriptstyle \pm 0.40}$ & $0.80 {\scriptstyle \pm 0.40}$ & $0.80 {\scriptstyle \pm 0.40}$ & $\mathbf{1.00} {\scriptstyle \pm 0.00}$ & $0.90 {\scriptstyle \pm 0.05}$ & $\mathbf{0.95} {\scriptstyle \pm 0.03}$ \\
\multicolumn{8}{c}{\textit{continued on next page}}
\end{tabular}
}
\end{table}

\begin{table}[p]
    \centering
    \resizebox{\textwidth}{!}{
\begin{tabular}{l|ccccccc}
\textbf{Evaluation Results} & SHD & SHD\textsubscript{double} & SID\textsubscript{min} & SID\textsubscript{max} & Precision & Recall & F\textsubscript{1} \\
\hline \hline
Dataset Lucas & & & & & & & \\
\hline
PC & $1.00 {\scriptstyle \pm 0.00}$ & $1.00 {\scriptstyle \pm 0.00}$ & $\mathbf{0.00} {\scriptstyle \pm 0.00}$ & $7.00 {\scriptstyle \pm 0.00}$ & $\mathbf{1.00} {\scriptstyle \pm 0.00}$ & $0.92 {\scriptstyle \pm 0.00}$ & $0.96 {\scriptstyle \pm 0.00}$ \\
GES & $1.00 {\scriptstyle \pm 0.00}$ & $1.00 {\scriptstyle \pm 0.00}$ & $\mathbf{0.00} {\scriptstyle \pm 0.00}$ & $7.00 {\scriptstyle \pm 0.00}$ & $\mathbf{1.00} {\scriptstyle \pm 0.00}$ & $0.92 {\scriptstyle \pm 0.00}$ & $0.96 {\scriptstyle \pm 0.00}$ \\
Typed-PC (Naive) & $1.00 {\scriptstyle \pm 0.00}$ & $1.00 {\scriptstyle \pm 0.00}$ & $\mathbf{0.00} {\scriptstyle \pm 0.00}$ & $7.00 {\scriptstyle \pm 0.00}$ & $\mathbf{1.00} {\scriptstyle \pm 0.00}$ & $0.92 {\scriptstyle \pm 0.00}$ & $0.96 {\scriptstyle \pm 0.00}$ \\
Typed-PC (Maj.) & $1.00 {\scriptstyle \pm 0.00}$ & $1.00 {\scriptstyle \pm 0.00}$ & $\mathbf{0.00} {\scriptstyle \pm 0.00}$ & $7.00 {\scriptstyle \pm 0.00}$ & $\mathbf{1.00} {\scriptstyle \pm 0.00}$ & $0.92 {\scriptstyle \pm 0.00}$ & $0.96 {\scriptstyle \pm 0.00}$ \\
PC + t-Propagation & $12.00 {\scriptstyle \pm 0.00}$ & $12.00 {\scriptstyle \pm 0.00}$ & $\mathbf{0.00} {\scriptstyle \pm 0.00}$ & $99.00 {\scriptstyle \pm 0.00}$ & $0.00 {\scriptstyle \pm 0.00}$ & $0.00 {\scriptstyle \pm 0.00}$ & $0.00 {\scriptstyle \pm 0.00}$ \\
GES + t-Propagation & $1.00 {\scriptstyle \pm 0.00}$ & $1.00 {\scriptstyle \pm 0.00}$ & $\mathbf{0.00} {\scriptstyle \pm 0.00}$ & $7.00 {\scriptstyle \pm 0.00}$ & $\mathbf{1.00} {\scriptstyle \pm 0.00}$ & $0.92 {\scriptstyle \pm 0.00}$ & $0.96 {\scriptstyle \pm 0.00}$ \\
Tagged-PC (AntiV) & $2.00 {\scriptstyle \pm 0.00}$ & $4.00 {\scriptstyle \pm 0.00}$ & $16.00 {\scriptstyle \pm 0.00}$ & $16.00 {\scriptstyle \pm 0.00}$ & $0.83 {\scriptstyle \pm 0.00}$ & $0.83 {\scriptstyle \pm 0.00}$ & $0.83 {\scriptstyle \pm 0.00}$ \\
Tagged-PC & $\mathbf{0.00} {\scriptstyle \pm 0.00}$ & $\mathbf{0.00} {\scriptstyle \pm 0.00}$ & $\mathbf{0.00} {\scriptstyle \pm 0.00}$ & $\mathbf{0.00} {\scriptstyle \pm 0.00}$ & $\mathbf{1.00} {\scriptstyle \pm 0.00}$ & $\mathbf{1.00} {\scriptstyle \pm 0.00}$ & $\mathbf{1.00} {\scriptstyle \pm 0.00}$ \\
Tagged-GES & $\mathbf{0.00} {\scriptstyle \pm 0.00}$ & $\mathbf{0.00} {\scriptstyle \pm 0.00}$ & $\mathbf{0.00} {\scriptstyle \pm 0.00}$ & $\mathbf{0.00} {\scriptstyle \pm 0.00}$ & $\mathbf{1.00} {\scriptstyle \pm 0.00}$ & $\mathbf{1.00} {\scriptstyle \pm 0.00}$ & $\mathbf{1.00} {\scriptstyle \pm 0.00}$ \\
\hline
LLM Root Recursive & $12.00 {\scriptstyle \pm 0.00}$ & $12.00 {\scriptstyle \pm 0.00}$ & $51.00 {\scriptstyle \pm 0.00}$ & $51.00 {\scriptstyle \pm 0.00}$ & $0.00 {\scriptstyle \pm 0.00}$ & $0.00 {\scriptstyle \pm 0.00}$ & $0.00 {\scriptstyle \pm 0.00}$ \\
LLM on PC & $\mathbf{0.00} {\scriptstyle \pm 0.00}$ & $\mathbf{0.00} {\scriptstyle \pm 0.00}$ & $\mathbf{0.00} {\scriptstyle \pm 0.00}$ & $\mathbf{0.00} {\scriptstyle \pm 0.00}$ & $\mathbf{1.00} {\scriptstyle \pm 0.00}$ & $\mathbf{1.00} {\scriptstyle \pm 0.00}$ & $\mathbf{1.00} {\scriptstyle \pm 0.00}$ \\
LLM on GES & $\mathbf{0.00} {\scriptstyle \pm 0.00}$ & $\mathbf{0.00} {\scriptstyle \pm 0.00}$ & $\mathbf{0.00} {\scriptstyle \pm 0.00}$ & $\mathbf{0.00} {\scriptstyle \pm 0.00}$ & $\mathbf{1.00} {\scriptstyle \pm 0.00}$ & $\mathbf{1.00} {\scriptstyle \pm 0.00}$ & $\mathbf{1.00} {\scriptstyle \pm 0.00}$ \\
\hline \hline
Dataset Child & & & & & & & \\
\hline
PC & $5.20 {\scriptstyle \pm 1.40}$ & $8.60 {\scriptstyle \pm 2.24}$ & $76.00 {\scriptstyle \pm 22.26}$ & $100.80 {\scriptstyle \pm 26.43}$ & $0.85 {\scriptstyle \pm 0.04}$ & $0.80 {\scriptstyle \pm 0.06}$ & $0.82 {\scriptstyle \pm 0.05}$ \\
GES & $12.00 {\scriptstyle \pm 0.00}$ & $12.00 {\scriptstyle \pm 0.00}$ & $\mathbf{0.00} {\scriptstyle \pm 0.00}$ & $228.00 {\scriptstyle \pm 0.00}$ & $\mathbf{1.00} {\scriptstyle \pm 0.00}$ & $0.52 {\scriptstyle \pm 0.00}$ & $0.68 {\scriptstyle \pm 0.00}$ \\
Typed-PC (Naive) & $4.30 {\scriptstyle \pm 2.05}$ & $7.50 {\scriptstyle \pm 3.17}$ & $69.60 {\scriptstyle \pm 26.76}$ & $78.20 {\scriptstyle \pm 34.51}$ & $0.86 {\scriptstyle \pm 0.05}$ & $0.83 {\scriptstyle \pm 0.09}$ & $0.85 {\scriptstyle \pm 0.07}$ \\
Typed-PC (Maj.) & $2.60 {\scriptstyle \pm 1.20}$ & $4.70 {\scriptstyle \pm 2.28}$ & $38.10 {\scriptstyle \pm 23.41}$ & $44.30 {\scriptstyle \pm 20.53}$ & $0.91 {\scriptstyle \pm 0.05}$ & $0.90 {\scriptstyle \pm 0.05}$ & $0.91 {\scriptstyle \pm 0.05}$ \\
PC + t-Propagation & $11.10 {\scriptstyle \pm 2.07}$ & $20.60 {\scriptstyle \pm 3.95}$ & $156.00 {\scriptstyle \pm 25.65}$ & $184.40 {\scriptstyle \pm 29.29}$ & $0.59 {\scriptstyle \pm 0.08}$ & $0.56 {\scriptstyle \pm 0.08}$ & $0.58 {\scriptstyle \pm 0.08}$ \\
GES + t-Propagation & $3.00 {\scriptstyle \pm 0.00}$ & $3.00 {\scriptstyle \pm 0.00}$ & $\mathbf{0.00} {\scriptstyle \pm 0.00}$ & $44.00 {\scriptstyle \pm 0.00}$ & $\mathbf{1.00} {\scriptstyle \pm 0.00}$ & $0.88 {\scriptstyle \pm 0.00}$ & $0.94 {\scriptstyle \pm 0.00}$ \\
Tagged-PC (AntiV) & $3.30 {\scriptstyle \pm 0.46}$ & $6.50 {\scriptstyle \pm 0.81}$ & $55.50 {\scriptstyle \pm 2.20}$ & $55.50 {\scriptstyle \pm 2.20}$ & $0.87 {\scriptstyle \pm 0.02}$ & $0.87 {\scriptstyle \pm 0.02}$ & $0.87 {\scriptstyle \pm 0.02}$ \\
Tagged-PC & $1.30 {\scriptstyle \pm 1.19}$ & $2.50 {\scriptstyle \pm 2.42}$ & $21.00 {\scriptstyle \pm 23.48}$ & $21.00 {\scriptstyle \pm 23.48}$ & $0.95 {\scriptstyle \pm 0.05}$ & $0.95 {\scriptstyle \pm 0.05}$ & $0.95 {\scriptstyle \pm 0.05}$ \\
Tagged-GES & $3.00 {\scriptstyle \pm 0.00}$ & $3.00 {\scriptstyle \pm 0.00}$ & $\mathbf{0.00} {\scriptstyle \pm 0.00}$ & $44.00 {\scriptstyle \pm 0.00}$ & $\mathbf{1.00} {\scriptstyle \pm 0.00}$ & $0.88 {\scriptstyle \pm 0.00}$ & $0.94 {\scriptstyle \pm 0.00}$ \\
\hline
LLM Root Recursive & $8.00 {\scriptstyle \pm 0.00}$ & $8.00 {\scriptstyle \pm 0.00}$ & $54.00 {\scriptstyle \pm 0.00}$ & $54.00 {\scriptstyle \pm 0.00}$ & $0.84 {\scriptstyle \pm 0.00}$ & $0.84 {\scriptstyle \pm 0.00}$ & $0.84 {\scriptstyle \pm 0.00}$ \\
LLM on PC & $4.00 {\scriptstyle \pm 1.26}$ & $7.90 {\scriptstyle \pm 2.62}$ & $76.80 {\scriptstyle \pm 22.99}$ & $76.80 {\scriptstyle \pm 22.99}$ & $0.84 {\scriptstyle \pm 0.05}$ & $0.84 {\scriptstyle \pm 0.05}$ & $0.84 {\scriptstyle \pm 0.05}$ \\
LLM on GES & $\mathbf{1.00} {\scriptstyle \pm 0.00}$ & $\mathbf{2.00} {\scriptstyle \pm 0.00}$ & $4.00 {\scriptstyle \pm 0.00}$ & $\mathbf{4.00} {\scriptstyle \pm 0.00}$ & $0.96 {\scriptstyle \pm 0.00}$ & $\mathbf{0.96} {\scriptstyle \pm 0.00}$ & $\mathbf{0.96} {\scriptstyle \pm 0.00}$ \\
\hline \hline
Dataset Alarm & & & & & & & \\
\hline
PC & $11.50 {\scriptstyle \pm 3.96}$ & $13.40 {\scriptstyle \pm 5.71}$ & $58.80 {\scriptstyle \pm 39.62}$ & $125.90 {\scriptstyle \pm 68.80}$ & $0.93 {\scriptstyle \pm 0.07}$ & $0.79 {\scriptstyle \pm 0.07}$ & $0.85 {\scriptstyle \pm 0.07}$ \\
GES & $8.80 {\scriptstyle \pm 2.18}$ & $9.90 {\scriptstyle \pm 2.70}$ & $47.40 {\scriptstyle \pm 9.87}$ & $96.20 {\scriptstyle \pm 12.66}$ & $0.94 {\scriptstyle \pm 0.03}$ & $0.84 {\scriptstyle \pm 0.03}$ & $0.89 {\scriptstyle \pm 0.03}$ \\
Typed-PC (Naive) & $9.10 {\scriptstyle \pm 2.51}$ & $10.90 {\scriptstyle \pm 3.83}$ & $53.40 {\scriptstyle \pm 35.37}$ & $93.90 {\scriptstyle \pm 36.60}$ & $0.94 {\scriptstyle \pm 0.05}$ & $0.84 {\scriptstyle \pm 0.04}$ & $0.89 {\scriptstyle \pm 0.04}$ \\
Typed-PC (Maj.) & $8.40 {\scriptstyle \pm 1.43}$ & $11.50 {\scriptstyle \pm 1.69}$ & $90.10 {\scriptstyle \pm 18.08}$ & $99.90 {\scriptstyle \pm 15.97}$ & $0.91 {\scriptstyle \pm 0.02}$ & $0.85 {\scriptstyle \pm 0.02}$ & $0.88 {\scriptstyle \pm 0.02}$ \\
PC + t-Propagation & $25.70 {\scriptstyle \pm 2.83}$ & $45.30 {\scriptstyle \pm 5.39}$ & $450.00 {\scriptstyle \pm 70.04}$ & $482.50 {\scriptstyle \pm 56.05}$ & $0.53 {\scriptstyle \pm 0.08}$ & $0.48 {\scriptstyle \pm 0.07}$ & $0.50 {\scriptstyle \pm 0.07}$ \\
GES + t-Propagation & $7.80 {\scriptstyle \pm 2.18}$ & $8.90 {\scriptstyle \pm 2.70}$ & $47.40 {\scriptstyle \pm 9.87}$ & $87.20 {\scriptstyle \pm 12.66}$ & $0.94 {\scriptstyle \pm 0.03}$ & $0.86 {\scriptstyle \pm 0.03}$ & $0.90 {\scriptstyle \pm 0.03}$ \\
Tagged-PC (AntiV) & $8.90 {\scriptstyle \pm 1.22}$ & $13.40 {\scriptstyle \pm 1.69}$ & $115.00 {\scriptstyle \pm 26.50}$ & $115.00 {\scriptstyle \pm 26.50}$ & $0.88 {\scriptstyle \pm 0.03}$ & $0.84 {\scriptstyle \pm 0.02}$ & $0.86 {\scriptstyle \pm 0.02}$ \\
Tagged-PC & $7.10 {\scriptstyle \pm 3.11}$ & $9.20 {\scriptstyle \pm 4.85}$ & $72.40 {\scriptstyle \pm 45.53}$ & $81.70 {\scriptstyle \pm 50.81}$ & $0.93 {\scriptstyle \pm 0.06}$ & $0.88 {\scriptstyle \pm 0.05}$ & $0.91 {\scriptstyle \pm 0.05}$ \\
Tagged-GES & $\mathbf{4.50} {\scriptstyle \pm 1.28}$ & $\mathbf{4.50} {\scriptstyle \pm 1.28}$ & $\mathbf{18.40} {\scriptstyle \pm 0.80}$ & $36.40 {\scriptstyle \pm 0.80}$ & $\mathbf{0.97} {\scriptstyle \pm 0.02}$ & $\mathbf{0.93} {\scriptstyle \pm 0.01}$ & $\mathbf{0.95} {\scriptstyle \pm 0.01}$ \\
\hline
LLM Root Recursive & $87.00 {\scriptstyle \pm 0.00}$ & $87.00 {\scriptstyle \pm 0.00}$ & $24.00 {\scriptstyle \pm 0.00}$ & $\mathbf{24.00} {\scriptstyle \pm 0.00}$ & $0.34 {\scriptstyle \pm 0.00}$ & $0.91 {\scriptstyle \pm 0.00}$ & $0.49 {\scriptstyle \pm 0.00}$ \\
LLM on PC & $6.50 {\scriptstyle \pm 2.50}$ & $8.20 {\scriptstyle \pm 4.24}$ & $62.60 {\scriptstyle \pm 40.67}$ & $62.60 {\scriptstyle \pm 40.67}$ & $0.93 {\scriptstyle \pm 0.06}$ & $0.90 {\scriptstyle \pm 0.04}$ & $0.91 {\scriptstyle \pm 0.05}$ \\
LLM on GES & $4.60 {\scriptstyle \pm 1.80}$ & $5.70 {\scriptstyle \pm 2.33}$ & $47.40 {\scriptstyle \pm 9.87}$ & $47.40 {\scriptstyle \pm 9.87}$ & $0.95 {\scriptstyle \pm 0.03}$ & $0.93 {\scriptstyle \pm 0.02}$ & $0.94 {\scriptstyle \pm 0.03}$ \\
\hline \hline
Dataset Insurance & & & & & & & \\
\hline
PC & $22.00 {\scriptstyle \pm 2.61}$ & $30.20 {\scriptstyle \pm 4.14}$ & $371.30 {\scriptstyle \pm 37.41}$ & $393.30 {\scriptstyle \pm 50.29}$ & $0.78 {\scriptstyle \pm 0.04}$ & $0.59 {\scriptstyle \pm 0.05}$ & $0.67 {\scriptstyle \pm 0.05}$ \\
GES & $24.50 {\scriptstyle \pm 2.73}$ & $28.10 {\scriptstyle \pm 5.63}$ & $\mathbf{230.70} {\scriptstyle \pm 61.36}$ & $341.70 {\scriptstyle \pm 23.83}$ & $0.85 {\scriptstyle \pm 0.11}$ & $0.57 {\scriptstyle \pm 0.03}$ & $0.68 {\scriptstyle \pm 0.06}$ \\
Typed-PC (Naive) & $18.70 {\scriptstyle \pm 2.05}$ & $26.20 {\scriptstyle \pm 3.57}$ & $354.10 {\scriptstyle \pm 33.73}$ & $354.10 {\scriptstyle \pm 33.73}$ & $0.81 {\scriptstyle \pm 0.05}$ & $0.65 {\scriptstyle \pm 0.03}$ & $0.72 {\scriptstyle \pm 0.04}$ \\
Typed-PC (Maj.) & $20.70 {\scriptstyle \pm 3.49}$ & $26.80 {\scriptstyle \pm 5.40}$ & $346.70 {\scriptstyle \pm 58.53}$ & $374.20 {\scriptstyle \pm 58.37}$ & $0.83 {\scriptstyle \pm 0.06}$ & $0.61 {\scriptstyle \pm 0.07}$ & $0.70 {\scriptstyle \pm 0.06}$ \\
PC + t-Propagation & $26.10 {\scriptstyle \pm 3.11}$ & $40.70 {\scriptstyle \pm 6.42}$ & $451.50 {\scriptstyle \pm 41.94}$ & $451.80 {\scriptstyle \pm 41.61}$ & $0.64 {\scriptstyle \pm 0.08}$ & $0.51 {\scriptstyle \pm 0.06}$ & $0.57 {\scriptstyle \pm 0.07}$ \\
GES + t-Propagation & $19.90 {\scriptstyle \pm 4.66}$ & $23.50 {\scriptstyle \pm 7.58}$ & $\mathbf{230.70} {\scriptstyle \pm 61.36}$ & $312.70 {\scriptstyle \pm 26.03}$ & $0.86 {\scriptstyle \pm 0.10}$ & $0.66 {\scriptstyle \pm 0.07}$ & $0.74 {\scriptstyle \pm 0.08}$ \\
Tagged-PC (AntiV) & $19.90 {\scriptstyle \pm 3.56}$ & $28.60 {\scriptstyle \pm 6.33}$ & $413.50 {\scriptstyle \pm 66.30}$ & $413.50 {\scriptstyle \pm 66.30}$ & $0.78 {\scriptstyle \pm 0.07}$ & $0.63 {\scriptstyle \pm 0.07}$ & $0.69 {\scriptstyle \pm 0.07}$ \\
Tagged-PC & $20.90 {\scriptstyle \pm 2.21}$ & $30.00 {\scriptstyle \pm 4.00}$ & $389.60 {\scriptstyle \pm 48.01}$ & $390.20 {\scriptstyle \pm 48.46}$ & $0.77 {\scriptstyle \pm 0.05}$ & $0.61 {\scriptstyle \pm 0.04}$ & $0.68 {\scriptstyle \pm 0.04}$ \\
Tagged-GES & $18.40 {\scriptstyle \pm 4.15}$ & $22.40 {\scriptstyle \pm 6.55}$ & $250.30 {\scriptstyle \pm 37.97}$ & $259.30 {\scriptstyle \pm 30.59}$ & $0.86 {\scriptstyle \pm 0.08}$ & $0.69 {\scriptstyle \pm 0.06}$ & $0.76 {\scriptstyle \pm 0.07}$ \\
\hline
LLM Root Recursive & $53.00 {\scriptstyle \pm 0.00}$ & $53.00 {\scriptstyle \pm 0.00}$ & $238.00 {\scriptstyle \pm 0.00}$ & $238.00 {\scriptstyle \pm 0.00}$ & $0.49 {\scriptstyle \pm 0.00}$ & $0.67 {\scriptstyle \pm 0.00}$ & $0.57 {\scriptstyle \pm 0.00}$ \\
LLM on PC & $19.30 {\scriptstyle \pm 2.10}$ & $27.40 {\scriptstyle \pm 3.69}$ & $371.30 {\scriptstyle \pm 37.41}$ & $371.30 {\scriptstyle \pm 37.41}$ & $0.79 {\scriptstyle \pm 0.04}$ & $0.64 {\scriptstyle \pm 0.04}$ & $0.71 {\scriptstyle \pm 0.04}$ \\
LLM on GES & $\mathbf{17.10} {\scriptstyle \pm 5.63}$ & $\mathbf{20.70} {\scriptstyle \pm 8.56}$ & $\mathbf{230.70} {\scriptstyle \pm 61.36}$ & $\mathbf{230.70} {\scriptstyle \pm 61.36}$ & $\mathbf{0.87} {\scriptstyle \pm 0.10}$ & $\mathbf{0.71} {\scriptstyle \pm 0.09}$ & $\mathbf{0.78} {\scriptstyle \pm 0.09}$ \\
\multicolumn{8}{c}{\textit{continued on next page}}
\end{tabular}
}
\end{table}

\begin{table}[p]
    \centering
    \resizebox{\textwidth}{!}{
\begin{tabular}{l|ccccccc}
\textbf{Evaluation Results} & SHD & SHD\textsubscript{double} & SID\textsubscript{min} & SID\textsubscript{max} & Precision & Recall & F\textsubscript{1} \\
\hline \hline
Dataset Hailfinder & & & & & & & \\
\hline
PC & $46.00 {\scriptstyle \pm 2.83}$ & $49.40 {\scriptstyle \pm 4.18}$ & - & - & $0.83 {\scriptstyle \pm 0.06}$ & $0.39 {\scriptstyle \pm 0.04}$ & $0.53 {\scriptstyle \pm 0.04}$ \\
GES & $\mathbf{39.00} {\scriptstyle \pm 15.21}$ & $\mathbf{42.20} {\scriptstyle \pm 15.28}$ & - & - & $0.69 {\scriptstyle \pm 0.12}$ & $\mathbf{0.65} {\scriptstyle \pm 0.12}$ & $\mathbf{0.67} {\scriptstyle \pm 0.12}$ \\
Typed-PC (Naive) & $45.40 {\scriptstyle \pm 2.50}$ & $48.90 {\scriptstyle \pm 3.48}$ & - & - & $0.83 {\scriptstyle \pm 0.06}$ & $0.40 {\scriptstyle \pm 0.02}$ & $0.54 {\scriptstyle \pm 0.03}$ \\
Typed-PC (Maj.) & $45.80 {\scriptstyle \pm 2.23}$ & $48.70 {\scriptstyle \pm 2.65}$ & - & - & $\mathbf{0.84} {\scriptstyle \pm 0.05}$ & $0.40 {\scriptstyle \pm 0.00}$ & $0.54 {\scriptstyle \pm 0.01}$ \\
PC + t-Propagation & $62.80 {\scriptstyle \pm 3.03}$ & $80.80 {\scriptstyle \pm 5.40}$ & - & - & $0.36 {\scriptstyle \pm 0.06}$ & $0.14 {\scriptstyle \pm 0.02}$ & $0.20 {\scriptstyle \pm 0.03}$ \\
GES + t-Propagation & $\mathbf{39.00} {\scriptstyle \pm 15.21}$ & $\mathbf{42.20} {\scriptstyle \pm 15.28}$ & - & - & $0.69 {\scriptstyle \pm 0.12}$ & $\mathbf{0.65} {\scriptstyle \pm 0.12}$ & $\mathbf{0.67} {\scriptstyle \pm 0.12}$ \\
Tagged-PC (AntiV) & $49.30 {\scriptstyle \pm 2.69}$ & $59.20 {\scriptstyle \pm 3.03}$ & - & - & $0.63 {\scriptstyle \pm 0.06}$ & $0.34 {\scriptstyle \pm 0.02}$ & $0.44 {\scriptstyle \pm 0.03}$ \\
Tagged-PC & $41.90 {\scriptstyle \pm 3.45}$ & $44.70 {\scriptstyle \pm 3.93}$ & - & - & $0.80 {\scriptstyle \pm 0.04}$ & $0.45 {\scriptstyle \pm 0.04}$ & $0.58 {\scriptstyle \pm 0.04}$ \\
Tagged-GES & $\mathbf{39.00} {\scriptstyle \pm 15.21}$ & $\mathbf{42.20} {\scriptstyle \pm 15.28}$ & - & - & $0.69 {\scriptstyle \pm 0.12}$ & $\mathbf{0.65} {\scriptstyle \pm 0.12}$ & $\mathbf{0.67} {\scriptstyle \pm 0.12}$ \\
\hline
LLM Root Recursive & $180.00 {\scriptstyle \pm 0.00}$ & $180.00 {\scriptstyle \pm 0.00}$ & - & - & $0.13 {\scriptstyle \pm 0.00}$ & $0.32 {\scriptstyle \pm 0.00}$ & $0.19 {\scriptstyle \pm 0.00}$ \\
LLM on PC & $41.50 {\scriptstyle \pm 2.54}$ & $45.10 {\scriptstyle \pm 3.51}$ & - & - & $0.76 {\scriptstyle \pm 0.05}$ & $0.46 {\scriptstyle \pm 0.02}$ & $0.57 {\scriptstyle \pm 0.03}$ \\
LLM on GES & $\mathbf{39.00} {\scriptstyle \pm 15.21}$ & $\mathbf{42.20} {\scriptstyle \pm 15.28}$ & - & - & $0.69 {\scriptstyle \pm 0.12}$ & $\mathbf{0.65} {\scriptstyle \pm 0.12}$ & $\mathbf{0.67} {\scriptstyle \pm 0.12}$ \\
\hline \hline
Dataset Hepar2 & & & & & & & \\
\hline
PC & $94.80 {\scriptstyle \pm 4.85}$ & $126.60 {\scriptstyle \pm 9.84}$ & - & - & $0.50 {\scriptstyle \pm 0.07}$ & $0.28 {\scriptstyle \pm 0.04}$ & $0.35 {\scriptstyle \pm 0.05}$ \\
GES & $57.60 {\scriptstyle \pm 3.23}$ & $64.90 {\scriptstyle \pm 3.36}$ & - & - & $0.90 {\scriptstyle \pm 0.01}$ & $0.53 {\scriptstyle \pm 0.03}$ & $0.67 {\scriptstyle \pm 0.02}$ \\
Typed-PC (Naive) & $95.40 {\scriptstyle \pm 3.20}$ & $127.70 {\scriptstyle \pm 8.44}$ & - & - & $0.49 {\scriptstyle \pm 0.04}$ & $0.27 {\scriptstyle \pm 0.02}$ & $0.35 {\scriptstyle \pm 0.03}$ \\
Typed-PC (Maj.) & $75.10 {\scriptstyle \pm 5.43}$ & $86.70 {\scriptstyle \pm 9.97}$ & - & - & $0.77 {\scriptstyle \pm 0.06}$ & $0.44 {\scriptstyle \pm 0.04}$ & $0.56 {\scriptstyle \pm 0.05}$ \\
PC + t-Propagation & $83.40 {\scriptstyle \pm 3.50}$ & $106.90 {\scriptstyle \pm 7.37}$ & - & - & $0.62 {\scriptstyle \pm 0.05}$ & $0.37 {\scriptstyle \pm 0.03}$ & $0.46 {\scriptstyle \pm 0.03}$ \\
GES + t-Propagation & $53.40 {\scriptstyle \pm 1.80}$ & $61.30 {\scriptstyle \pm 2.19}$ & - & - & $0.90 {\scriptstyle \pm 0.01}$ & $0.57 {\scriptstyle \pm 0.01}$ & $0.69 {\scriptstyle \pm 0.01}$ \\
Tagged-PC (AntiV) & $77.30 {\scriptstyle \pm 2.15}$ & $93.20 {\scriptstyle \pm 5.13}$ & - & - & $0.71 {\scriptstyle \pm 0.04}$ & $0.42 {\scriptstyle \pm 0.01}$ & $0.53 {\scriptstyle \pm 0.02}$ \\
Tagged-PC & $94.20 {\scriptstyle \pm 9.90}$ & $127.90 {\scriptstyle \pm 19.41}$ & - & - & $0.48 {\scriptstyle \pm 0.13}$ & $0.28 {\scriptstyle \pm 0.08}$ & $0.35 {\scriptstyle \pm 0.10}$ \\
Tagged-GES & $\mathbf{50.30} {\scriptstyle \pm 1.55}$ & $\mathbf{55.00} {\scriptstyle \pm 2.32}$ & - & - & $\mathbf{0.94} {\scriptstyle \pm 0.02}$ & $\mathbf{0.59} {\scriptstyle \pm 0.01}$ & $\mathbf{0.73} {\scriptstyle \pm 0.01}$ \\
\hline
LLM Root Recursive & $127.00 {\scriptstyle \pm 0.00}$ & $127.00 {\scriptstyle \pm 0.00}$ & - & - & $0.42 {\scriptstyle \pm 0.00}$ & $0.08 {\scriptstyle \pm 0.00}$ & $0.14 {\scriptstyle \pm 0.00}$ \\
LLM on PC & $90.90 {\scriptstyle \pm 5.24}$ & $122.20 {\scriptstyle \pm 9.96}$ & - & - & $0.51 {\scriptstyle \pm 0.07}$ & $0.31 {\scriptstyle \pm 0.04}$ & $0.38 {\scriptstyle \pm 0.05}$ \\
LLM on GES & $53.70 {\scriptstyle \pm 1.79}$ & $62.50 {\scriptstyle \pm 2.16}$ & - & - & $0.89 {\scriptstyle \pm 0.01}$ & $0.57 {\scriptstyle \pm 0.01}$ & $0.69 {\scriptstyle \pm 0.01}$ \\
\hline \hline
Dataset Win95Pts & & & & & & & \\
\hline
PC & $58.80 {\scriptstyle \pm 4.17}$ & $66.10 {\scriptstyle \pm 6.28}$ & - & - & $\mathbf{0.85} {\scriptstyle \pm 0.05}$ & $0.52 {\scriptstyle \pm 0.03}$ & $0.65 {\scriptstyle \pm 0.04}$ \\
GES & $52.20 {\scriptstyle \pm 6.43}$ & $57.90 {\scriptstyle \pm 6.82}$ & - & - & $0.76 {\scriptstyle \pm 0.03}$ & $0.71 {\scriptstyle \pm 0.03}$ & $0.73 {\scriptstyle \pm 0.03}$ \\
Typed-PC (Naive) & $60.70 {\scriptstyle \pm 2.79}$ & $69.90 {\scriptstyle \pm 2.91}$ & - & - & $0.82 {\scriptstyle \pm 0.03}$ & $0.51 {\scriptstyle \pm 0.02}$ & $0.63 {\scriptstyle \pm 0.02}$ \\
Typed-PC (Maj.) & $55.40 {\scriptstyle \pm 4.34}$ & $64.70 {\scriptstyle \pm 8.31}$ & - & - & $0.83 {\scriptstyle \pm 0.06}$ & $0.55 {\scriptstyle \pm 0.04}$ & $0.66 {\scriptstyle \pm 0.04}$ \\
PC + t-Propagation & $83.00 {\scriptstyle \pm 3.55}$ & $122.70 {\scriptstyle \pm 5.35}$ & - & - & $0.44 {\scriptstyle \pm 0.03}$ & $0.31 {\scriptstyle \pm 0.03}$ & $0.36 {\scriptstyle \pm 0.03}$ \\
GES + t-Propagation & $48.40 {\scriptstyle \pm 6.28}$ & $55.30 {\scriptstyle \pm 6.90}$ & - & - & $0.76 {\scriptstyle \pm 0.03}$ & $0.74 {\scriptstyle \pm 0.03}$ & $0.75 {\scriptstyle \pm 0.03}$ \\
Tagged-PC (AntiV) & $51.60 {\scriptstyle \pm 3.23}$ & $59.70 {\scriptstyle \pm 5.44}$ & - & - & $0.84 {\scriptstyle \pm 0.04}$ & $0.59 {\scriptstyle \pm 0.03}$ & $0.69 {\scriptstyle \pm 0.03}$ \\
Tagged-PC & $48.70 {\scriptstyle \pm 3.61}$ & $57.00 {\scriptstyle \pm 6.36}$ & - & - & $0.84 {\scriptstyle \pm 0.04}$ & $0.61 {\scriptstyle \pm 0.02}$ & $0.71 {\scriptstyle \pm 0.03}$ \\
Tagged-GES & $45.90 {\scriptstyle \pm 5.96}$ & $50.00 {\scriptstyle \pm 6.29}$ & - & - & $0.79 {\scriptstyle \pm 0.03}$ & $0.76 {\scriptstyle \pm 0.03}$ & $0.77 {\scriptstyle \pm 0.03}$ \\
\hline
LLM Root Recursive & $149.00 {\scriptstyle \pm 0.00}$ & $149.00 {\scriptstyle \pm 0.00}$ & - & - & $0.24 {\scriptstyle \pm 0.00}$ & $0.15 {\scriptstyle \pm 0.00}$ & $0.19 {\scriptstyle \pm 0.00}$ \\
LLM on PC & $47.90 {\scriptstyle \pm 3.56}$ & $55.30 {\scriptstyle \pm 6.02}$ & - & - & $0.85 {\scriptstyle \pm 0.04}$ & $0.62 {\scriptstyle \pm 0.03}$ & $0.72 {\scriptstyle \pm 0.03}$ \\
LLM on GES & $\mathbf{44.00} {\scriptstyle \pm 6.07}$ & $\mathbf{49.60} {\scriptstyle \pm 6.41}$ & - & - & $0.78 {\scriptstyle \pm 0.03}$ & $\mathbf{0.78} {\scriptstyle \pm 0.03}$ & $\mathbf{0.78} {\scriptstyle \pm 0.03}$ \\
\end{tabular}
}
    \caption{\textbf{Absolute Scores of Best Configuration per Dataset.} For each method, the parameter configuration (including LLM) that performed best for itself was chosen. Split across two pages due to the large table size. Certain SID results were omitted due to the high computation time. ${}_{\pm y}$ indicates standard deviation.}
    \label{tab:detailedOverallResultsAppx}
\end{table}

\textbf{Investigating Specific Prediction Counts.}
We show the exact number of correct, incorrect, and abstained predictions on undirected edges for the best-performing parameter configuration of our main evaluation in Tab.~\ref{tab:specific_decisions}.
As expected from the success of our tagging approach in the previously reported metric, tagging predicts edges correctly far more often than incorrectly.
Further, several edges remain undirected, as evidence is insufficient for directing.
Also note that one undirected edge had no corresponding edge in the ground-truth graph (as the GES skeleton was incorrect at this place) and was omitted from this table.

\begin{table*}[t]
    \centering
\resizebox{\textwidth}{!}{
\begin{tabular}{lcccccccccc|c}
    \toprule
    Dataset & 0 & 1 & 2 & 3 & 4 & 5 & 6 & 7 & 8 & 9 & Average \\
    \midrule
    Cancer & \textcolor{blue}{0}/\textcolor{red}{0}/\textcolor{black}{0} & \textcolor{blue}{0}/\textcolor{red}{0}/\textcolor{black}{3} & \textcolor{blue}{0}/\textcolor{red}{0}/\textcolor{black}{0} & \textcolor{blue}{0}/\textcolor{red}{0}/\textcolor{black}{0} & \textcolor{blue}{0}/\textcolor{red}{0}/\textcolor{black}{0} & \textcolor{blue}{0}/\textcolor{red}{0}/\textcolor{black}{0} & \textcolor{blue}{0}/\textcolor{red}{0}/\textcolor{black}{0} & \textcolor{blue}{0}/\textcolor{red}{0}/\textcolor{black}{3} & \textcolor{blue}{0}/\textcolor{red}{0}/\textcolor{black}{0} & \textcolor{blue}{0}/\textcolor{red}{0}/\textcolor{black}{0} & \textcolor{blue}{0.0}/\textcolor{red}{0.0}/\textcolor{black}{0.6} \\
    Earthquake & \textcolor{blue}{0}/\textcolor{red}{0}/\textcolor{black}{0} & \textcolor{blue}{0}/\textcolor{red}{0}/\textcolor{black}{0} & \textcolor{blue}{0}/\textcolor{red}{0}/\textcolor{black}{0} & \textcolor{blue}{0}/\textcolor{red}{0}/\textcolor{black}{0} & \textcolor{blue}{0}/\textcolor{red}{0}/\textcolor{black}{0} & \textcolor{blue}{0}/\textcolor{red}{0}/\textcolor{black}{0} & \textcolor{blue}{0}/\textcolor{red}{0}/\textcolor{black}{0} & \textcolor{blue}{0}/\textcolor{red}{0}/\textcolor{black}{0} & \textcolor{blue}{0}/\textcolor{red}{0}/\textcolor{black}{0} & \textcolor{blue}{0}/\textcolor{red}{0}/\textcolor{black}{0} & \textcolor{blue}{0.0}/\textcolor{red}{0.0}/\textcolor{black}{0.0} \\
    Survey & \textcolor{blue}{0}/\textcolor{red}{0}/\textcolor{black}{4} & \textcolor{blue}{0}/\textcolor{red}{0}/\textcolor{black}{4} & \textcolor{blue}{0}/\textcolor{red}{0}/\textcolor{black}{1} & \textcolor{blue}{0}/\textcolor{red}{0}/\textcolor{black}{0} & \textcolor{blue}{0}/\textcolor{red}{0}/\textcolor{black}{0} & \textcolor{blue}{0}/\textcolor{red}{0}/\textcolor{black}{3} & \textcolor{blue}{0}/\textcolor{red}{0}/\textcolor{black}{1} & \textcolor{blue}{0}/\textcolor{red}{0}/\textcolor{black}{4} & \textcolor{blue}{0}/\textcolor{red}{0}/\textcolor{black}{4} & \textcolor{blue}{0}/\textcolor{red}{0}/\textcolor{black}{1} & \textcolor{blue}{0.0}/\textcolor{red}{0.0}/\textcolor{black}{2.2} \\
    Asia & \textcolor{blue}{0}/\textcolor{red}{0}/\textcolor{black}{2} & \textcolor{blue}{0}/\textcolor{red}{0}/\textcolor{black}{2} & \textcolor{blue}{0}/\textcolor{red}{0}/\textcolor{black}{3} & \textcolor{blue}{0}/\textcolor{red}{0}/\textcolor{black}{2} & \textcolor{blue}{0}/\textcolor{red}{0}/\textcolor{black}{2} & \textcolor{blue}{0}/\textcolor{red}{0}/\textcolor{black}{2} & \textcolor{blue}{0}/\textcolor{red}{0}/\textcolor{black}{2} & \textcolor{blue}{0}/\textcolor{red}{0}/\textcolor{black}{2} & \textcolor{blue}{0}/\textcolor{red}{0}/\textcolor{black}{2} & \textcolor{blue}{0}/\textcolor{red}{0}/\textcolor{black}{3} & \textcolor{blue}{0.0}/\textcolor{red}{0.0}/\textcolor{black}{2.2} \\
    Lucas & \textcolor{blue}{1}/\textcolor{red}{0}/\textcolor{black}{0} & \textcolor{blue}{1}/\textcolor{red}{0}/\textcolor{black}{0} & \textcolor{blue}{1}/\textcolor{red}{0}/\textcolor{black}{0} & \textcolor{blue}{1}/\textcolor{red}{0}/\textcolor{black}{0} & \textcolor{blue}{1}/\textcolor{red}{0}/\textcolor{black}{0} & \textcolor{blue}{1}/\textcolor{red}{0}/\textcolor{black}{0} & \textcolor{blue}{1}/\textcolor{red}{0}/\textcolor{black}{0} & \textcolor{blue}{1}/\textcolor{red}{0}/\textcolor{black}{0} & \textcolor{blue}{1}/\textcolor{red}{0}/\textcolor{black}{0} & \textcolor{blue}{1}/\textcolor{red}{0}/\textcolor{black}{0} & \textcolor{blue}{1.0}/\textcolor{red}{0.0}/\textcolor{black}{0.0} \\
    Child & \textcolor{blue}{8}/\textcolor{red}{0}/\textcolor{black}{4} & \textcolor{blue}{8}/\textcolor{red}{0}/\textcolor{black}{4} & \textcolor{blue}{8}/\textcolor{red}{0}/\textcolor{black}{4} & \textcolor{blue}{8}/\textcolor{red}{0}/\textcolor{black}{4} & \textcolor{blue}{8}/\textcolor{red}{0}/\textcolor{black}{4} & \textcolor{blue}{8}/\textcolor{red}{0}/\textcolor{black}{4} & \textcolor{blue}{8}/\textcolor{red}{0}/\textcolor{black}{4} & \textcolor{blue}{8}/\textcolor{red}{0}/\textcolor{black}{4} & \textcolor{blue}{8}/\textcolor{red}{0}/\textcolor{black}{4} & \textcolor{blue}{8}/\textcolor{red}{0}/\textcolor{black}{4} & \textcolor{blue}{8.0}/\textcolor{red}{0.0}/\textcolor{black}{4.0} \\
    Insurance & \textcolor{blue}{6}/\textcolor{red}{1}/\textcolor{black}{4} & \textcolor{blue}{5}/\textcolor{red}{0}/\textcolor{black}{0} & \textcolor{blue}{5}/\textcolor{red}{0}/\textcolor{black}{0} & \textcolor{blue}{5}/\textcolor{red}{1}/\textcolor{black}{5} & \textcolor{blue}{8}/\textcolor{red}{1}/\textcolor{black}{2} & \textcolor{blue}{5}/\textcolor{red}{1}/\textcolor{black}{5} & \textcolor{blue}{5}/\textcolor{red}{0}/\textcolor{black}{0} & \textcolor{blue}{5}/\textcolor{red}{0}/\textcolor{black}{0} & \textcolor{blue}{5}/\textcolor{red}{0}/\textcolor{black}{0} & \textcolor{blue}{5}/\textcolor{red}{0}/\textcolor{black}{0} & \textcolor{blue}{5.4}/\textcolor{red}{0.4}/\textcolor{black}{1.6} \\
    Alarm & \textcolor{blue}{3}/\textcolor{red}{0}/\textcolor{black}{1} & \textcolor{blue}{3}/\textcolor{red}{0}/\textcolor{black}{1} & \textcolor{blue}{4}/\textcolor{red}{0}/\textcolor{black}{1} & \textcolor{blue}{3}/\textcolor{red}{0}/\textcolor{black}{1} & \textcolor{blue}{3}/\textcolor{red}{0}/\textcolor{black}{1} & \textcolor{blue}{4}/\textcolor{red}{0}/\textcolor{black}{1} & \textcolor{blue}{3}/\textcolor{red}{0}/\textcolor{black}{1} & \textcolor{blue}{3}/\textcolor{red}{0}/\textcolor{black}{1} & \textcolor{blue}{3}/\textcolor{red}{0}/\textcolor{black}{1} & \textcolor{blue}{3}/\textcolor{red}{0}/\textcolor{black}{1} & \textcolor{blue}{3.2}/\textcolor{red}{0.0}/\textcolor{black}{1.0} \\
    Hailfinder & \textcolor{blue}{0}/\textcolor{red}{0}/\textcolor{black}{0} & \textcolor{blue}{0}/\textcolor{red}{0}/\textcolor{black}{0} & \textcolor{blue}{0}/\textcolor{red}{0}/\textcolor{black}{0} & \textcolor{blue}{0}/\textcolor{red}{0}/\textcolor{black}{0} & \textcolor{blue}{0}/\textcolor{red}{0}/\textcolor{black}{0} & \textcolor{blue}{0}/\textcolor{red}{0}/\textcolor{black}{0} & \textcolor{blue}{0}/\textcolor{red}{0}/\textcolor{black}{0} & \textcolor{blue}{0}/\textcolor{red}{0}/\textcolor{black}{0} & \textcolor{blue}{0}/\textcolor{red}{0}/\textcolor{black}{0} & \textcolor{blue}{0}/\textcolor{red}{0}/\textcolor{black}{0} & \textcolor{blue}{0.0}/\textcolor{red}{0.0}/\textcolor{black}{0.0} \\
    Hepar2 & \textcolor{blue}{2}/\textcolor{red}{0}/\textcolor{black}{0} & \textcolor{blue}{2}/\textcolor{red}{2}/\textcolor{black}{0} & \textcolor{blue}{0}/\textcolor{red}{0}/\textcolor{black}{1} & \textcolor{blue}{2}/\textcolor{red}{0}/\textcolor{black}{3} & \textcolor{blue}{3}/\textcolor{red}{0}/\textcolor{black}{1} & \textcolor{blue}{2}/\textcolor{red}{3}/\textcolor{black}{0} & \textcolor{blue}{10}/\textcolor{red}{0}/\textcolor{black}{0} & \textcolor{blue}{6}/\textcolor{red}{0}/\textcolor{black}{3} & \textcolor{blue}{10}/\textcolor{red}{0}/\textcolor{black}{0} & \textcolor{blue}{2}/\textcolor{red}{2}/\textcolor{black}{0} & \textcolor{blue}{3.9}/\textcolor{red}{0.7}/\textcolor{black}{0.8} \\
    Win95Pts & \textcolor{blue}{4}/\textcolor{red}{1}/\textcolor{black}{4} & \textcolor{blue}{3}/\textcolor{red}{1}/\textcolor{black}{3} & \textcolor{blue}{4}/\textcolor{red}{2}/\textcolor{black}{3} & \textcolor{blue}{4}/\textcolor{red}{1}/\textcolor{black}{4} & \textcolor{blue}{4}/\textcolor{red}{1}/\textcolor{black}{3} & \textcolor{blue}{4}/\textcolor{red}{1}/\textcolor{black}{4} & \textcolor{blue}{3}/\textcolor{red}{2}/\textcolor{black}{3} & \textcolor{blue}{3}/\textcolor{red}{0}/\textcolor{black}{4} & \textcolor{blue}{4}/\textcolor{red}{2}/\textcolor{black}{3} & \textcolor{blue}{3}/\textcolor{red}{1}/\textcolor{black}{3} & \textcolor{blue}{3.6}/\textcolor{red}{1.2}/\textcolor{black}{3.4} \\
    \bottomrule
\end{tabular}
}
\caption{\textbf{Number of Correct/Incorrect/Abstained Predictions per Dataset and Seed.} The number of correctly directed edges (blue, left number per cell) is substantially larger than the number of incorrectly directed edges (red, middle number per cell). However, many edges remain undirected, indicating an abstention of the tagging procedure (black, right number per cell).}
\label{tab:specific_decisions}
\end{table*}

\textbf{Incorrect Predictions.}
In Tab.~\ref{tab:incorrect_directions}, we show all incorrect directions produced by tagging on GES using the best-performing parameter configuration.\footnote{This evaluation excludes mistakes during the redirection step. We found only one mistake among several correct redirection choices. Since redirection happens relatively rarely overall, we do not focus on it here in detail.}
Any shown error is present in at least one seed (repeated mistakes are not listed multiple times).
We include one example tag pair that was relevant for influencing the tagging decision.
Overall, the number of errors is very small and, on this small set of instances, we observe no clear systematic patterns.

\begin{table}[t]
    \centering
    \begin{align}
 & \textbf{Hepar2} & & \nonumber \\
\text{``fat''} &~\rightarrow~ \text{``gallstones''} && (\text{``upstream\_cause''}~\rightarrow~\text{``patient\_history''}) \nonumber \\
\text{``choledocholithotomy''} &~\rightarrow~ \text{``gallstones''} && (\text{``biliary\_system''}~\rightarrow~\text{``abdominal\_finding''}) \nonumber \\
\text{``obesity''} &~\rightarrow~ \text{``diabetes''} && (\text{``upstream\_cause''}~\rightarrow~\text{``chronic\_condition''}) \nonumber \\
 & \textbf{Insurance} & & \nonumber \\
\text{``SeniorTrain''} &~\rightarrow~ \text{``Age''} && (\text{``age\_dependent''}~\rightarrow~\text{``person\_related''}) \nonumber \\
 & \textbf{Win95Pts} & & \nonumber \\
\text{``PrtStatMem''} &~\rightarrow~ \text{``PrtMem''} && (\text{``memory''}~\rightarrow~\text{``root\_cause''}) \nonumber \\
\text{``AppDtGnTm''} &~\rightarrow~ \text{``PrtSpool''} && (\text{``application''}~\rightarrow~\text{``intermediate\_state''}) \nonumber \\
\text{``PSERRMEM''} &~\rightarrow~ \text{``TstpsTxt''} && (\text{``memory''}~\rightarrow~\text{``postscript''}) \nonumber %
    \end{align}
    \vspace{-25pt}
        \caption{\textbf{Incorrectly Predicted Edge Directions.} Shown are edges that are directed incorrectly by tagging at least once across all 10 seeds. To the right, one tag pair is shown that was contained in the evidence for directing the edge incorrectly.
        }
    \label{tab:incorrect_directions}
\end{table}

\subsection{Evaluation on Ground-Truth CPDAG}
\label{sec:eval_cpdag}
In addition to our main evaluation, we considered the ground-truth CPDAG and applied tagging to discover whether our approach successfully directs undirected edges when the causal discovery algorithm is applied before identifying the CPDAG without error.
Tab.~\ref{tab:skeleton} shows that tagging does, on average, improve the given CPDAGs by directing more undirected edges correctly than incorrectly.
In addition to the ranked results, we include absolute scores on all datasets in Tab.~\ref{tab:skeleton_all}.

\begin{table}[th]
    \centering
    \resizebox{\textwidth}{!}{
\begin{tabular}{l|ccccccc}
& SHD & SHD\textsubscript{double} & SID\textsubscript{min} & SID\textsubscript{max} & Precision & Recall & F\textsubscript{1} \\
 & Ranks & Ranks & Ranks & Ranks & Ranks & Ranks & Ranks \\
\hline
GT CPDAG & $1.64 {\scriptstyle \pm 0.00}$ & $1.45 {\scriptstyle \pm 0.00}$ & $\mathbf{1.00} {\scriptstyle \pm 0.00}$ & $1.38 {\scriptstyle \pm 0.00}$ & $\mathbf{1.00} {\scriptstyle \pm 0.00}$ & $1.64 {\scriptstyle \pm 0.00}$ & $1.55 {\scriptstyle \pm 0.00}$ \\
Tagging on GT CPDAG & $\mathbf{1.00} {\scriptstyle \pm 0.00}$ & $\mathbf{1.18} {\scriptstyle \pm 0.00}$ & $1.25 {\scriptstyle \pm 0.00}$ & $\mathbf{1.12} {\scriptstyle \pm 0.00}$ & $1.36 {\scriptstyle \pm 0.00}$ & $\mathbf{1.00} {\scriptstyle \pm 0.00}$ & $\mathbf{1.09} {\scriptstyle \pm 0.00}$ \\
\end{tabular}
}
    \caption{\textbf{Average Ranks for Ground Truth CPDAG over all Datasets (lower is better).} We use the parameters that performed best in the main evaluation as shown in Table~\ref{tab:avgBestRankedResults}. SID\textsubscript{min} and Precision are highest on the GT CPDAG, as there are no errors in the graph, so these metrics can not be improved upon. However, the other metrics show that our approach, on average, improves the graph by directing more edges correctly than incorrectly. ${}_{\pm y}$ indicates std. deviation.}
    \label{tab:skeleton}
\end{table}

\begin{table}[h]
    \centering
    \resizebox{\textwidth}{!}{
\begin{tabular}{l|ccccccc}
\textbf{Evaluation Results} & SHD & SHD\textsubscript{double} & SID\textsubscript{min} & SID\textsubscript{max} & Precision & Recall & F\textsubscript{1} \\
\hline \hline
Dataset Cancer & & & & & & & \\
\hline
GT CPDAG & $\mathbf{0.00} {\scriptstyle \pm 0.00}$ & $\mathbf{0.00} {\scriptstyle \pm 0.00}$ & $\mathbf{0.00} {\scriptstyle \pm 0.00}$ & $\mathbf{0.00} {\scriptstyle \pm 0.00}$ & $\mathbf{1.00} {\scriptstyle \pm 0.00}$ & $\mathbf{1.00} {\scriptstyle \pm 0.00}$ & $\mathbf{1.00} {\scriptstyle \pm 0.00}$ \\
Tagging on GT CPDAG & $\mathbf{0.00} {\scriptstyle \pm 0.00}$ & $\mathbf{0.00} {\scriptstyle \pm 0.00}$ & $\mathbf{0.00} {\scriptstyle \pm 0.00}$ & $\mathbf{0.00} {\scriptstyle \pm 0.00}$ & $\mathbf{1.00} {\scriptstyle \pm 0.00}$ & $\mathbf{1.00} {\scriptstyle \pm 0.00}$ & $\mathbf{1.00} {\scriptstyle \pm 0.00}$ \\
\hline \hline
Dataset Earthquake & & & & & & & \\
\hline
GT CPDAG & $\mathbf{0.00} {\scriptstyle \pm 0.00}$ & $\mathbf{0.00} {\scriptstyle \pm 0.00}$ & $\mathbf{0.00} {\scriptstyle \pm 0.00}$ & $\mathbf{0.00} {\scriptstyle \pm 0.00}$ & $\mathbf{1.00} {\scriptstyle \pm 0.00}$ & $\mathbf{1.00} {\scriptstyle \pm 0.00}$ & $\mathbf{1.00} {\scriptstyle \pm 0.00}$ \\
Tagging on GT CPDAG & $\mathbf{0.00} {\scriptstyle \pm 0.00}$ & $\mathbf{0.00} {\scriptstyle \pm 0.00}$ & $\mathbf{0.00} {\scriptstyle \pm 0.00}$ & $\mathbf{0.00} {\scriptstyle \pm 0.00}$ & $\mathbf{1.00} {\scriptstyle \pm 0.00}$ & $\mathbf{1.00} {\scriptstyle \pm 0.00}$ & $\mathbf{1.00} {\scriptstyle \pm 0.00}$ \\
\hline \hline
Dataset Survey & & & & & & & \\
\hline
GT CPDAG & $\mathbf{0.00} {\scriptstyle \pm 0.00}$ & $\mathbf{0.00} {\scriptstyle \pm 0.00}$ & $\mathbf{0.00} {\scriptstyle \pm 0.00}$ & $\mathbf{0.00} {\scriptstyle \pm 0.00}$ & $\mathbf{1.00} {\scriptstyle \pm 0.00}$ & $\mathbf{1.00} {\scriptstyle \pm 0.00}$ & $\mathbf{1.00} {\scriptstyle \pm 0.00}$ \\
Tagging on GT CPDAG & $\mathbf{0.00} {\scriptstyle \pm 0.00}$ & $\mathbf{0.00} {\scriptstyle \pm 0.00}$ & $\mathbf{0.00} {\scriptstyle \pm 0.00}$ & $\mathbf{0.00} {\scriptstyle \pm 0.00}$ & $\mathbf{1.00} {\scriptstyle \pm 0.00}$ & $\mathbf{1.00} {\scriptstyle \pm 0.00}$ & $\mathbf{1.00} {\scriptstyle \pm 0.00}$ \\
\hline \hline
Dataset Asia & & & & & & & \\
\hline
GT CPDAG & $\mathbf{3.00} {\scriptstyle \pm 0.00}$ & $\mathbf{3.00} {\scriptstyle \pm 0.00}$ & $\mathbf{0.00} {\scriptstyle \pm 0.00}$ & $\mathbf{12.00} {\scriptstyle \pm 0.00}$ & $\mathbf{1.00} {\scriptstyle \pm 0.00}$ & $\mathbf{0.62} {\scriptstyle \pm 0.00}$ & $\mathbf{0.77} {\scriptstyle \pm 0.00}$ \\
Tagging on GT CPDAG & $\mathbf{3.00} {\scriptstyle \pm 0.00}$ & $\mathbf{3.00} {\scriptstyle \pm 0.00}$ & $\mathbf{0.00} {\scriptstyle \pm 0.00}$ & $\mathbf{12.00} {\scriptstyle \pm 0.00}$ & $\mathbf{1.00} {\scriptstyle \pm 0.00}$ & $\mathbf{0.62} {\scriptstyle \pm 0.00}$ & $\mathbf{0.77} {\scriptstyle \pm 0.00}$ \\
\hline \hline
Dataset Lucas & & & & & & & \\
\hline
GT CPDAG & $1.00 {\scriptstyle \pm 0.00}$ & $1.00 {\scriptstyle \pm 0.00}$ & $\mathbf{0.00} {\scriptstyle \pm 0.00}$ & $7.00 {\scriptstyle \pm 0.00}$ & $\mathbf{1.00} {\scriptstyle \pm 0.00}$ & $0.92 {\scriptstyle \pm 0.00}$ & $0.96 {\scriptstyle \pm 0.00}$ \\
Tagging on GT CPDAG & $\mathbf{0.00} {\scriptstyle \pm 0.00}$ & $\mathbf{0.00} {\scriptstyle \pm 0.00}$ & $\mathbf{0.00} {\scriptstyle \pm 0.00}$ & $\mathbf{0.00} {\scriptstyle \pm 0.00}$ & $\mathbf{1.00} {\scriptstyle \pm 0.00}$ & $\mathbf{1.00} {\scriptstyle \pm 0.00}$ & $\mathbf{1.00} {\scriptstyle \pm 0.00}$ \\
\hline \hline
Dataset Child & & & & & & & \\
\hline
GT CPDAG & $10.00 {\scriptstyle \pm 0.00}$ & $10.00 {\scriptstyle \pm 0.00}$ & $\mathbf{0.00} {\scriptstyle \pm 0.00}$ & $209.00 {\scriptstyle \pm 0.00}$ & $\mathbf{1.00} {\scriptstyle \pm 0.00}$ & $0.60 {\scriptstyle \pm 0.00}$ & $0.75 {\scriptstyle \pm 0.00}$ \\
Tagging on GT CPDAG & $\mathbf{2.00} {\scriptstyle \pm 0.00}$ & $\mathbf{3.00} {\scriptstyle \pm 0.00}$ & $20.00 {\scriptstyle \pm 0.00}$ & $\mathbf{43.00} {\scriptstyle \pm 0.00}$ & $0.96 {\scriptstyle \pm 0.00}$ & $\mathbf{0.92} {\scriptstyle \pm 0.00}$ & $\mathbf{0.94} {\scriptstyle \pm 0.00}$ \\
\hline \hline
Dataset Alarm & & & & & & & \\
\hline
GT CPDAG & $4.00 {\scriptstyle \pm 0.00}$ & $4.00 {\scriptstyle \pm 0.00}$ & $\mathbf{0.00} {\scriptstyle \pm 0.00}$ & $\mathbf{46.00} {\scriptstyle \pm 0.00}$ & $\mathbf{1.00} {\scriptstyle \pm 0.00}$ & $0.91 {\scriptstyle \pm 0.00}$ & $0.95 {\scriptstyle \pm 0.00}$ \\
Tagging on GT CPDAG & $\mathbf{2.00} {\scriptstyle \pm 0.00}$ & $\mathbf{3.00} {\scriptstyle \pm 0.00}$ & $42.00 {\scriptstyle \pm 0.00}$ & $60.00 {\scriptstyle \pm 0.00}$ & $0.98 {\scriptstyle \pm 0.00}$ & $\mathbf{0.96} {\scriptstyle \pm 0.00}$ & $\mathbf{0.97} {\scriptstyle \pm 0.00}$ \\
\hline \hline
Dataset Insurance & & & & & & & \\
\hline
GT CPDAG & $2.00 {\scriptstyle \pm 0.00}$ & $2.00 {\scriptstyle \pm 0.00}$ & $\mathbf{0.00} {\scriptstyle \pm 0.00}$ & $75.00 {\scriptstyle \pm 0.00}$ & $\mathbf{1.00} {\scriptstyle \pm 0.00}$ & $0.96 {\scriptstyle \pm 0.00}$ & $0.98 {\scriptstyle \pm 0.00}$ \\
Tagging on GT CPDAG & $\mathbf{1.00} {\scriptstyle \pm 0.00}$ & $\mathbf{1.00} {\scriptstyle \pm 0.00}$ & $\mathbf{0.00} {\scriptstyle \pm 0.00}$ & $\mathbf{49.00} {\scriptstyle \pm 0.00}$ & $\mathbf{1.00} {\scriptstyle \pm 0.00}$ & $\mathbf{0.98} {\scriptstyle \pm 0.00}$ & $\mathbf{0.99} {\scriptstyle \pm 0.00}$ \\
\hline \hline
Dataset Hailfinder & & & & & & & \\
\hline
GT CPDAG & $17.00 {\scriptstyle \pm 0.00}$ & $\mathbf{17.00} {\scriptstyle \pm 0.00}$ & - & - & $\mathbf{1.00} {\scriptstyle \pm 0.00}$ & $0.74 {\scriptstyle \pm 0.00}$ & $0.85 {\scriptstyle \pm 0.00}$ \\
Tagging on GT CPDAG & $\mathbf{9.00} {\scriptstyle \pm 0.00}$ & $18.00 {\scriptstyle \pm 0.00}$ & - & - & $0.86 {\scriptstyle \pm 0.00}$ & $\mathbf{0.86} {\scriptstyle \pm 0.00}$ & $\mathbf{0.86} {\scriptstyle \pm 0.00}$ \\
\hline \hline
Dataset Hepar2 & & & & & & & \\
\hline
GT CPDAG & $7.00 {\scriptstyle \pm 0.00}$ & $7.00 {\scriptstyle \pm 0.00}$ & - & - & $\mathbf{1.00} {\scriptstyle \pm 0.00}$ & $0.94 {\scriptstyle \pm 0.00}$ & $0.97 {\scriptstyle \pm 0.00}$ \\
Tagging on GT CPDAG & $\mathbf{3.00} {\scriptstyle \pm 0.00}$ & $\mathbf{3.00} {\scriptstyle \pm 0.00}$ & - & - & $\mathbf{1.00} {\scriptstyle \pm 0.00}$ & $\mathbf{0.98} {\scriptstyle \pm 0.00}$ & $\mathbf{0.99} {\scriptstyle \pm 0.00}$ \\
\hline \hline
Dataset Win95Pts & & & & & & & \\
\hline
GT CPDAG & $8.00 {\scriptstyle \pm 0.00}$ & $\mathbf{8.00} {\scriptstyle \pm 0.00}$ & - & - & $\mathbf{1.00} {\scriptstyle \pm 0.00}$ & $0.93 {\scriptstyle \pm 0.00}$ & $\mathbf{0.96} {\scriptstyle \pm 0.00}$ \\
Tagging on GT CPDAG & $\mathbf{7.00} {\scriptstyle \pm 0.00}$ & $11.00 {\scriptstyle \pm 0.00}$ & - & - & $0.96 {\scriptstyle \pm 0.00}$ & $\mathbf{0.94} {\scriptstyle \pm 0.00}$ & $0.95 {\scriptstyle \pm 0.00}$ \\
\end{tabular}
}
    \caption{\textbf{Absolute Scores for Ground Truth CPDAG over all Datasets.} Certain SID results were omitted due to the high computation time. We use the parameters that performed best in the main evaluation as shown in Table~\ref{tab:avgBestRankedResults}. Direction using tagging improves results on several datasets. ${}_{\pm y}$ indicates std. deviation.}
    \label{tab:skeleton_all}
\end{table}

\subsection{Different Calibration Thresholds}
\label{sec:vary_thresholds}
In this experiment, we want to investigate how the calibration threshold impacts tagging predictions.
The calibration threshold determines how high the average tag informative value must be for an undirected edge to be directed; the tagging algorithm abstains from making a decision otherwise.
If we want to analyze whether individual edges are directed correctly, we need to compare them to a ground truth.
As PC or GES might result in skeletons where undirected edges have no corresponding edge in the ground truth graph, and, hence, we can not easily determine whether the choice made by tagging was ``good'', we run a new experimental evaluation on the ground truth CPDAG using the best-performing parameter configuration for it obtained during the main evaluation.
While the main evaluation used the cross-validation procedure to determine the thresholds, we now disable this automatic computation and rerun our experiments for different thresholds from $0.5$ to $1.0$ in steps of $0.05$.\footnote{Usually, a probability must only reach that specific threshold (hence, $1.0$ is a valid, albeit strongly conservative choice), except for $0.5$, where it is hard-coded to accept any edge directions larger than $0.5$ (to avoid fully random decisions).}

Fig.~\ref{fig:thresholds} shows how the number of predictions quickly decreases with stricter thresholds.
In particular, no more incorrect predictions are made for the most conservative thresholds, although this comes at a cost of abstaining from most or all other predictions as well.

\begin{figure}
    \centering
    \includegraphics[width=0.5\linewidth]{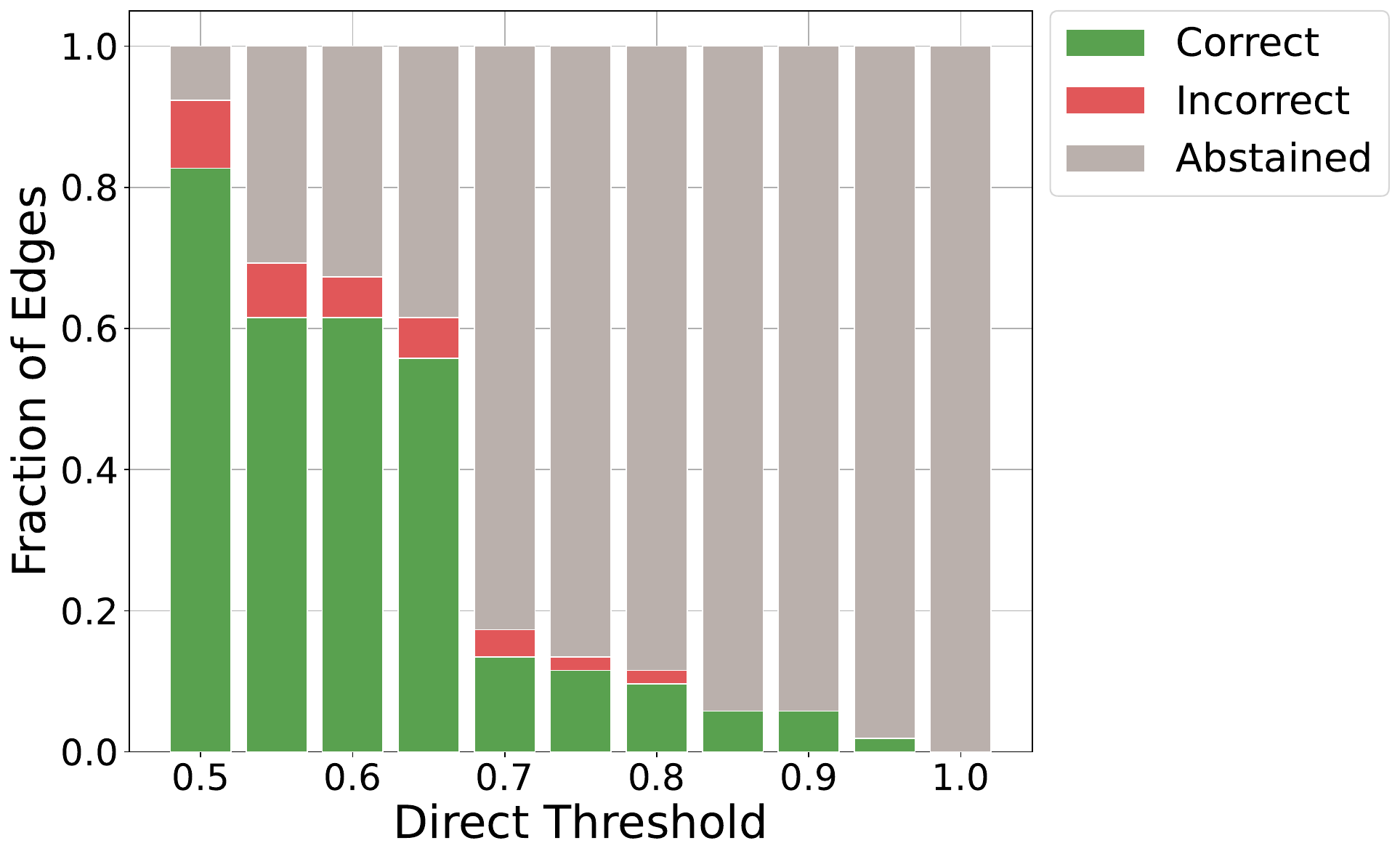}
    \caption{\textbf{Correct, Incorrect, and Abstained Decisions Across Different Thresholds.} Tagging applied on the ground truth CPDAG, comparing edge predictions with the ground truth graph. As the threshold becomes stricter, abstentions increase, and incorrect predictions vanish.}
    \label{fig:thresholds}
\end{figure}

\subsection{Laplace Correction Experiment}
\label{sec:laplace_correction}
Currently, tag pairs with small but clear evidence where all edges point towards the same direction are all assigned the same tag informative value, independently of the exact amount of evidence used.
To test whether an approach that takes the evidence number into account more directly can be beneficial for the tagging procedure, we ran our main experiment again, now applying Laplace correction on all tag pairs before computing the tag informative values.
Thus, tag pairs with little evidence indicate a much smaller preference for an edge direction than those supported by a large number of edges.
Our results in Tab.~\ref{tab:laplace} show no substantial difference, with results dropping just slightly.
While these experiments are insufficient for making well-founded claims about this, we assume this to be due to the relatively small size of the graphs, where it appears to be more helpful to use evidence to its fullest extent, ignoring epistemic uncertainty.
We think it is possible for the Laplace correction to show greater benefits in larger settings.

\begin{table*}[t]
    \centering
\resizebox{\textwidth}{!}{
\begin{tabular}{l|ccccccc}
& SHD & SHD\textsubscript{double} & SID\textsubscript{min} & SID\textsubscript{max} & Precision & Recall & F\textsubscript{1} \\
 & Ranks & Ranks & Ranks & Ranks & Ranks & Ranks & Ranks \\
\hline
PC & $3.96 {\scriptstyle \pm 0.42}$ & $3.94 {\scriptstyle \pm 0.36}$ & $2.92 {\scriptstyle \pm 0.32}$ & $3.54 {\scriptstyle \pm 0.54}$ & $2.95 {\scriptstyle \pm 0.39}$ & $4.06 {\scriptstyle \pm 0.38}$ & $4.09 {\scriptstyle \pm 0.37}$ \\
GES & $3.35 {\scriptstyle \pm 0.53}$ & $2.93 {\scriptstyle \pm 0.49}$ & $\mathbf{1.29} {\scriptstyle \pm 0.14}$ & $2.99 {\scriptstyle \pm 0.50}$ & $2.68 {\scriptstyle \pm 0.54}$ & $2.98 {\scriptstyle \pm 0.44}$ & $2.91 {\scriptstyle \pm 0.49}$ \\
\hline
Typed-PC (Naive) & $3.14 {\scriptstyle \pm 0.35}$ & $3.33 {\scriptstyle \pm 0.40}$ & $2.94 {\scriptstyle \pm 0.28}$ & $2.84 {\scriptstyle \pm 0.32}$ & $2.95 {\scriptstyle \pm 0.43}$ & $3.19 {\scriptstyle \pm 0.36}$ & $3.30 {\scriptstyle \pm 0.40}$ \\
Typed-PC (Maj.) & $2.72 {\scriptstyle \pm 0.31}$ & $2.92 {\scriptstyle \pm 0.33}$ & $3.25 {\scriptstyle \pm 0.35}$ & $2.99 {\scriptstyle \pm 0.32}$ & $2.70 {\scriptstyle \pm 0.37}$ & $2.77 {\scriptstyle \pm 0.33}$ & $2.92 {\scriptstyle \pm 0.31}$ \\
\hline
Tagged-PC (AntiV) & $3.36 {\scriptstyle \pm 0.49}$ & $3.87 {\scriptstyle \pm 0.39}$ & $4.36 {\scriptstyle \pm 0.30}$ & $3.80 {\scriptstyle \pm 0.45}$ & $4.13 {\scriptstyle \pm 0.38}$ & $3.45 {\scriptstyle \pm 0.43}$ & $3.91 {\scriptstyle \pm 0.42}$ \\
Tagged-PC & $3.11 {\scriptstyle \pm 0.38}$ & $3.41 {\scriptstyle \pm 0.33}$ & $3.17 {\scriptstyle \pm 0.33}$ & $3.21 {\scriptstyle \pm 0.53}$ & $3.24 {\scriptstyle \pm 0.52}$ & $3.24 {\scriptstyle \pm 0.29}$ & $3.46 {\scriptstyle \pm 0.25}$ \\
Tagged-GES & $\mathbf{2.10} {\scriptstyle \pm 0.56}$ & $\mathbf{1.85} {\scriptstyle \pm 0.41}$ & $1.55 {\scriptstyle \pm 0.21}$ & $\mathbf{1.82} {\scriptstyle \pm 0.42}$ & $\mathbf{2.58} {\scriptstyle \pm 0.55}$ & $\mathbf{1.84} {\scriptstyle \pm 0.42}$ & $\mathbf{1.82} {\scriptstyle \pm 0.46}$ \\
\end{tabular}
}
\caption{\textbf{Average \underline{Ranks} for Best Configurations over all Datasets (lower is better) using Laplace Correction.} Results for tagging worsen slightly (resulting in marginally better ranks for other methods), but stay very close to the original evaluation results, yielding no qualitatively different results.  ${}_{\pm y}$ indicates standard deviation.}
\label{tab:laplace}
\end{table*}

\subsection{Ablating the Tag Generation Procedure}
\label{sec:appxPromptAblation}
In this section, we investigate how much variation in generating tags influences downstream performance.
To this end, we consider different LLM versions (\ref{sec:llm_versions}) and different prompting strategies (\ref{sec:llm_prompting}).

\subsubsection{Comparison of Different LLM Versions}
\label{sec:llm_different_versions}
We consider three LLMs of our main evaluation, where several different versions are accessible online.
For each LLM family (Claude, GPT, Qwen3), we use the same parameter configuration that was identified as best for it in our main evaluation (see Tab.~\ref{tab:best_f1_configs}).
Using these, we run our main evaluation (see Sec.~\ref{sec:main_eval}) for Tagged-PC and Tagged-GES, with 10 different seeds per configuration (tags remain identical per seed, but different dataset samples are used, introducing variation in the base PC and GES results).
We rank the resulting metrics per LLM family and compare intra-family ranks, showing results for Tagged-PC and Tagged-GES in Tabs.~\ref{tab:llm_versions_pc} and~\ref{tab:llm_versions_ges}.\footnote{Contrary to our main evaluation, we resolve ties between methods by averaging ranks instead of using the minimum rank. This makes it easier to quickly assess how much any metric differs from the mean result at $2.5$.}
In both cases, we see that the LLM version has little impact on the result quality, with most metrics staying very close to the mean rank of $2.5$.
There is not a single LLM version or metric where the average ranks lie outside the $[2,3]$ interval.
Even the minor differences that do exist show no clear preference towards newer or older LLM versions, further indicating that the choice of a specific LLM model in obtaining tags is not of crucial importance.
Overall, we conclude that our method is robust to the choice of LLM, as older and newer versions perform similarly.
\label{sec:llm_versions}
\begin{table}[]
    \centering
\resizebox{\textwidth}{!}{
\begin{tabular}{l|ccccccc}
LLM & SHD & SHD\textsubscript{double} & SID\textsubscript{min} & SID\textsubscript{max} & Precision & Recall & F\textsubscript{1} \\
 & Ranks & Ranks & Ranks & Ranks & Ranks & Ranks & Ranks \\
\hline
Claude Opus 5 & $\mathbf{2.08} {\scriptstyle \pm 0.09}$ & $\mathbf{2.08} {\scriptstyle \pm 0.12}$ & $2.48 {\scriptstyle \pm 0.13}$ & $\mathbf{2.26} {\scriptstyle \pm 0.13}$ & $\mathbf{2.37} {\scriptstyle \pm 0.16}$ & $\mathbf{2.08} {\scriptstyle \pm 0.09}$ & $\mathbf{2.12} {\scriptstyle \pm 0.12}$ \\
Claude Opus 4.8 & $2.89 {\scriptstyle \pm 0.10}$ & $2.85 {\scriptstyle \pm 0.08}$ & $2.55 {\scriptstyle \pm 0.10}$ & $2.76 {\scriptstyle \pm 0.16}$ & $2.64 {\scriptstyle \pm 0.19}$ & $2.89 {\scriptstyle \pm 0.10}$ & $2.88 {\scriptstyle \pm 0.11}$ \\
Claude Opus 4.7 & $2.66 {\scriptstyle \pm 0.15}$ & $2.73 {\scriptstyle \pm 0.19}$ & $2.52 {\scriptstyle \pm 0.20}$ & $2.56 {\scriptstyle \pm 0.16}$ & $2.57 {\scriptstyle \pm 0.20}$ & $2.66 {\scriptstyle \pm 0.15}$ & $2.65 {\scriptstyle \pm 0.18}$ \\
Claude Opus 4.6 & $2.37 {\scriptstyle \pm 0.13}$ & $2.34 {\scriptstyle \pm 0.11}$ & $\mathbf{2.45} {\scriptstyle \pm 0.18}$ & $2.42 {\scriptstyle \pm 0.15}$ & $2.43 {\scriptstyle \pm 0.13}$ & $2.37 {\scriptstyle \pm 0.13}$ & $2.35 {\scriptstyle \pm 0.11}$ \\
\hline
GPT 5.6 Sol & $2.47 {\scriptstyle \pm 0.12}$ & $2.61 {\scriptstyle \pm 0.11}$ & $2.70 {\scriptstyle \pm 0.10}$ & $2.65 {\scriptstyle \pm 0.10}$ & $2.68 {\scriptstyle \pm 0.17}$ & $2.47 {\scriptstyle \pm 0.12}$ & $2.59 {\scriptstyle \pm 0.12}$ \\
GPT 5.5 & $\mathbf{2.38} {\scriptstyle \pm 0.14}$ & $\mathbf{2.39} {\scriptstyle \pm 0.16}$ & $\mathbf{2.34} {\scriptstyle \pm 0.11}$ & $2.54 {\scriptstyle \pm 0.15}$ & $2.39 {\scriptstyle \pm 0.10}$ & $\mathbf{2.38} {\scriptstyle \pm 0.14}$ & $\mathbf{2.37} {\scriptstyle \pm 0.18}$ \\
GPT 5.4 & $2.47 {\scriptstyle \pm 0.16}$ & $2.45 {\scriptstyle \pm 0.15}$ & $2.55 {\scriptstyle \pm 0.08}$ & $2.44 {\scriptstyle \pm 0.16}$ & $2.67 {\scriptstyle \pm 0.12}$ & $2.47 {\scriptstyle \pm 0.16}$ & $2.54 {\scriptstyle \pm 0.17}$ \\
GPT 5.2 & $2.68 {\scriptstyle \pm 0.11}$ & $2.55 {\scriptstyle \pm 0.15}$ & $2.41 {\scriptstyle \pm 0.15}$ & $\mathbf{2.38} {\scriptstyle \pm 0.15}$ & $\mathbf{2.26} {\scriptstyle \pm 0.17}$ & $2.68 {\scriptstyle \pm 0.11}$ & $2.50 {\scriptstyle \pm 0.16}$ \\
\hline
Qwen3 8 Max & $2.55 {\scriptstyle \pm 0.19}$ & $2.56 {\scriptstyle \pm 0.14}$ & $\mathbf{2.29} {\scriptstyle \pm 0.14}$ & $\mathbf{2.42} {\scriptstyle \pm 0.14}$ & $\mathbf{2.35} {\scriptstyle \pm 0.22}$ & $2.55 {\scriptstyle \pm 0.19}$ & $2.51 {\scriptstyle \pm 0.18}$ \\
Qwen3 7 Max & $2.46 {\scriptstyle \pm 0.14}$ & $\mathbf{2.45} {\scriptstyle \pm 0.17}$ & $2.58 {\scriptstyle \pm 0.07}$ & $2.54 {\scriptstyle \pm 0.15}$ & $2.49 {\scriptstyle \pm 0.17}$ & $2.46 {\scriptstyle \pm 0.14}$ & $2.48 {\scriptstyle \pm 0.17}$ \\
Qwen3 6 Max Preview & $\mathbf{2.45} {\scriptstyle \pm 0.19}$ & $2.47 {\scriptstyle \pm 0.20}$ & $2.71 {\scriptstyle \pm 0.14}$ & $2.60 {\scriptstyle \pm 0.20}$ & $2.67 {\scriptstyle \pm 0.14}$ & $\mathbf{2.45} {\scriptstyle \pm 0.19}$ & $\mathbf{2.47} {\scriptstyle \pm 0.21}$ \\
Qwen3 5 397B & $2.53 {\scriptstyle \pm 0.18}$ & $2.52 {\scriptstyle \pm 0.14}$ & $2.42 {\scriptstyle \pm 0.12}$ & $2.44 {\scriptstyle \pm 0.14}$ & $2.49 {\scriptstyle \pm 0.15}$ & $2.53 {\scriptstyle \pm 0.18}$ & $2.54 {\scriptstyle \pm 0.17}$ \\
\end{tabular}
}
    \caption{\textbf{Comparison of Different LLM Versions for Tagged-PC.} Shown are the ranks between four different LLM versions of the same LLM family. Ranks are based on 10 runs and use averaging in case of ties. It can be seen that not a single LLM version is substantially better or worse than other versions of the same family, as all ranks are close to the average rank of 2.5, never below 2 and never above 3.  ${}_{\pm y}$ indicates standard deviation.}
    \label{tab:llm_versions_pc}
\end{table}

\begin{table}[]
    \centering
\resizebox{\textwidth}{!}{
\begin{tabular}{l|ccccccc}
LLM & SHD & SHD\textsubscript{double} & SID\textsubscript{min} & SID\textsubscript{max} & Precision & Recall & F\textsubscript{1} \\
 & Ranks & Ranks & Ranks & Ranks & Ranks & Ranks & Ranks \\
\hline
Claude Opus 5 & $\mathbf{2.20} {\scriptstyle \pm 0.07}$ & $2.38 {\scriptstyle \pm 0.08}$ & $2.78 {\scriptstyle \pm 0.05}$ & $\mathbf{2.36} {\scriptstyle \pm 0.06}$ & $2.52 {\scriptstyle \pm 0.08}$ & $\mathbf{2.20} {\scriptstyle \pm 0.07}$ & $2.32 {\scriptstyle \pm 0.08}$ \\
Claude Opus 4.8 & $2.62 {\scriptstyle \pm 0.13}$ & $2.65 {\scriptstyle \pm 0.11}$ & $2.69 {\scriptstyle \pm 0.17}$ & $2.48 {\scriptstyle \pm 0.15}$ & $2.75 {\scriptstyle \pm 0.12}$ & $2.62 {\scriptstyle \pm 0.13}$ & $2.65 {\scriptstyle \pm 0.11}$ \\
Claude Opus 4.7 & $2.81 {\scriptstyle \pm 0.13}$ & $2.77 {\scriptstyle \pm 0.13}$ & $2.34 {\scriptstyle \pm 0.08}$ & $2.66 {\scriptstyle \pm 0.09}$ & $2.55 {\scriptstyle \pm 0.13}$ & $2.81 {\scriptstyle \pm 0.13}$ & $2.84 {\scriptstyle \pm 0.12}$ \\
Claude Opus 4.6 & $2.37 {\scriptstyle \pm 0.09}$ & $\mathbf{2.20} {\scriptstyle \pm 0.08}$ & $\mathbf{2.19} {\scriptstyle \pm 0.06}$ & $2.51 {\scriptstyle \pm 0.07}$ & $\mathbf{2.17} {\scriptstyle \pm 0.13}$ & $2.37 {\scriptstyle \pm 0.09}$ & $\mathbf{2.20} {\scriptstyle \pm 0.08}$ \\
\hline
GPT 5.6 Sol & $2.49 {\scriptstyle \pm 0.10}$ & $2.47 {\scriptstyle \pm 0.12}$ & $2.56 {\scriptstyle \pm 0.10}$ & $2.36 {\scriptstyle \pm 0.11}$ & $2.44 {\scriptstyle \pm 0.13}$ & $2.49 {\scriptstyle \pm 0.10}$ & $2.50 {\scriptstyle \pm 0.13}$ \\
GPT 5.5 & $2.53 {\scriptstyle \pm 0.10}$ & $2.50 {\scriptstyle \pm 0.07}$ & $\mathbf{2.39} {\scriptstyle \pm 0.09}$ & $2.79 {\scriptstyle \pm 0.08}$ & $\mathbf{2.36} {\scriptstyle \pm 0.13}$ & $2.53 {\scriptstyle \pm 0.10}$ & $2.54 {\scriptstyle \pm 0.08}$ \\
GPT 5.4 & $\mathbf{2.03} {\scriptstyle \pm 0.06}$ & $\mathbf{2.26} {\scriptstyle \pm 0.10}$ & $2.56 {\scriptstyle \pm 0.14}$ & $\mathbf{2.04} {\scriptstyle \pm 0.07}$ & $2.64 {\scriptstyle \pm 0.14}$ & $\mathbf{2.03} {\scriptstyle \pm 0.06}$ & $\mathbf{2.12} {\scriptstyle \pm 0.15}$ \\
GPT 5.2 & $2.95 {\scriptstyle \pm 0.07}$ & $2.77 {\scriptstyle \pm 0.09}$ & $2.48 {\scriptstyle \pm 0.09}$ & $2.81 {\scriptstyle \pm 0.13}$ & $2.56 {\scriptstyle \pm 0.13}$ & $2.95 {\scriptstyle \pm 0.07}$ & $2.84 {\scriptstyle \pm 0.08}$ \\
\hline
Qwen3 8 Max & $2.58 {\scriptstyle \pm 0.12}$ & $2.37 {\scriptstyle \pm 0.17}$ & $\mathbf{2.21} {\scriptstyle \pm 0.08}$ & $2.56 {\scriptstyle \pm 0.13}$ & $\mathbf{2.35} {\scriptstyle \pm 0.10}$ & $2.58 {\scriptstyle \pm 0.12}$ & $2.39 {\scriptstyle \pm 0.17}$ \\
Qwen3 7 Max & $2.46 {\scriptstyle \pm 0.15}$ & $2.58 {\scriptstyle \pm 0.18}$ & $2.82 {\scriptstyle \pm 0.05}$ & $2.56 {\scriptstyle \pm 0.05}$ & $2.73 {\scriptstyle \pm 0.16}$ & $2.46 {\scriptstyle \pm 0.15}$ & $2.55 {\scriptstyle \pm 0.20}$ \\
Qwen3 6 Max Preview & $2.84 {\scriptstyle \pm 0.10}$ & $2.82 {\scriptstyle \pm 0.10}$ & $2.46 {\scriptstyle \pm 0.15}$ & $2.85 {\scriptstyle \pm 0.11}$ & $2.54 {\scriptstyle \pm 0.10}$ & $2.84 {\scriptstyle \pm 0.10}$ & $2.83 {\scriptstyle \pm 0.11}$ \\
Qwen3 5 397B & $\mathbf{2.12} {\scriptstyle \pm 0.11}$ & $\mathbf{2.23} {\scriptstyle \pm 0.12}$ & $2.52 {\scriptstyle \pm 0.10}$ & $\mathbf{2.04} {\scriptstyle \pm 0.13}$ & $2.38 {\scriptstyle \pm 0.07}$ & $\mathbf{2.12} {\scriptstyle \pm 0.11}$ & $\mathbf{2.23} {\scriptstyle \pm 0.12}$ \\
\end{tabular}
}
    \caption{\textbf{Comparison of Different LLM Versions for Tagged-GES.} Shown are the ranks between four different LLM versions of the same LLM family. Ranks are based on 10 runs and use averaging in case of ties. It can be seen that not a single LLM version is substantially better or worse than other versions of the same family, as all ranks are close to the average rank of 2.5, never below 2 and never above 3.  ${}_{\pm y}$ indicates standard deviation.}
    \label{tab:llm_versions_ges}
\end{table}

\subsubsection{Different Prompting Strategies}
\label{sec:llm_prompting}
We now evaluate the impact of varying the prompting strategies. In particular, we (1) identify 7 optional components within the prompt (stating the definition of a tag, explaining the desired tag specificity tradeoff, describing the 'no duplicate' requirement, \dots) and query LLMs under leave-one-out schema; (2) we paraphrase the prompt in a bullet point list of standard (matching the original prompt), 'terse' and 'verbose' length, and (3) analyze different prompt politeness levels of 'polite', 'demanding' and 'imperative'. Similar to the previous prompt variations experiments that alter the content and style of the prompt in order to analyze changes in the quality of the produced tags (and possible effects on downstream performance), we (4) also consider effects on the number of produced tags. Instead of letting the LLM emit the number of tags it finds best fitting, we alter the prompt to demand a specific number of tags ($\texttt{numTags} \in \{2, 3, 4, 6, 8, 10\}$) per variable. All prompt details are given in App.~\ref{sec:prompt_variations}.
This experiment is based on our main evaluation (Sec.~\ref{sec:main_eval}).
As our main evaluation covers a large variety of parameter configurations, pairing all of these with every prompting strategy is expensive in terms of both prompting for all LLMs and running the full parameter grid.
Therefore, we restrict ourselves to the best-performing parameter configuration involving Claude for Tagged-GES.
Tab.~\ref{tab:different_tagging_prompts} shows the results in the same style as Tab.~\ref{tab:avgBestRankedResults}.
Importantly, to simplify comparing the original and the new prompting strategies, we do not rank the new prompting strategies among themselves, but place any of them as a substitute for the original prompt, and rank this setup again.
In other words, we show what rank this configuration would have achieved if used instead of our default prompt.
The results show that, while our prompting strategy performs best, the other prompting strategies still perform well, remaining to be ranked similarly to the default prompt.
Importantly, for F\textsubscript{1}, a metric that takes both prediction accuracy and abstention rate into account, the rank for any prompting strategy is smaller than that of the other methods recorded in our original evaluation.
\begin{table}[]
    \centering
\resizebox{\textwidth}{!}{
\begin{tabular}{l|ccccccc}
Method & SHD & SHD\textsubscript{double} & SID\textsubscript{min} & SID\textsubscript{max} & Precision & Recall & F\textsubscript{1} \\
 & Ranks & Ranks & Ranks & Ranks & Ranks & Ranks & Ranks \\
\hline
PC$^{\textbf{(ref)}}$ & $4.32 {\scriptstyle \pm 0.38}$ & $4.27 {\scriptstyle \pm 0.33}$ & $3.15 {\scriptstyle \pm 0.28}$ & $3.98 {\scriptstyle \pm 0.48}$ & $3.18 {\scriptstyle \pm 0.39}$ & $4.41 {\scriptstyle \pm 0.34}$ & $4.45 {\scriptstyle \pm 0.31}$ \\
GES$^{\textbf{(ref)}}$ & $3.54 {\scriptstyle \pm 0.51}$ & $3.07 {\scriptstyle \pm 0.47}$ & $1.27 {\scriptstyle \pm 0.13}$ & $3.17 {\scriptstyle \pm 0.51}$ & $2.69 {\scriptstyle \pm 0.51}$ & $3.17 {\scriptstyle \pm 0.42}$ & $3.13 {\scriptstyle \pm 0.51}$ \\
Typed-PC (Naive)$^{\textbf{(ref)}}$ & $3.42 {\scriptstyle \pm 0.34}$ & $3.62 {\scriptstyle \pm 0.33}$ & $3.10 {\scriptstyle \pm 0.27}$ & $3.20 {\scriptstyle \pm 0.33}$ & $3.07 {\scriptstyle \pm 0.42}$ & $3.47 {\scriptstyle \pm 0.34}$ & $3.59 {\scriptstyle \pm 0.39}$ \\
Typed-PC (Maj.)$^{\textbf{(ref)}}$ & $2.97 {\scriptstyle \pm 0.28}$ & $3.25 {\scriptstyle \pm 0.34}$ & $3.40 {\scriptstyle \pm 0.38}$ & $3.27 {\scriptstyle \pm 0.37}$ & $2.84 {\scriptstyle \pm 0.35}$ & $3.05 {\scriptstyle \pm 0.26}$ & $3.20 {\scriptstyle \pm 0.24}$ \\
Tagged-PC (AntiV)$^{\textbf{(ref)}}$ & $3.60 {\scriptstyle \pm 0.45}$ & $4.05 {\scriptstyle \pm 0.38}$ & $4.54 {\scriptstyle \pm 0.32}$ & $4.05 {\scriptstyle \pm 0.46}$ & $4.25 {\scriptstyle \pm 0.40}$ & $3.67 {\scriptstyle \pm 0.37}$ & $4.11 {\scriptstyle \pm 0.40}$ \\
Tagged-PC$^{\textbf{(ref)}}$ & $2.66 {\scriptstyle \pm 0.32}$ & $3.00 {\scriptstyle \pm 0.41}$ & $2.95 {\scriptstyle \pm 0.49}$ & $2.67 {\scriptstyle \pm 0.50}$ & $3.07 {\scriptstyle \pm 0.48}$ & $2.89 {\scriptstyle \pm 0.28}$ & $2.99 {\scriptstyle \pm 0.38}$ \\
\hline
\hline
Tagged-GES (Baseline (Default Prompt)) & $\mathbf{2.11} {\scriptstyle \pm 0.55}$ & $\mathbf{1.73} {\scriptstyle \pm 0.39}$ & $\mathbf{1.11} {\scriptstyle \pm 0.09}$ & $\mathbf{1.68} {\scriptstyle \pm 0.37}$ & $\mathbf{2.24} {\scriptstyle \pm 0.51}$ & $\mathbf{1.77} {\scriptstyle \pm 0.42}$ & $\mathbf{1.70} {\scriptstyle \pm 0.47}$ \\
\hline
\hline
Tagged-GES (Tag Definition) & $2.39 {\scriptstyle \pm 0.56}$ & $2.07 {\scriptstyle \pm 0.40}$ & $1.80 {\scriptstyle \pm 0.29}$ & $2.45 {\scriptstyle \pm 0.45}$ & $2.81 {\scriptstyle \pm 0.52}$ & $2.05 {\scriptstyle \pm 0.42}$ & $2.12 {\scriptstyle \pm 0.43}$ \\
Tagged-GES (Specificity Tradeoff) & $2.39 {\scriptstyle \pm 0.53}$ & $2.03 {\scriptstyle \pm 0.42}$ & $1.34 {\scriptstyle \pm 0.18}$ & $2.14 {\scriptstyle \pm 0.41}$ & $2.40 {\scriptstyle \pm 0.53}$ & $2.06 {\scriptstyle \pm 0.40}$ & $2.04 {\scriptstyle \pm 0.48}$ \\
Tagged-GES (Downstream Purpose) & $2.69 {\scriptstyle \pm 0.62}$ & $2.30 {\scriptstyle \pm 0.55}$ & $1.59 {\scriptstyle \pm 0.24}$ & $1.91 {\scriptstyle \pm 0.43}$ & $2.82 {\scriptstyle \pm 0.46}$ & $2.19 {\scriptstyle \pm 0.47}$ & $2.32 {\scriptstyle \pm 0.56}$ \\
Tagged-GES (Discriminativeness) & $2.61 {\scriptstyle \pm 0.62}$ & $2.61 {\scriptstyle \pm 0.33}$ & $1.85 {\scriptstyle \pm 0.34}$ & $2.56 {\scriptstyle \pm 0.52}$ & $2.97 {\scriptstyle \pm 0.43}$ & $2.18 {\scriptstyle \pm 0.46}$ & $2.53 {\scriptstyle \pm 0.37}$ \\
Tagged-GES (Multiple Tags) & $2.37 {\scriptstyle \pm 0.61}$ & $2.36 {\scriptstyle \pm 0.49}$ & $2.01 {\scriptstyle \pm 0.21}$ & $2.06 {\scriptstyle \pm 0.52}$ & $3.00 {\scriptstyle \pm 0.53}$ & $2.05 {\scriptstyle \pm 0.45}$ & $2.29 {\scriptstyle \pm 0.53}$ \\
Tagged-GES (Tag Balance) & $2.52 {\scriptstyle \pm 0.56}$ & $2.20 {\scriptstyle \pm 0.41}$ & $1.71 {\scriptstyle \pm 0.33}$ & $2.52 {\scriptstyle \pm 0.42}$ & $2.80 {\scriptstyle \pm 0.49}$ & $2.15 {\scriptstyle \pm 0.42}$ & $2.26 {\scriptstyle \pm 0.45}$ \\
Tagged-GES (No Duplicated) & $2.55 {\scriptstyle \pm 0.60}$ & $2.27 {\scriptstyle \pm 0.49}$ & $1.52 {\scriptstyle \pm 0.23}$ & $1.99 {\scriptstyle \pm 0.48}$ & $2.73 {\scriptstyle \pm 0.56}$ & $2.10 {\scriptstyle \pm 0.43}$ & $2.24 {\scriptstyle \pm 0.54}$ \\
\hline
Tagged-GES (Bullets) & $2.47 {\scriptstyle \pm 0.61}$ & $2.27 {\scriptstyle \pm 0.45}$ & $1.89 {\scriptstyle \pm 0.31}$ & $2.40 {\scriptstyle \pm 0.47}$ & $2.98 {\scriptstyle \pm 0.45}$ & $2.06 {\scriptstyle \pm 0.45}$ & $2.15 {\scriptstyle \pm 0.44}$ \\
Tagged-GES (Terse) & $2.67 {\scriptstyle \pm 0.59}$ & $2.60 {\scriptstyle \pm 0.39}$ & $1.90 {\scriptstyle \pm 0.33}$ & $2.62 {\scriptstyle \pm 0.47}$ & $3.03 {\scriptstyle \pm 0.45}$ & $2.31 {\scriptstyle \pm 0.45}$ & $2.55 {\scriptstyle \pm 0.39}$ \\
Tagged-GES (Verbose) & $2.54 {\scriptstyle \pm 0.65}$ & $2.45 {\scriptstyle \pm 0.44}$ & $2.35 {\scriptstyle \pm 0.31}$ & $2.41 {\scriptstyle \pm 0.52}$ & $3.33 {\scriptstyle \pm 0.48}$ & $2.15 {\scriptstyle \pm 0.48}$ & $2.42 {\scriptstyle \pm 0.49}$ \\
\hline
Tagged-GES (Polite) & $2.54 {\scriptstyle \pm 0.59}$ & $2.22 {\scriptstyle \pm 0.46}$ & $1.24 {\scriptstyle \pm 0.17}$ & $1.99 {\scriptstyle \pm 0.50}$ & $2.45 {\scriptstyle \pm 0.52}$ & $2.12 {\scriptstyle \pm 0.44}$ & $2.20 {\scriptstyle \pm 0.46}$ \\
Tagged-GES (Demanding) & $2.35 {\scriptstyle \pm 0.56}$ & $2.08 {\scriptstyle \pm 0.43}$ & $1.73 {\scriptstyle \pm 0.18}$ & $2.10 {\scriptstyle \pm 0.48}$ & $2.66 {\scriptstyle \pm 0.56}$ & $2.03 {\scriptstyle \pm 0.43}$ & $2.06 {\scriptstyle \pm 0.49}$ \\
Tagged-GES (Imperative) & $2.65 {\scriptstyle \pm 0.54}$ & $2.40 {\scriptstyle \pm 0.47}$ & $1.29 {\scriptstyle \pm 0.17}$ & $2.34 {\scriptstyle \pm 0.47}$ & $2.86 {\scriptstyle \pm 0.51}$ & $2.18 {\scriptstyle \pm 0.40}$ & $2.36 {\scriptstyle \pm 0.47}$ \\
\hline
Tagged-GES (Budget = 2) & $2.30 {\scriptstyle \pm 0.62}$ & $2.10 {\scriptstyle \pm 0.47}$ & $1.88 {\scriptstyle \pm 0.17}$ & $2.02 {\scriptstyle \pm 0.45}$ & $2.77 {\scriptstyle \pm 0.53}$ & $1.88 {\scriptstyle \pm 0.47}$ & $2.05 {\scriptstyle \pm 0.48}$ \\
Tagged-GES (Budget = 3) & $2.49 {\scriptstyle \pm 0.64}$ & $2.43 {\scriptstyle \pm 0.43}$ & $2.66 {\scriptstyle \pm 0.36}$ & $2.62 {\scriptstyle \pm 0.49}$ & $3.55 {\scriptstyle \pm 0.44}$ & $2.13 {\scriptstyle \pm 0.48}$ & $2.45 {\scriptstyle \pm 0.42}$ \\
Tagged-GES (Budget = 4) & $2.19 {\scriptstyle \pm 0.55}$ & $2.21 {\scriptstyle \pm 0.37}$ & $2.08 {\scriptstyle \pm 0.14}$ & $2.02 {\scriptstyle \pm 0.45}$ & $3.25 {\scriptstyle \pm 0.52}$ & $1.86 {\scriptstyle \pm 0.42}$ & $2.19 {\scriptstyle \pm 0.47}$ \\
Tagged-GES (Budget = 6) & $2.52 {\scriptstyle \pm 0.56}$ & $2.41 {\scriptstyle \pm 0.37}$ & $2.16 {\scriptstyle \pm 0.26}$ & $2.24 {\scriptstyle \pm 0.51}$ & $3.15 {\scriptstyle \pm 0.50}$ & $2.19 {\scriptstyle \pm 0.43}$ & $2.32 {\scriptstyle \pm 0.47}$ \\
Tagged-GES (Budget = 8) & $2.37 {\scriptstyle \pm 0.57}$ & $2.25 {\scriptstyle \pm 0.35}$ & $2.51 {\scriptstyle \pm 0.33}$ & $2.31 {\scriptstyle \pm 0.46}$ & $3.36 {\scriptstyle \pm 0.48}$ & $2.02 {\scriptstyle \pm 0.43}$ & $2.25 {\scriptstyle \pm 0.42}$ \\
Tagged-GES (Budget = 10) & $2.15 {\scriptstyle \pm 0.55}$ & $1.91 {\scriptstyle \pm 0.37}$ & $1.86 {\scriptstyle \pm 0.19}$ & $1.99 {\scriptstyle \pm 0.48}$ & $2.65 {\scriptstyle \pm 0.55}$ & $1.84 {\scriptstyle \pm 0.44}$ & $1.88 {\scriptstyle \pm 0.45}$ \\
\end{tabular}

}
    \caption{\textbf{Comparison of Different Prompting Strategies.} We use the main evaluation (Tab.~\ref{tab:avgBestRankedResults}) as a baseline to rank new results. We reevaluate and rank Tagged-GES under the same best hyperparameter settings, but swapping out LLM instructions according to the different prompt variations described in Sec.~\ref{sec:prompt_variations}. Tagged-GES with its `standard' prompt and is labeled ``Default Prompt''. Rows marked with ${\small\square}^{\textbf{(ref)}}$ are carried over from Tab.~\ref{tab:avgBestRankedResults} and provided as reference. Our chosen default prompt performs best. For most metrics (and always for F\textsubscript{1}), the alternative prompting strategies still surpass the reference ranks.  ${}_{\pm y}$ indicates standard deviation.}
    \label{tab:different_tagging_prompts}
\end{table}

\subsection{Directing of Undirected Edges}
\label{sec:undirected}
We conduct an additional experiment to investigate how our approach manages to direct undirected edges in settings without errors.
To this end, we use the ground-truth graphs and undirect 10\% to 50\% of all edges.
In the same manner as in Sec.~\ref{sec:faults}, we either consider all possible combinations or 20,000 random edge combinations per dataset and number of edges to undirect.
We also disable any application of Meek rules and acyclicity constraints, measuring only the correctness of directing the undirected edges according to the tagging statistics.
We report results in Tab.~\ref{tab:undirect}, containing the number of correct and incorrect predictions, as well as the number of edges that our approach kept as undirected edges.
Every edge direction is counted as one correct/incorrect/undecided count (therefore, the sum of these can exceed 20k).
Overall, tagging predicts edges correctly much more often than not, although there are many cases where the edge remains undirected because of high uncertainty.

\begin{table}[t]
    \centering
\rotatebox{90}{%
\resizebox{1.2\textwidth}{!}{
\begin{tabular}{cl|ccccccccccc}
 & & Ca & Ea & Su & As & Lu & Ch & In & Al & Ha & He & Wi \\
\hline
\hline
\multirow{5}{*}{\rotatebox{90}{GPT 5.2}} & 10\% & 0 / 0 / 0 & 0 / 0 / 0 & 5 / 0 / 1 & 7 / 0 / 1 & 7 / 0 / 5 & 272 / 0 / 328 & 60.9k / 4.1k / 35k & 33.7k / 5.9k / 60.5k & 10.5k / 0 / 129.5k & 62.7k / 0 / 177.3k & 37.5k / 5.7k / 176.9k \\
 & 20\% & 0 / 0 / 4 & 4 / 0 / 0 & 5 / 0 / 1 & 43 / 0 / 13 & 74 / 2 / 56 & 37.7k / 0 / 62.3k & 114k / 8.5k / 77.4k & 55.5k / 9.5k / 115k & 15.8k / 0 / 244.2k & 124.9k / 0 / 375.1k & 74.1k / 13.3k / 352.7k \\
 & 30\% & 0 / 0 / 4 & 4 / 0 / 0 & 16 / 1 / 13 & 43 / 0 / 13 & 992 / 72 / 916 & 48.8k / 0 / 111.2k & 167k / 13.5k / 139.5k & 74.2k / 12.5k / 193.3k & 18.2k / 0 / 381.8k & 172.5k / 0 / 567.5k & 109.3k / 20.6k / 550.1k \\
 & 40\% & 4 / 0 / 8 & 8 / 0 / 4 & 16 / 1 / 13 & 110 / 0 / 58 & 1.8k / 172 / 2k & 51.8k / 0 / 148.2k & 201k / 16.8k / 202.2k & 83k / 13.6k / 263.4k & 17.8k / 0 / 502.2k & 205.2k / 0 / 774.8k & 134.4k / 25.4k / 740.2k \\
 & 50\% & 4 / 0 / 8 & 8 / 0 / 4 & 18 / 5 / 37 & 150 / 0 / 130 & 2.3k / 270 / 3k & 52.4k / 0 / 187.6k & 224.1k / 18.4k / 277.4k & 87.2k / 14k / 358.8k & 15.3k / 0 / 644.7k & 216.5k / 0 / 1023.5k & 148k / 27.3k / 944.7k \\
\hline
\multirow{5}{*}{\rotatebox{90}{Claude 4.6}} & 10\% & 0 / 0 / 0 & 0 / 0 / 0 & 5 / 0 / 1 & 5 / 0 / 3 & 9 / 1 / 2 & 409 / 79 / 112 & 86.1k / 3.3k / 10.7k & 78.6k / 5.3k / 16.1k & 80.6k / 19.3k / 40.1k & 207.7k / 6.9k / 25.4k & 178.7k / 17.6k / 23.7k \\
 & 20\% & 0 / 0 / 4 & 4 / 0 / 0 & 5 / 0 / 1 & 33 / 4 / 19 & 91 / 13 / 28 & 67.6k / 10.8k / 21.6k & 170.7k / 7.4k / 21.9k & 142.3k / 10.7k / 27k & 148.5k / 33.8k / 77.6k & 430.8k / 16.6k / 52.5k & 351.9k / 35.5k / 52.6k \\
 & 30\% & 0 / 0 / 4 & 4 / 0 / 0 & 20 / 0 / 10 & 33 / 4 / 19 & 1.1k / 184 / 686 & 107.1k / 15.5k / 37.4k & 267.6k / 14.1k / 38.3k & 222.1k / 17.3k / 40.6k & 224.2k / 50.2k / 125.6k & 632.2k / 27k / 80.8k & 534.1k / 54.9k / 91k \\
 & 40\% & 0 / 4 / 8 & 9 / 0 / 3 & 20 / 0 / 10 & 88 / 14 / 66 & 2k / 385 / 1.6k & 131.4k / 18.6k / 50k & 342.2k / 21.8k / 56k & 284.9k / 22.4k / 52.7k & 283.9k / 63.5k / 172.6k & 827.8k / 38.8k / 113.4k & 691.8k / 73.7k / 134.5k \\
 & 50\% & 0 / 4 / 8 & 9 / 0 / 3 & 34 / 0 / 26 & 132 / 18 / 130 & 2.5k / 527 / 2.6k & 153.4k / 21.4k / 65.2k & 408.8k / 31.8k / 79.4k & 359.7k / 28.4k / 71.8k & 342.1k / 76.8k / 241.1k & 1027.7k / 52.5k / 159.7k & 832.8k / 93k / 194.2k \\
\hline
\multirow{5}{*}{\rotatebox{90}{Gemini 3 Pro}} & 10\% & 0 / 0 / 0 & 0 / 0 / 0 & 4 / 1 / 1 & 7 / 0 / 1 & 6 / 0 / 6 & 522 / 25 / 53 & 67k / 9.4k / 23.6k & 77.8k / 8.1k / 14.1k & 50.8k / 21.6k / 67.6k & 174.4k / 11.1k / 54.5k & 155.7k / 20.4k / 43.9k \\
 & 20\% & 2 / 0 / 2 & 3 / 0 / 1 & 4 / 1 / 1 & 46 / 1 / 9 & 72 / 12 / 48 & 84.9k / 4.8k / 10.3k & 132.6k / 19.6k / 47.8k & 140k / 14k / 26k & 97.4k / 43.5k / 119.1k & 361k / 17.5k / 121.6k & 302.1k / 40.9k / 97k \\
 & 30\% & 2 / 0 / 2 & 3 / 0 / 1 & 16 / 0 / 14 & 46 / 1 / 9 & 1.0k / 220 / 743 & 129.7k / 9k / 21.3k & 204k / 31.5k / 84.4k & 216.8k / 21.3k / 42k & 150.6k / 68.4k / 181k & 526.5k / 21.2k / 192.3k & 442.8k / 61.6k / 175.6k \\
 & 40\% & 8 / 0 / 4 & 2 / 1 / 9 & 16 / 0 / 14 & 123 / 8 / 37 & 1.9k / 487 / 1.6k & 155.3k / 12.1k / 32.6k & 254.1k / 41.6k / 124.4k & 276.4k / 26.7k / 56.9k & 194.9k / 89.5k / 235.6k & 681.3k / 23.1k / 275.7k & 547.6k / 77.7k / 274.7k \\
 & 50\% & 8 / 0 / 4 & 2 / 1 / 9 & 26 / 2 / 32 & 177 / 22 / 81 & 2.4k / 686 / 2.4k & 176.5k / 15.2k / 48.2k & 291.8k / 51.6k / 176.6k & 345.5k / 33.1k / 81.3k & 239.5k / 112.9k / 307.6k & 834.5k / 22.6k / 382.8k & 620.2k / 88.7k / 411.1k \\
\hline
\multirow{5}{*}{\rotatebox{90}{Llama 3.3}} & 10\% & 0 / 0 / 0 & 0 / 0 / 0 & 5 / 0 / 1 & 5 / 0 / 3 & 3 / 1 / 8 & 298 / 24 / 278 & 41.4k / 14.1k / 44.5k & 33.3k / 5.5k / 61.2k & 23.9k / 3.9k / 112.3k & 141.8k / 8.3k / 89.9k & 103.6k / 9.8k / 106.6k \\
 & 20\% & 0 / 0 / 4 & 4 / 0 / 0 & 5 / 0 / 1 & 26 / 1 / 29 & 33 / 11 / 88 & 43.3k / 4k / 52.6k & 81.9k / 24.3k / 93.8k & 53.8k / 8k / 118.2k & 39.7k / 6.6k / 213.7k & 289.1k / 15.8k / 195.1k & 194.4k / 19.3k / 226.3k \\
 & 30\% & 0 / 0 / 4 & 4 / 0 / 0 & 22 / 0 / 8 & 26 / 1 / 29 & 468 / 140 / 1.4k & 60.8k / 7k / 92.2k & 128.8k / 35.4k / 155.9k & 70.4k / 10.3k / 199.3k & 51.7k / 8.9k / 339.4k & 418.3k / 21.1k / 300.6k & 275k / 27.6k / 377.4k \\
 & 40\% & 0 / 0 / 12 & 12 / 0 / 0 & 22 / 0 / 8 & 66 / 4 / 98 & 882 / 238 / 2.8k & 69.7k / 9k / 121.2k & 162.3k / 43.9k / 213.8k & 76.7k / 12k / 271.3k & 56.1k / 10.2k / 453.7k & 537.7k / 25.1k / 417.2k & 327.6k / 32.9k / 539.5k \\
 & 50\% & 0 / 0 / 12 & 12 / 0 / 0 & 36 / 0 / 24 & 104 / 6 / 170 & 1.1k / 266 / 4.1k & 76k / 10.6k / 153.4k & 187.3k / 51.1k / 281.6k & 76.6k / 13.8k / 369.7k & 54.7k / 11k / 594.2k & 654.1k / 27.2k / 558.7k & 358.1k / 35.4k / 726.5k \\
\hline
\multirow{5}{*}{\rotatebox{90}{Qwen 3.5}} & 10\% & 0 / 0 / 0 & 0 / 0 / 0 & 6 / 0 / 0 & 3 / 0 / 5 & 7 / 2 / 3 & 335 / 57 / 208 & 18.4k / 3.8k / 77.9k & 32.1k / 0 / 67.9k & 24.2k / 9.5k / 106.3k & 170.3k / 10.7k / 59k & 74.3k / 8.5k / 137.2k \\
 & 20\% & 2 / 0 / 2 & 4 / 0 / 0 & 6 / 0 / 0 & 15 / 0 / 41 & 62 / 20 / 50 & 49.5k / 11.4k / 39.1k & 36.5k / 6.7k / 156.7k & 53.5k / 0 / 126.5k & 43.3k / 17.6k / 199.1k & 352.8k / 19.6k / 127.6k & 139.3k / 17.1k / 283.7k \\
 & 30\% & 2 / 0 / 2 & 4 / 0 / 0 & 22 / 0 / 8 & 15 / 0 / 41 & 873 / 280 / 827 & 69.6k / 20.3k / 70.1k & 54k / 9.2k / 256.8k & 73.3k / 0 / 206.7k & 61.7k / 25.4k / 312.9k & 515.2k / 26.1k / 198.8k & 190.9k / 23.2k / 465.9k \\
 & 40\% & 3 / 0 / 9 & 12 / 0 / 0 & 22 / 0 / 8 & 30 / 0 / 138 & 1.6k / 529 / 1.9k & 78.2k / 25.8k / 96k & 62k / 10.6k / 347.4k & 83.3k / 0 / 276.7k & 71.6k / 29.7k / 418.7k & 667.3k / 30.9k / 281.8k & 216.4k / 25.4k / 658.2k \\
 & 50\% & 3 / 0 / 9 & 12 / 0 / 0 & 32 / 0 / 28 & 30 / 0 / 250 & 1.9k / 683 / 3k & 82k / 30.6k / 127.4k & 63.5k / 11.5k / 445k & 86.5k / 0 / 373.5k & 74.4k / 31.1k / 554.5k & 815.7k / 33.2k / 391.1k & 217.7k / 24.6k / 877.7k \\
\hline
\multirow{5}{*}{\rotatebox{90}{GLM 5}} & 10\% & 0 / 0 / 0 & 0 / 0 / 0 & 4 / 0 / 2 & 8 / 0 / 0 & 6 / 0 / 6 & 313 / 0 / 287 & 68.7k / 7.8k / 23.5k & 60.8k / 6.5k / 32.7k & 52.4k / 18.4k / 69.2k & 175.3k / 8.3k / 56.4k & 98.5k / 27.4k / 94k \\
 & 20\% & 0 / 0 / 4 & 0 / 0 / 4 & 4 / 0 / 2 & 50 / 0 / 6 & 71 / 8 / 53 & 52.8k / 49 / 47.2k & 133.6k / 15.6k / 50.8k & 110.9k / 11.2k / 57.9k & 89.3k / 31.1k / 139.6k & 364.3k / 18.6k / 117.1k & 185.4k / 51.5k / 203.1k \\
 & 30\% & 0 / 0 / 4 & 0 / 0 / 4 & 18 / 0 / 12 & 50 / 0 / 6 & 908 / 167 / 905 & 80.6k / 654 / 78.7k & 202.7k / 24k / 93.3k & 171.1k / 17.3k / 91.7k & 119.6k / 42.2k / 238.3k & 537.6k / 27k / 175.4k & 262.6k / 73.1k / 344.3k \\
 & 40\% & 0 / 0 / 12 & 0 / 0 / 12 & 18 / 0 / 12 & 130 / 0 / 38 & 1.7k / 323 / 1.9k & 95.2k / 1.8k / 103k & 251.7k / 30.5k / 137.7k & 215.2k / 22k / 122.8k & 133.3k / 48.5k / 338.2k & 706.5k / 34.2k / 239.3k & 315.6k / 88.2k / 496.3k \\
 & 50\% & 0 / 0 / 12 & 0 / 0 / 12 & 34 / 0 / 26 & 180 / 0 / 100 & 2.2k / 407 / 3k & 105.9k / 3.8k / 130.3k & 290.6k / 36.5k / 193k & 260k / 28k / 172k & 134.7k / 51.8k / 473.5k & 884.2k / 39.9k / 315.9k & 346k / 98.6k / 675.4k \\
\hline
\multirow{5}{*}{\rotatebox{90}{Minimax 2.5}} & 10\% & 0 / 0 / 0 & 0 / 0 / 0 & 4 / 0 / 2 & 6 / 0 / 2 & 3 / 1 / 8 & 132 / 0 / 468 & 6.1k / 1.9k / 92k & 21.5k / 4k / 74.5k & 88.3k / 20.4k / 31.3k & 198.4k / 10.1k / 31.5k & 80.8k / 19.8k / 119.5k \\
 & 20\% & 2 / 0 / 2 & 4 / 0 / 0 & 4 / 0 / 2 & 36 / 0 / 20 & 27 / 13 / 92 & 16.6k / 0 / 83.4k & 9.4k / 3.8k / 186.8k & 33.2k / 6.4k / 140.4k & 161.7k / 40.8k / 57.5k & 404.8k / 20.3k / 74.8k & 153.8k / 42k / 244.1k \\
 & 30\% & 2 / 0 / 2 & 4 / 0 / 0 & 17 / 0 / 13 & 36 / 0 / 20 & 274 / 195 / 1.5k & 19k / 0 / 141k & 11.6k / 5.7k / 302.8k & 41.3k / 8k / 230.7k & 239.9k / 64.5k / 95.6k & 585k / 29.3k / 125.7k & 225.2k / 64.4k / 390.4k \\
 & 40\% & 8 / 0 / 4 & 12 / 0 / 0 & 17 / 0 / 13 & 98 / 4 / 66 & 430 / 364 / 3.2k & 18.2k / 0 / 181.8k & 12.5k / 6.7k / 400.8k & 44.3k / 8.4k / 307.3k & 297.7k / 83.6k / 138.7k & 751.2k / 37.8k / 191.1k & 279k / 83.3k / 537.8k \\
 & 50\% & 8 / 0 / 4 & 12 / 0 / 0 & 29 / 4 / 27 & 145 / 7 / 128 & 447 / 459 / 4.6k & 16.4k / 0 / 223.6k & 12.8k / 7.1k / 500.1k & 44.9k / 7.8k / 407.4k & 352.1k / 103.8k / 204.1k & 913.2k / 46.5k / 280.3k & 314k / 98.2k / 707.9k \\
\end{tabular}
}%
}
    \caption{\textbf{Predictions on Undirected Edges.} In this experiment, $\{10\%, \dots, 50\%\}$ edges of the ground truth graph were undirected. The evaluation only tests the tagging quality, disabling any application of Meek rules or acyclicity constraints. All results are shown as \textit{correct predictions} / \textit{incorrect predictions} / \textit{no prediction}, where \textit{no prediction} indicates that the tagging algorithm kept the edge undirected.}
    \label{tab:undirect}
\end{table}

\subsection{Further Results on Graph Faults}
\label{sec:all_faults}
We report the plots on removed and inverted edges in Fig.~\ref{fig:ablations_all}.
All results were obtained in the same manner as in Sec.~\ref{sec:faults} of the main body.
While performance varies slightly between the LLMs that generated the tags, we observe no qualitative difference to our reported results on Claude 4.6.
Note that some outlier behavior can be observed for Cancer and Earthquake.
As these datasets are very small (4 edges each), our tagging statistics are highly susceptible to noise, and edges can be directed incorrectly due to just a small amount of bad evidence.

\begin{figure}[h]
    \centering
    \begin{subfigure}{0.4\textwidth}
        \centering
        \includegraphics[width=\textwidth]{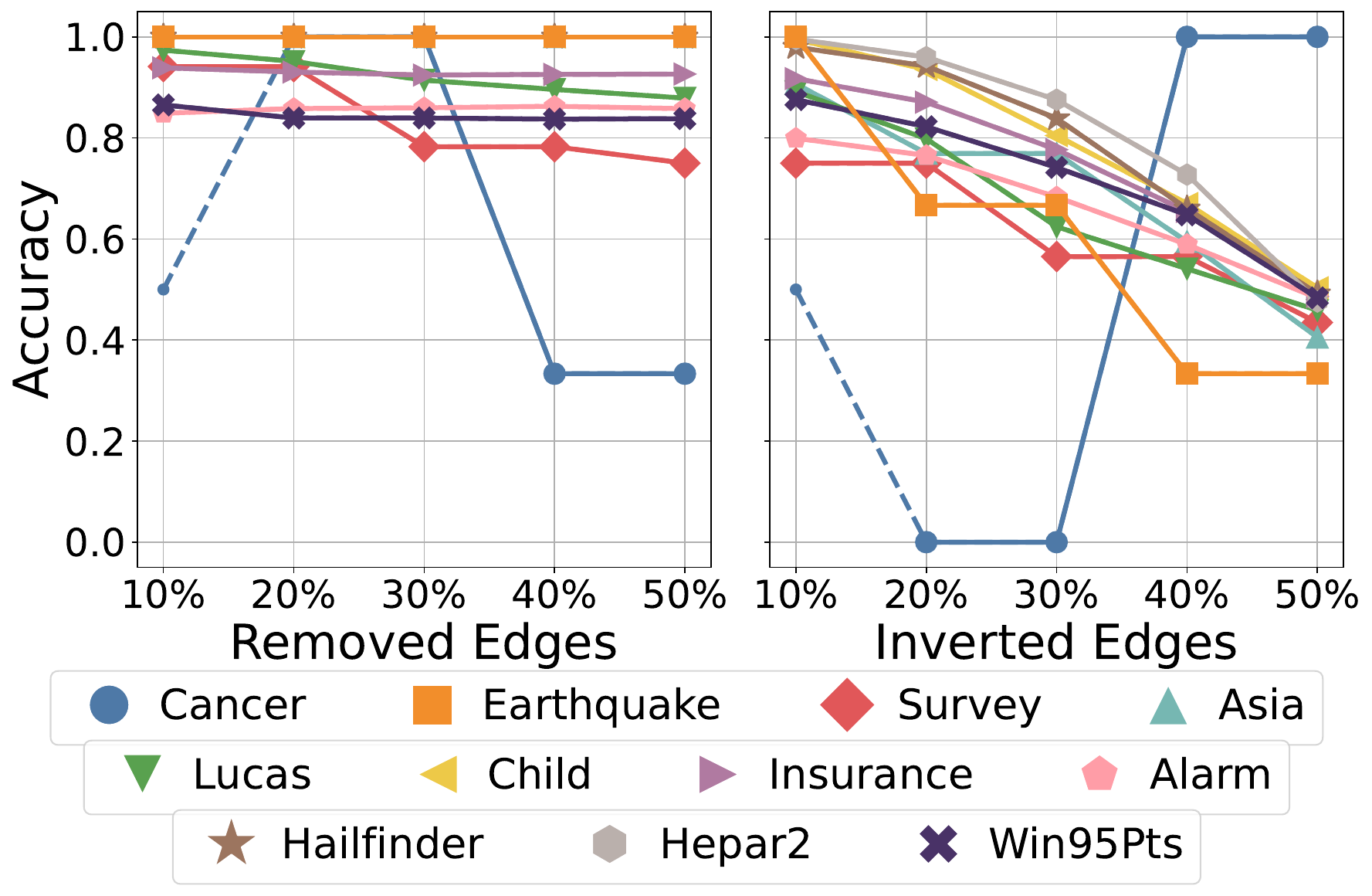}
        \caption{\textbf{GPT 5.2}}
        \label{fig:ablation_sub1}
    \end{subfigure}
    \hfill
    \begin{subfigure}{0.4\textwidth}
        \centering
        \includegraphics[width=\textwidth]{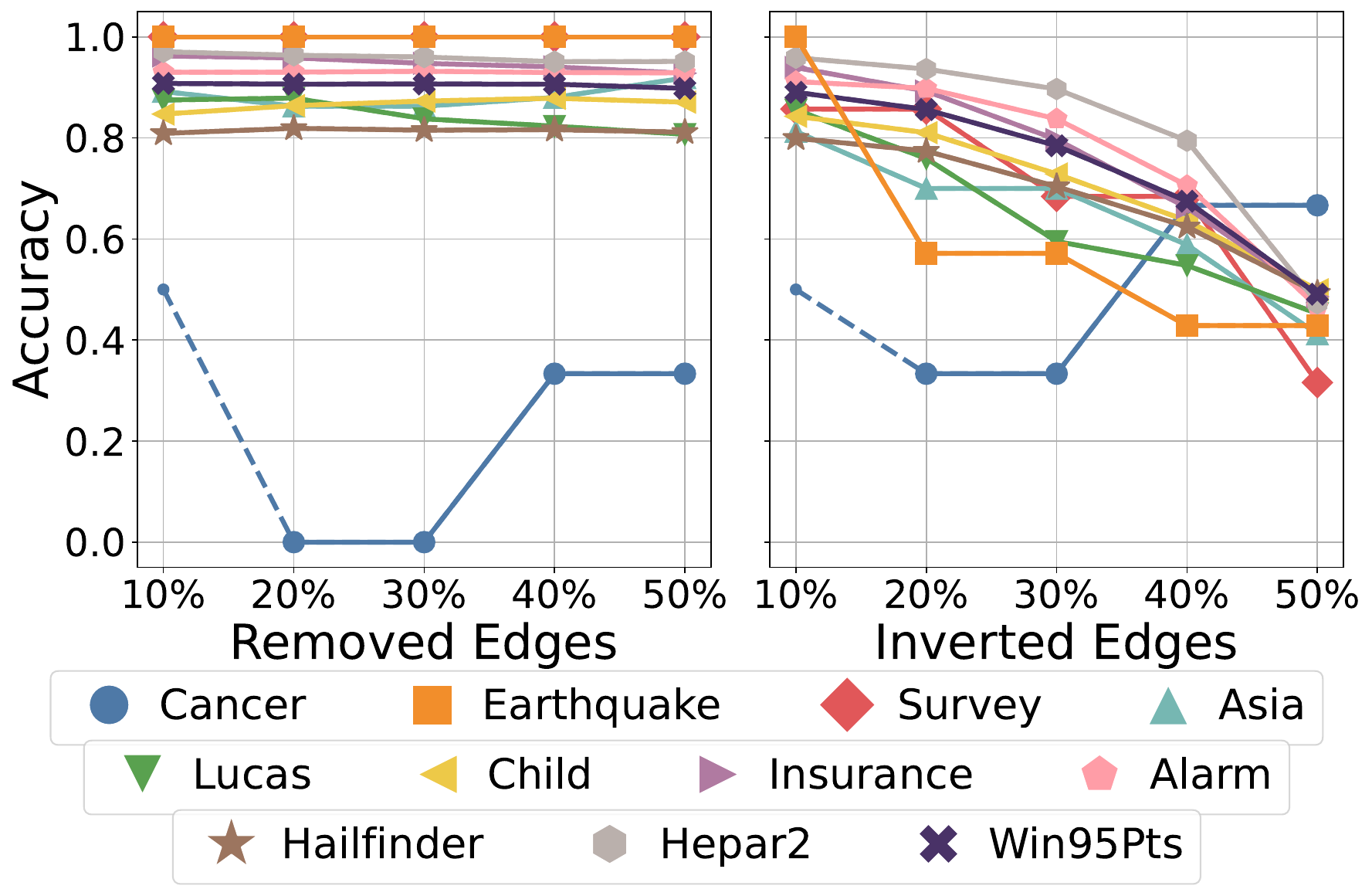}
        \caption{\textbf{Claude 4.6}}
        \label{fig:ablation_sub2}
    \end{subfigure}

    \vspace{0.3cm} %
    
    \begin{subfigure}{0.4\textwidth}
        \centering
        \includegraphics[width=\textwidth]{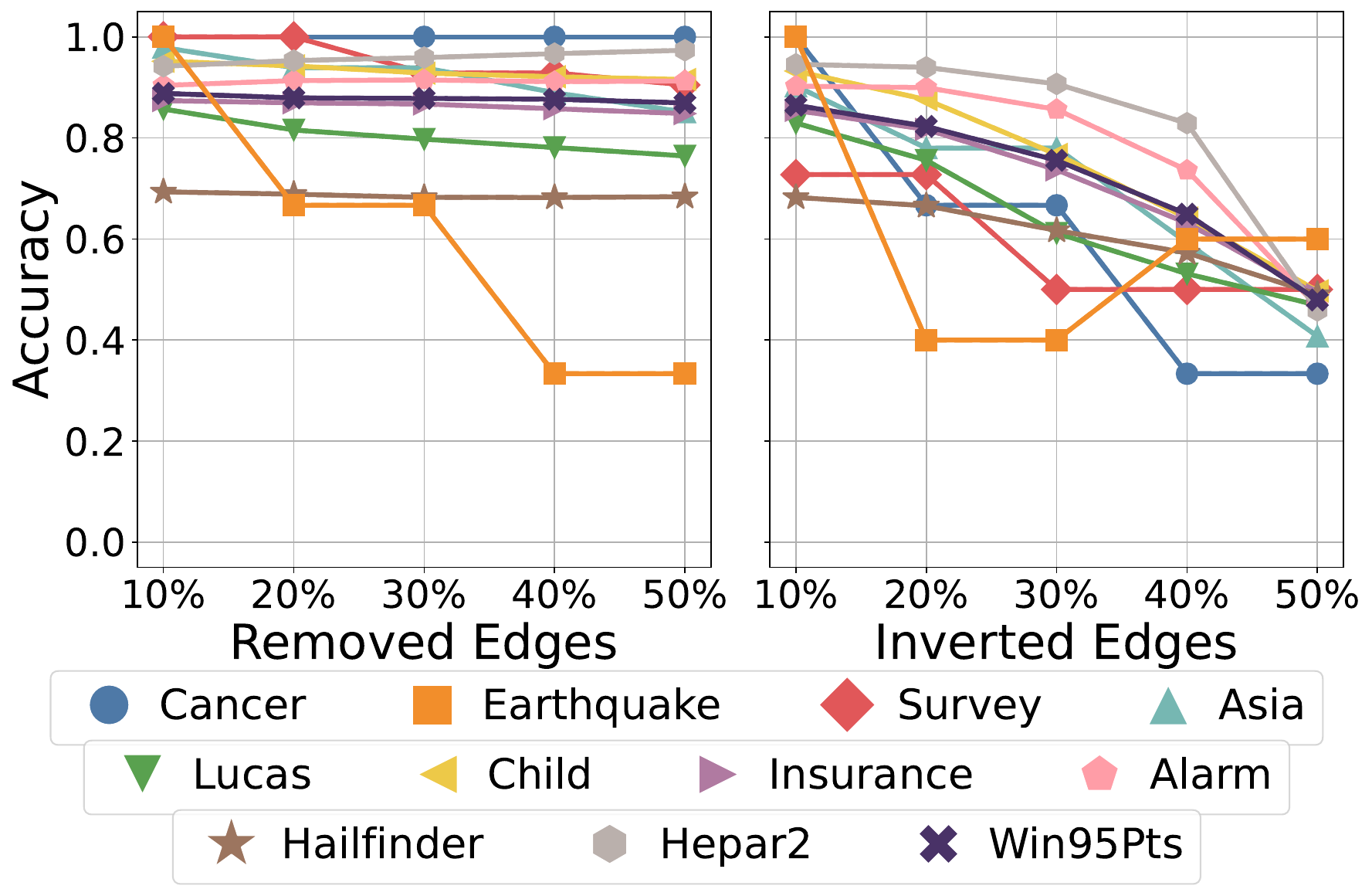}
        \caption{\textbf{Gemini 3 Pro}}
        \label{fig:ablation_sub3}
    \end{subfigure}
    \hfill
    \begin{subfigure}{0.4\textwidth}
        \centering
        \includegraphics[width=\textwidth]{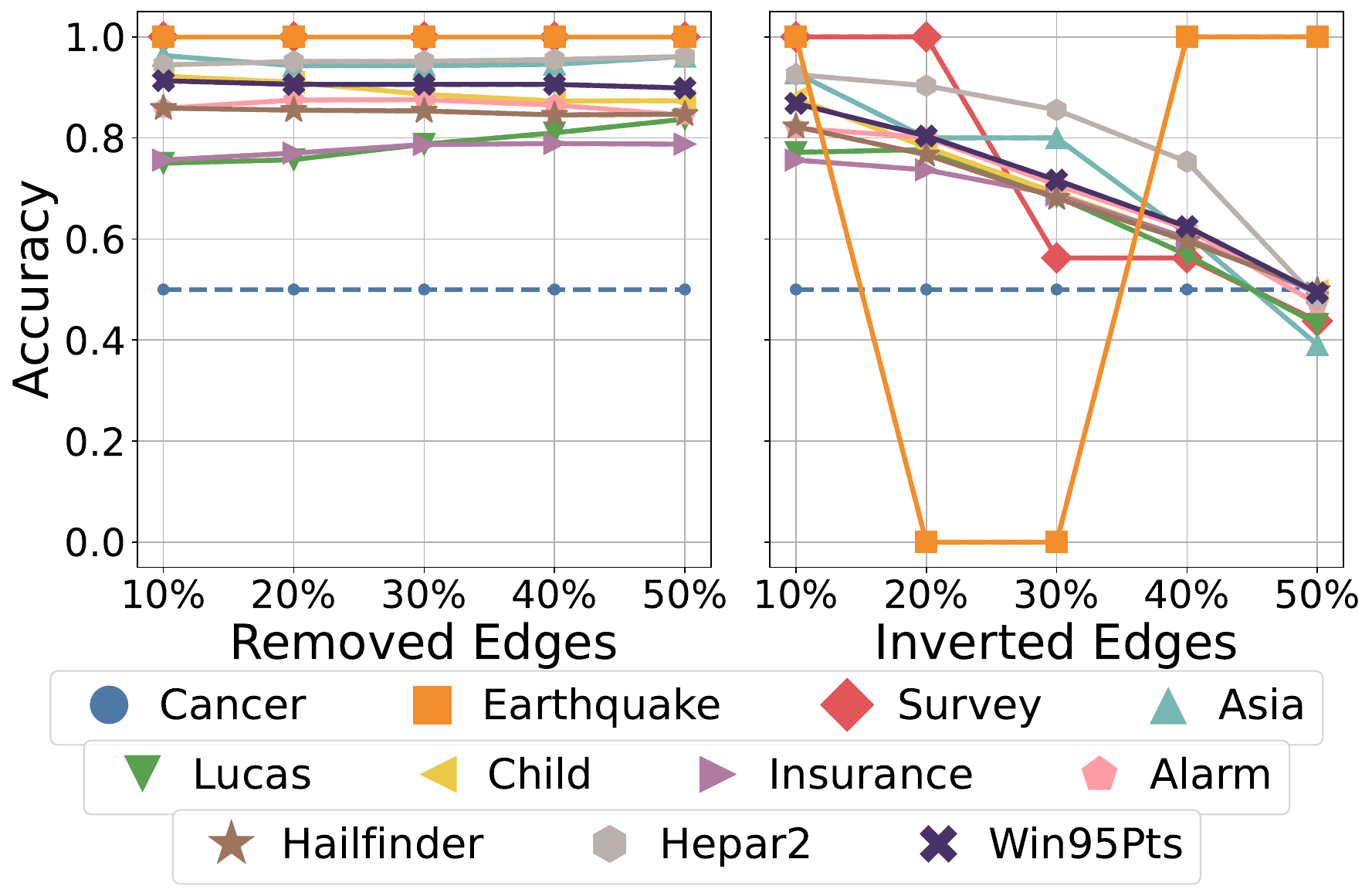}
        \caption{\textbf{Llama 3.3}}
        \label{fig:ablation_sub4}
    \end{subfigure}

    \vspace{0.3cm} %
    
    \begin{subfigure}{0.4\textwidth}
        \centering
        \includegraphics[width=\textwidth]{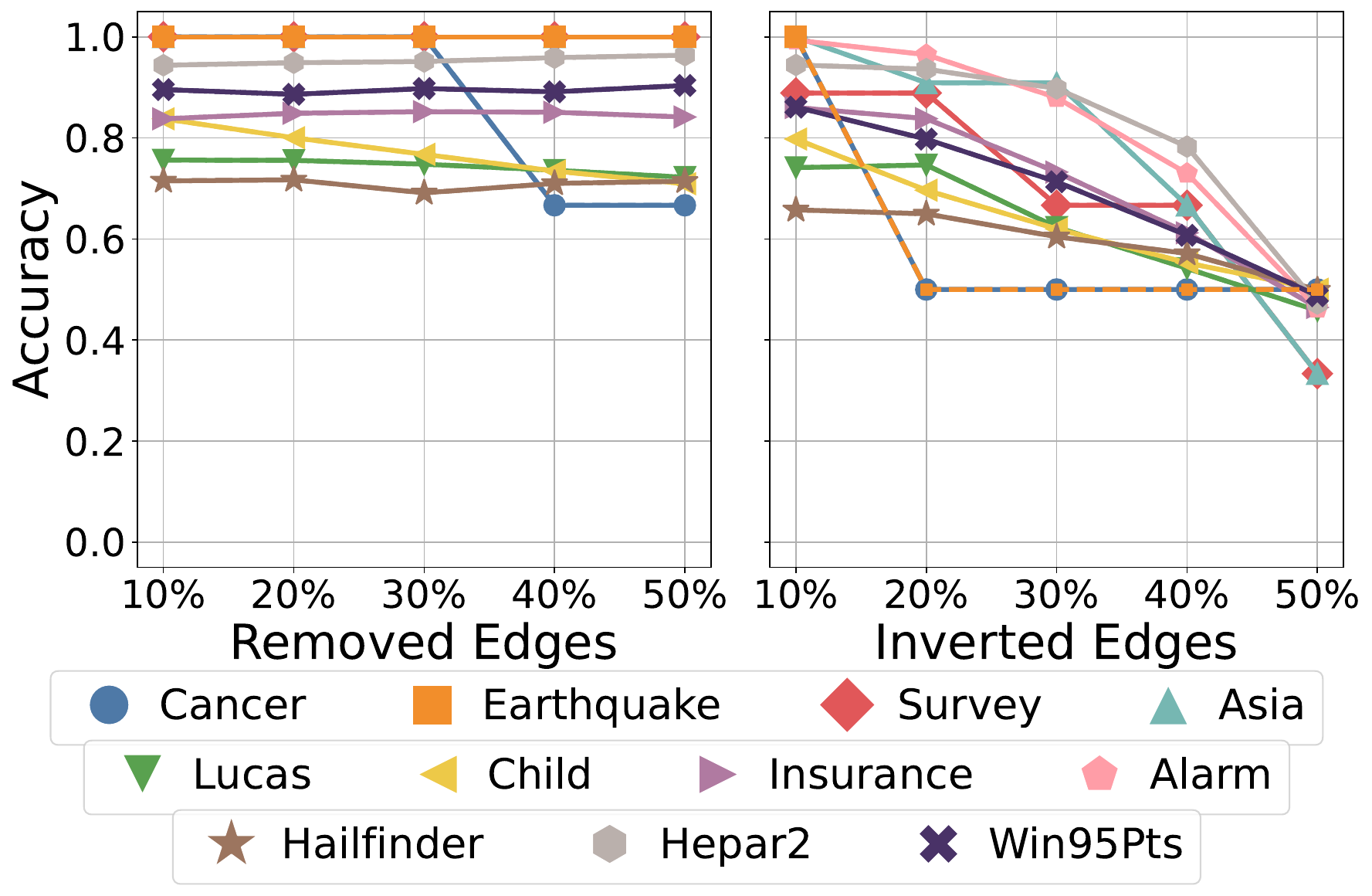}
        \caption{\textbf{Qwen 3.5}}
        \label{fig:ablation_sub5}
    \end{subfigure}
    \hfill
    \begin{subfigure}{0.4\textwidth}
        \centering
        \includegraphics[width=\textwidth]{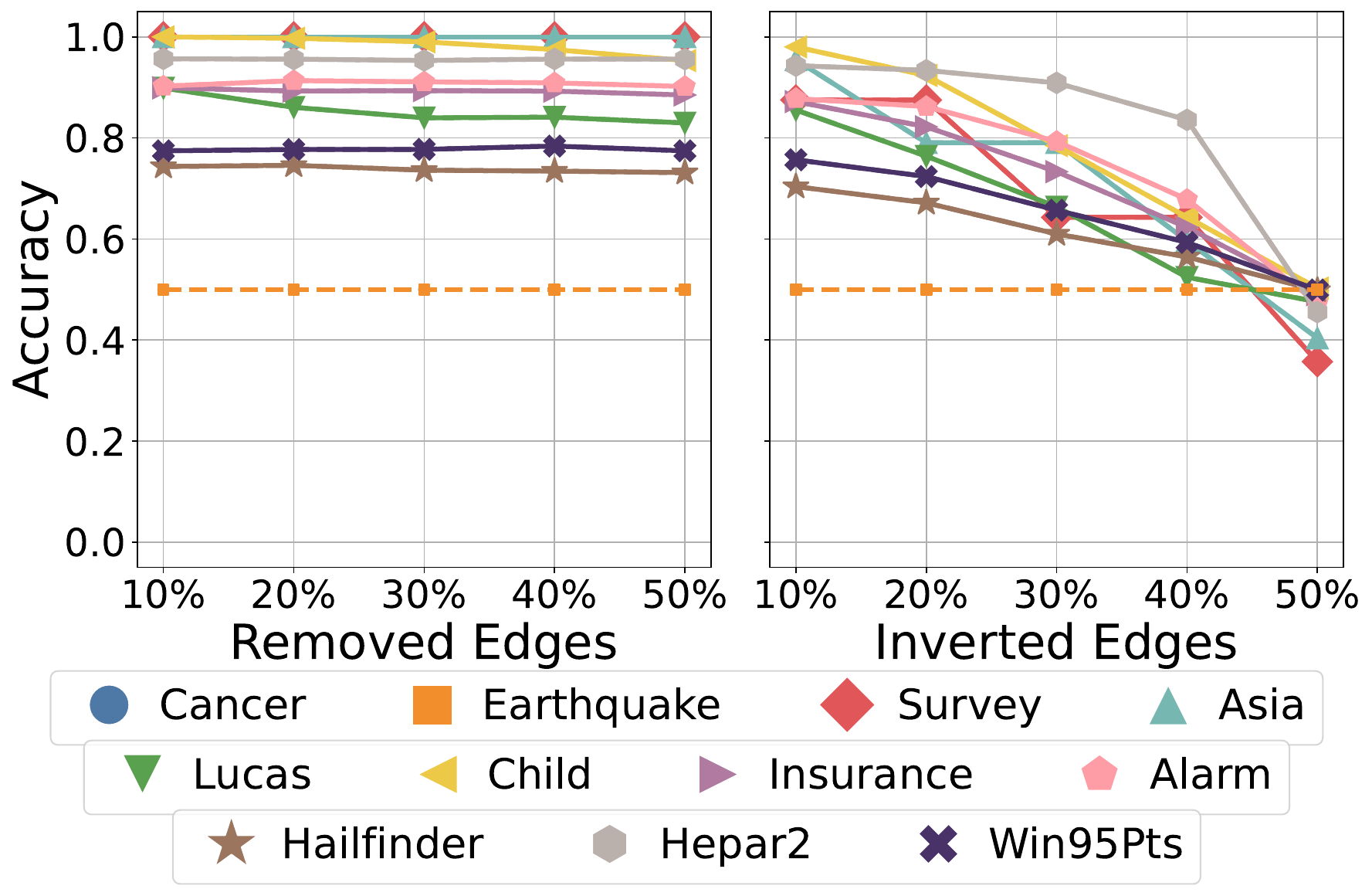}
        \caption{\textbf{GLM 5}}
        \label{fig:ablation_sub6}
    \end{subfigure}

    \vspace{0.3cm} %
    
    \begin{subfigure}{0.4\textwidth}
        \centering
        \includegraphics[width=\textwidth]{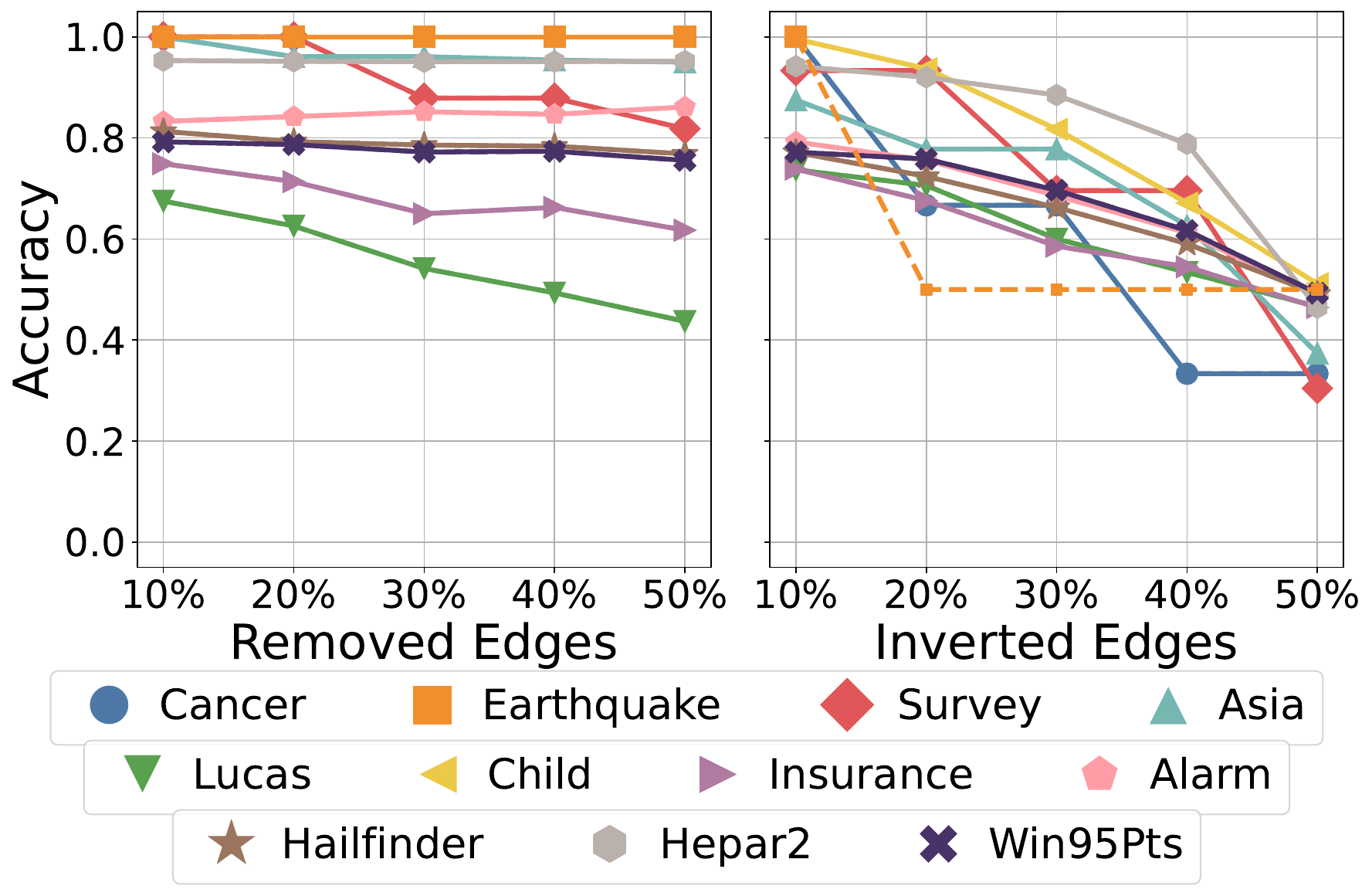}
        \caption{\textbf{Minimax 2.5}}
        \label{fig:ablation_sub7}
    \end{subfigure}
    
    \caption{\textbf{Accuracy with Errors in the Graph.} Tags were generated by the respective LLM. We can see the same general patterns as for Claude 4.6 for the other LLMs, where removed edges decrease prediction accuracy slightly, while adding errors impacts performance more strongly. (We set accuracy to 50\% for datasets where no edge was directed; shown by a smaller point and a dotted line.)}
    \label{fig:ablations_all}
\end{figure}

\subsection{Edge Directions under Noisy or Sparse Tags}
\label{app:tag_noise}
In this section, we include several experiments to show our method's robustness in settings with noisy and sparse tags.
The first experiment here uses a setup similar to the previous experiment on undirecting directed edges (Sec.~\ref{sec:undirected}).
We consider the ground truth causal graph and, considering every edge once, undirect this single edge and record whether it again gets directed correctly, incorrectly, or remains undirected. Edges that would be directed using Meek rules are excluded from this evaluation, as we only want to investigate the effects of tagging.
We differ from the previous experiment by adding $N\%$ tag faults with $N=[0,10,20,30,40,50,60,70,80,90]$ (with $100\%$ representing the current number of tags).
To be precise, we add $\frac{N}{2}\%$ random tags to variables that do not already have this tag and remove $\frac{X}{2}\%$ tags that were originally assigned by the LLM.
Then, for any edge in the graph, we undirected it and record the correctness of the prediction according to tagging based on the noisy tag set and all the other directed edges.
We average across 10 seeds and record the accuracy of correct predictions (ignoring edges that are not directed).
The results are shown in Figure~\ref{fig:ablations_tag_all}.

We can see that even if a large number of faults are introduced, the average accuracy remains relatively stable. This demonstrates that our approach is quite robust to suboptimal tagging, as incoherent tags get down-weighted by our approach. Note that even if $50\%$ of the tags are faulty, tagging still results in better-than-random predictions on average.
While accuracy further decreases with noise going up to $90\%$, average accuracy is still not clearly harmful (i.e., below $50\%$), and the larger datasets in particular tend to maintain a substantially higher accuracy than random guessing across LLMs.
Note that some outlier behavior can be observed for Cancer and Earthquake.
As these datasets are very small (4 edges each), our tagging statistics are highly susceptible to noise, and edges can be directed incorrectly due to just a small amount of bad evidence.
The black, dotted line shows that the number of abstentions increases when evidence becomes less clear, highlighting the robustness of our approach in noisy settings (abstention rates are shown on a per-dataset basis in Fig.~\ref{fig:ablations_tag_abstain_all}).

\begin{figure}[h]
    \centering
    \begin{subfigure}{0.45\textwidth}
        \centering
        \includegraphics[width=\textwidth]{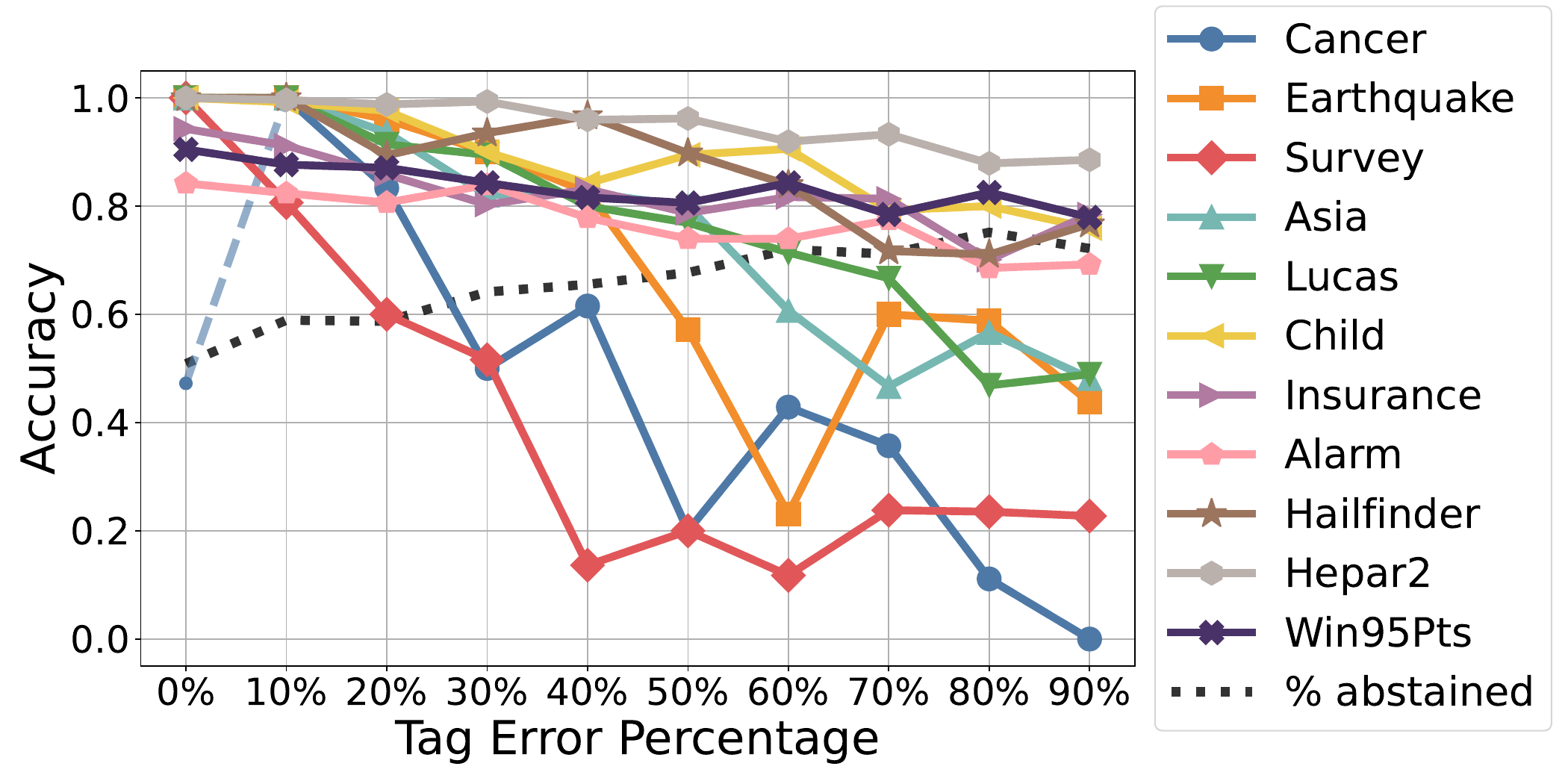}
        \caption{\textbf{GPT 5.2}}
        \label{fig:ablation_tag1}
    \end{subfigure}
    \hfill
    \begin{subfigure}{0.45\textwidth}
        \centering
        \includegraphics[width=\textwidth]{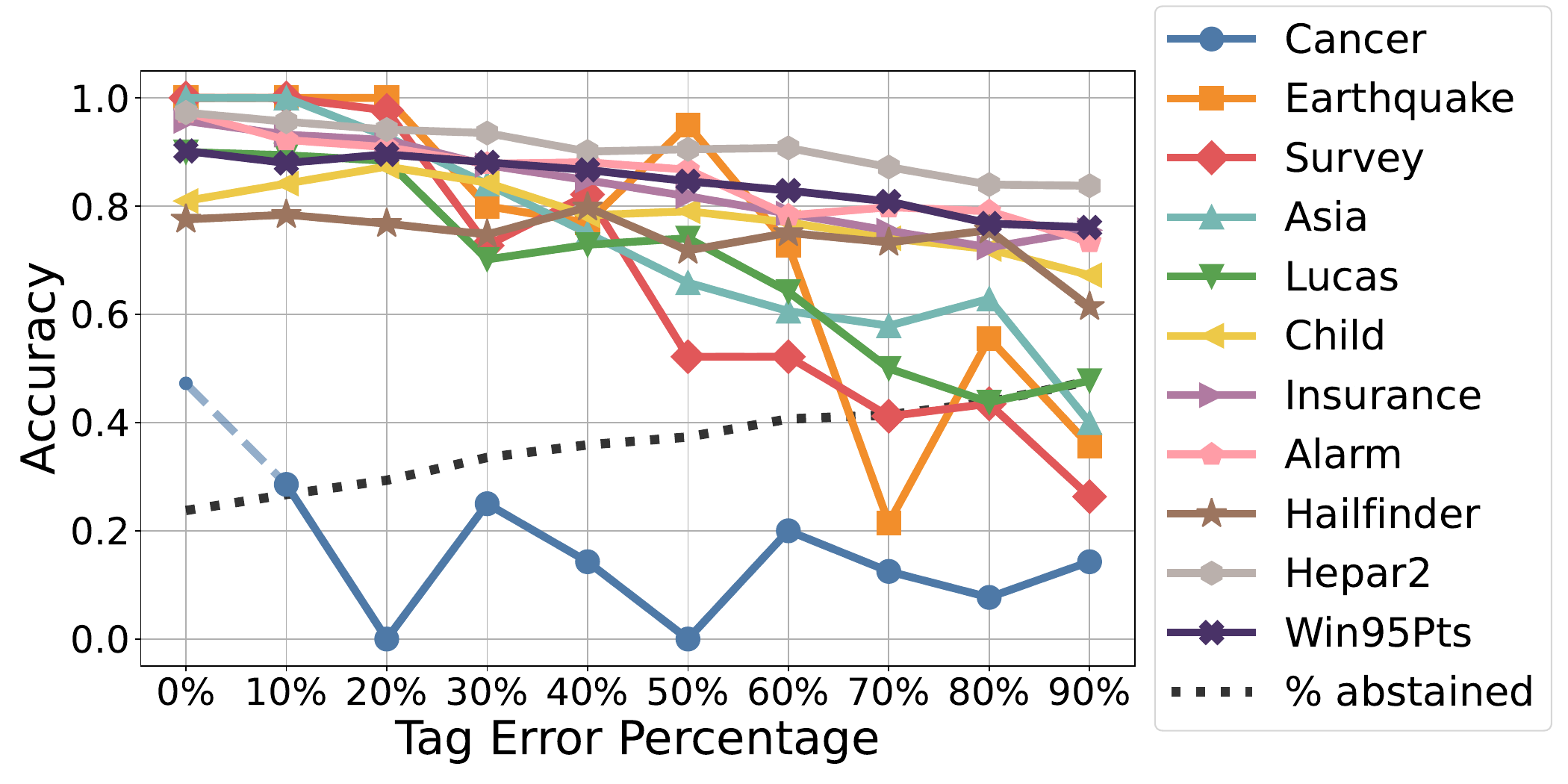}
        \caption{\textbf{Claude 4.6}}
        \label{fig:ablation_tag2}
    \end{subfigure}

    \vspace{0.5cm} %
    
    \begin{subfigure}{0.45\textwidth}
        \centering
        \includegraphics[width=\textwidth]{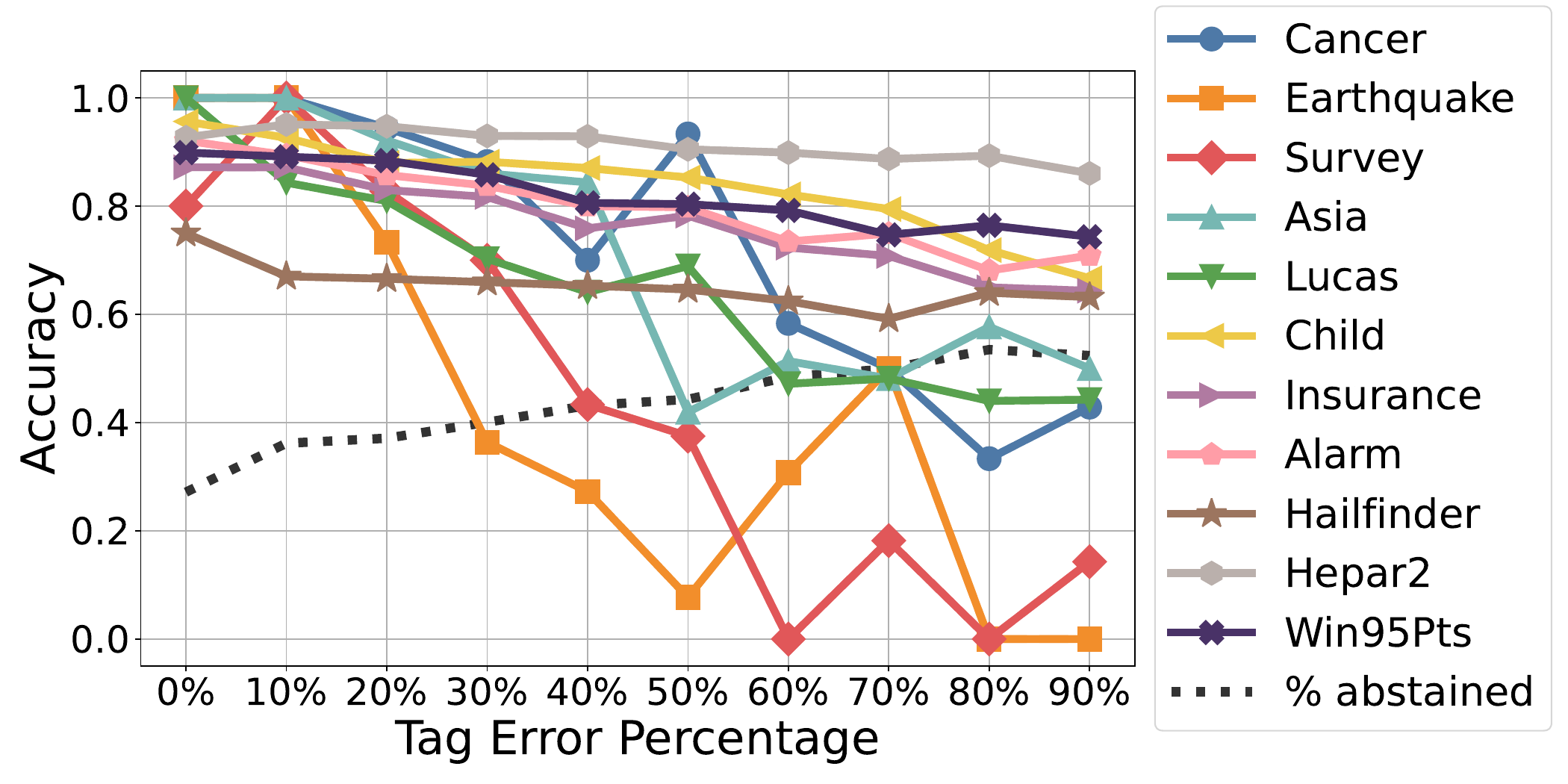}
        \caption{\textbf{Gemini 3 Pro}}
        \label{fig:ablation_tag3}
    \end{subfigure}
    \hfill
    \begin{subfigure}{0.45\textwidth}
        \centering
        \includegraphics[width=\textwidth]{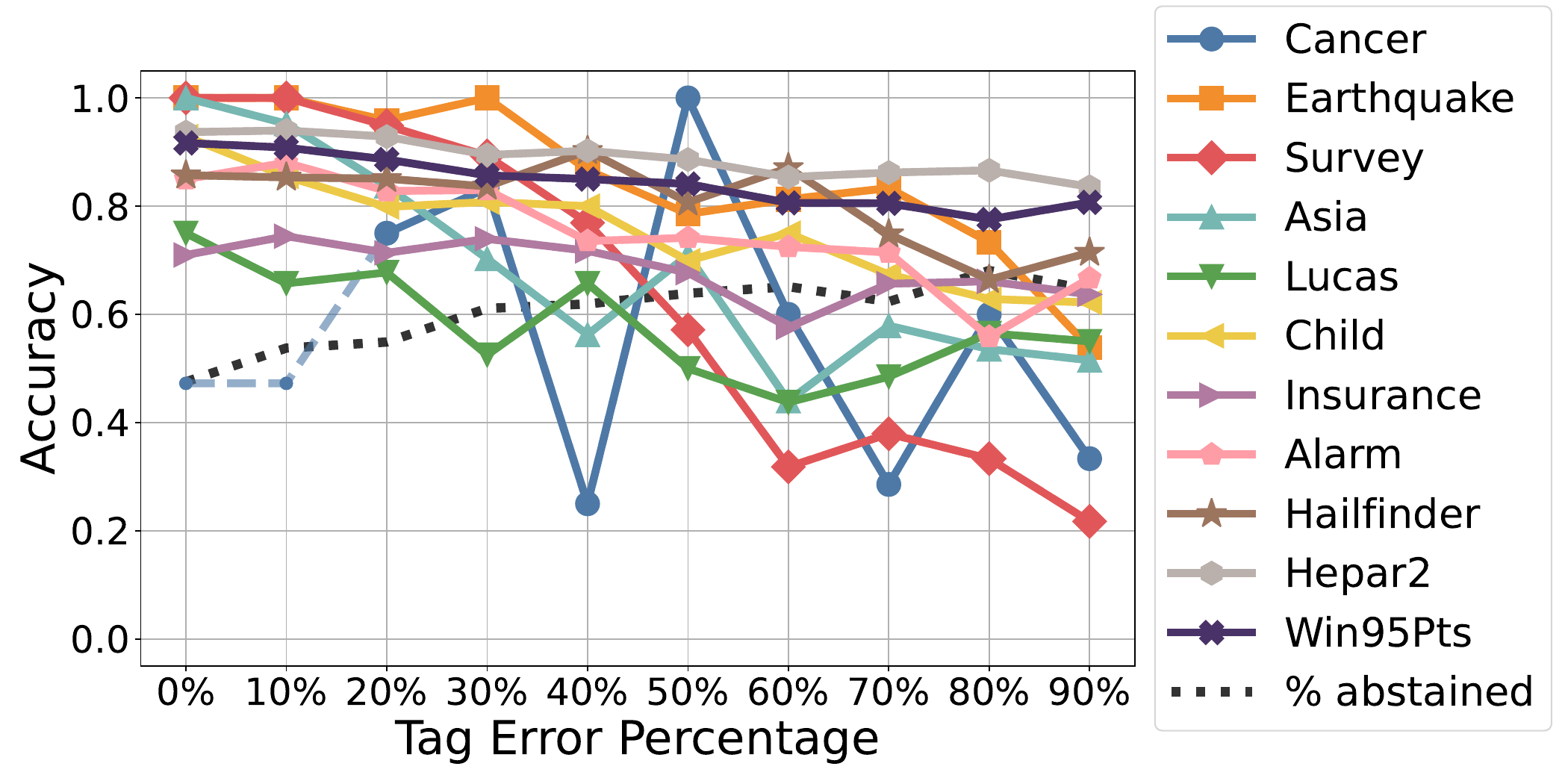}
        \caption{\textbf{Llama 3.3}}
        \label{fig:ablation_tag4}
    \end{subfigure}

    \vspace{0.5cm} %
    
    \begin{subfigure}{0.45\textwidth}
        \centering
        \includegraphics[width=\textwidth]{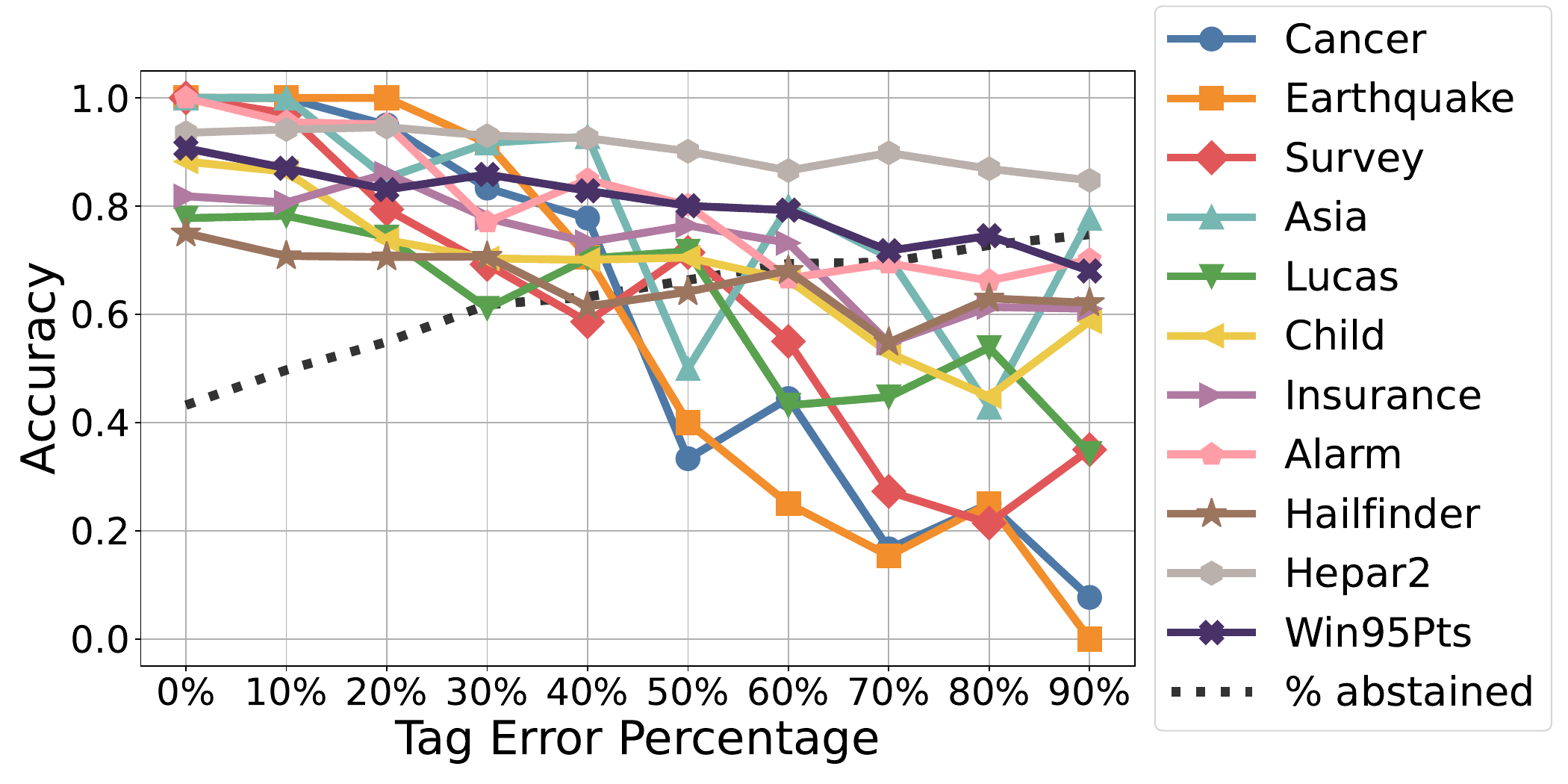}
        \caption{\textbf{Qwen 3.5}}
        \label{fig:ablation_tag5}
    \end{subfigure}
    \hfill
    \begin{subfigure}{0.45\textwidth}
        \centering
        \includegraphics[width=\textwidth]{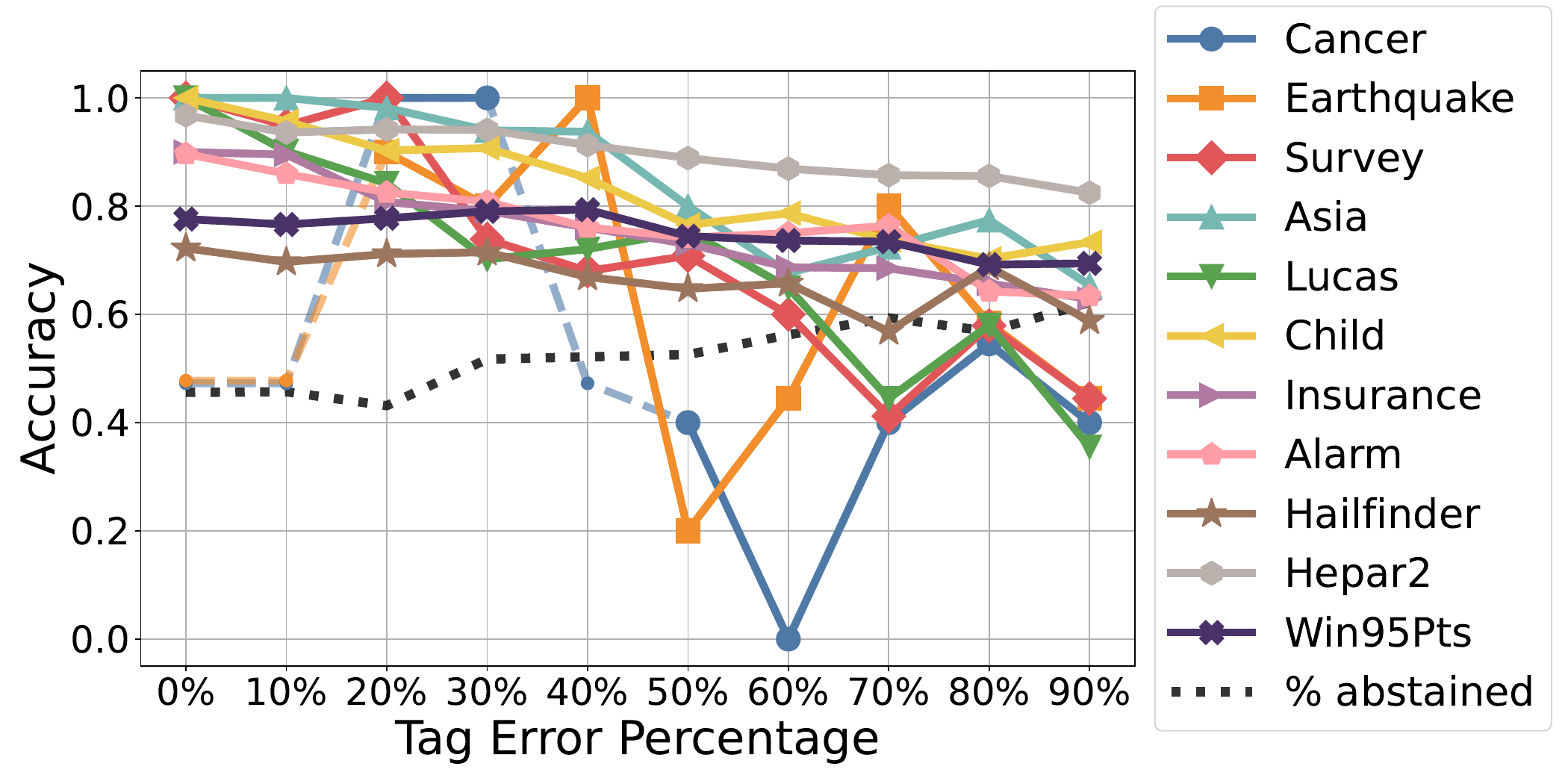}
        \caption{\textbf{GLM 5}}
        \label{fig:ablation_tag6}
    \end{subfigure}

    \vspace{0.5cm} %
    
    \begin{subfigure}{0.45\textwidth}
        \centering
        \includegraphics[width=\textwidth]{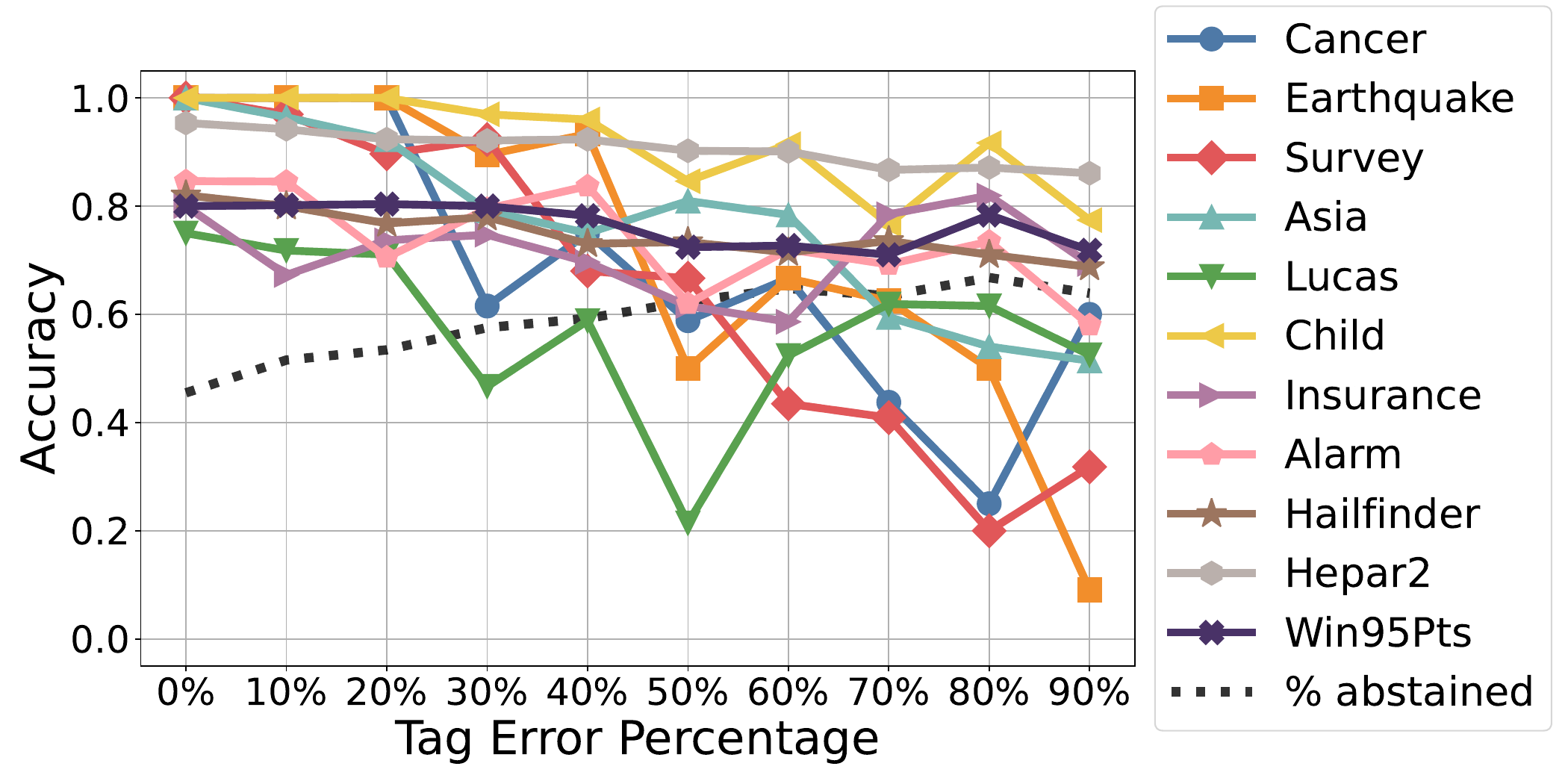}
        \caption{\textbf{Minimax 2.5}}
        \label{fig:ablation_tag7}
    \end{subfigure}
    
    \caption{\textbf{Accuracy with Noisy Tags.} Tagging predicts the correct edge direction more often than not, even with high tag error percentages. (We set accuracy to 50\% for datasets where no edge was directed; shown by a smaller point and a dashed line.) The black, dotted line shows how often the algorithm abstains from making a decision (also see Fig.~\ref{fig:ablations_tag_abstain_all}).}
    \label{fig:ablations_tag_all}
\end{figure}

\begin{figure}[h]
    \centering
    \begin{subfigure}{0.45\textwidth}
        \centering
        \includegraphics[width=\textwidth]{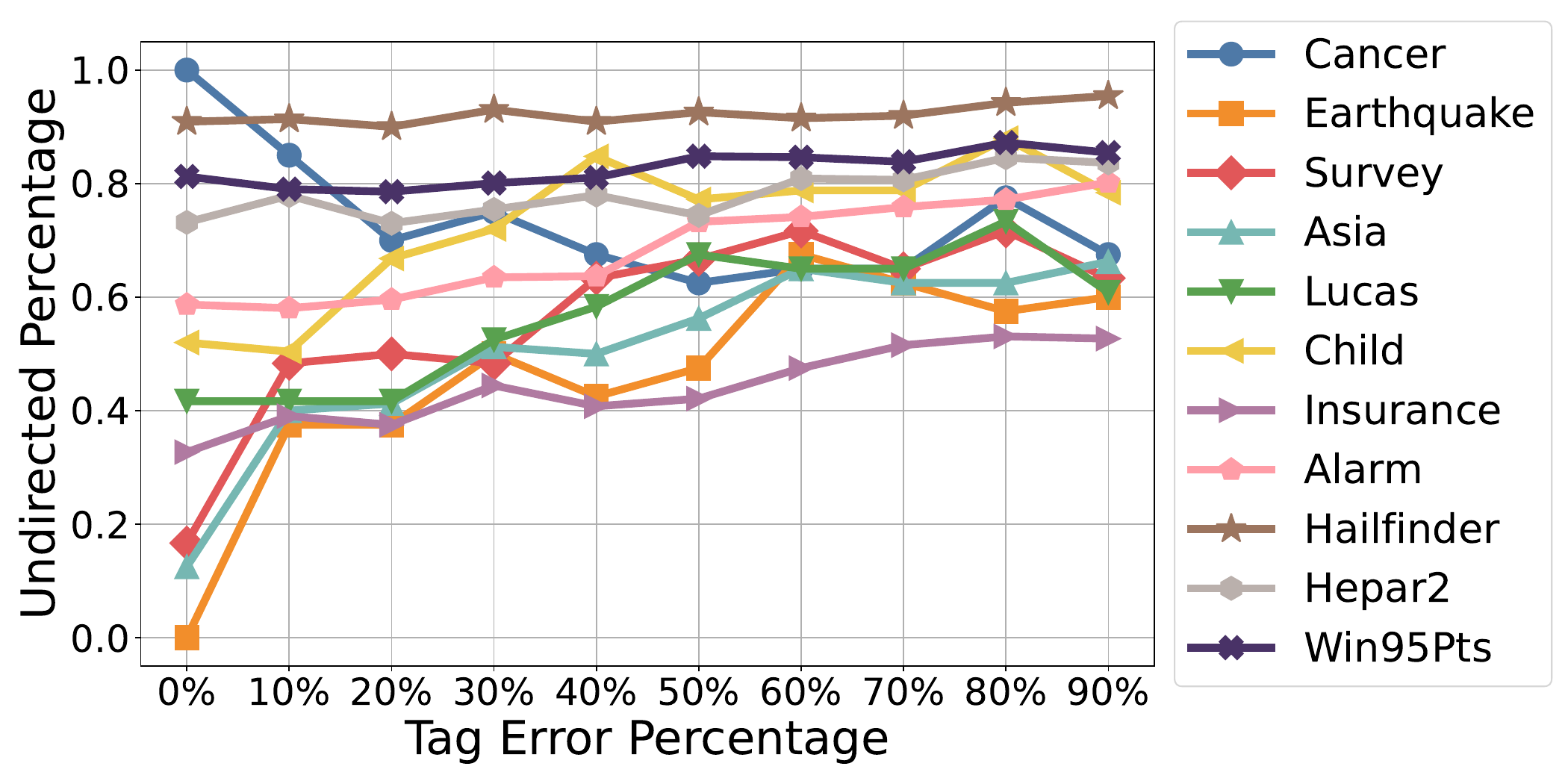}
        \caption{\textbf{GPT 5.2}}
        \label{fig:ablation_abstain_tag1}
    \end{subfigure}
    \hfill
    \begin{subfigure}{0.45\textwidth}
        \centering
        \includegraphics[width=\textwidth]{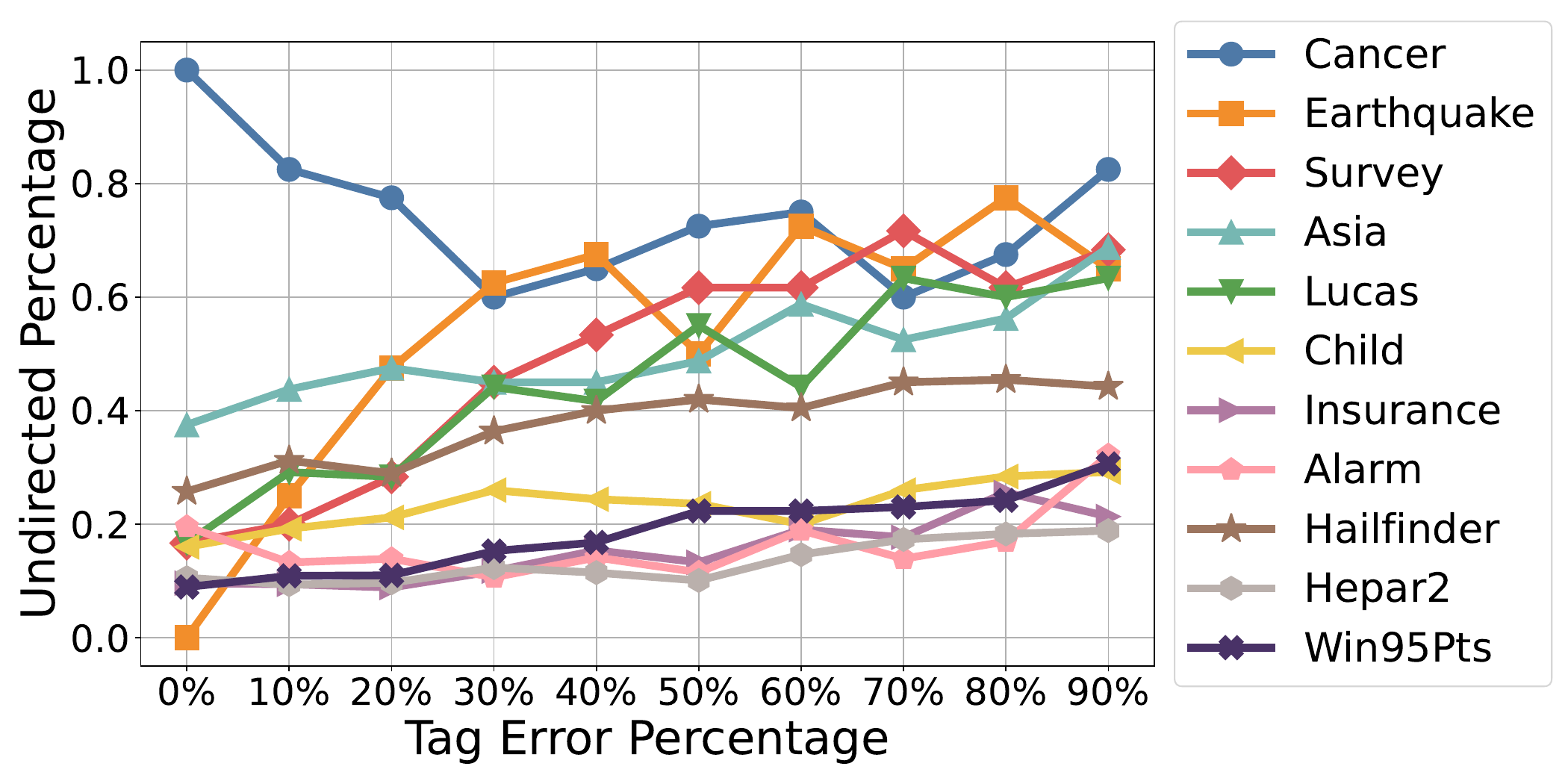}
        \caption{\textbf{Claude 4.6}}
        \label{fig:ablation_abstain_tag2}
    \end{subfigure}

    \vspace{0.5cm} %
    
    \begin{subfigure}{0.45\textwidth}
        \centering
        \includegraphics[width=\textwidth]{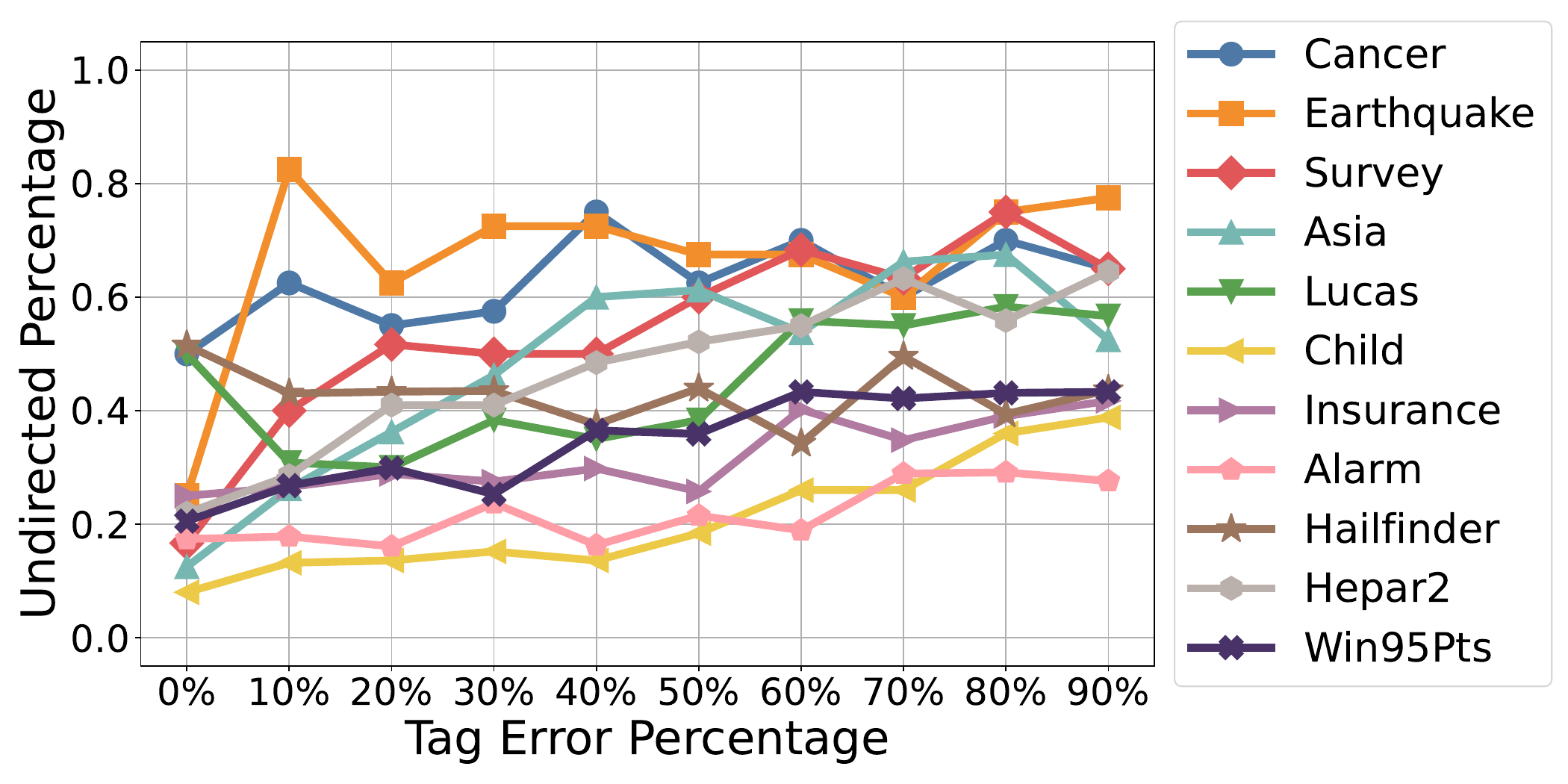}
        \caption{\textbf{Gemini 3 Pro}}
        \label{fig:ablation_abstain_tag3}
    \end{subfigure}
    \hfill
    \begin{subfigure}{0.45\textwidth}
        \centering
        \includegraphics[width=\textwidth]{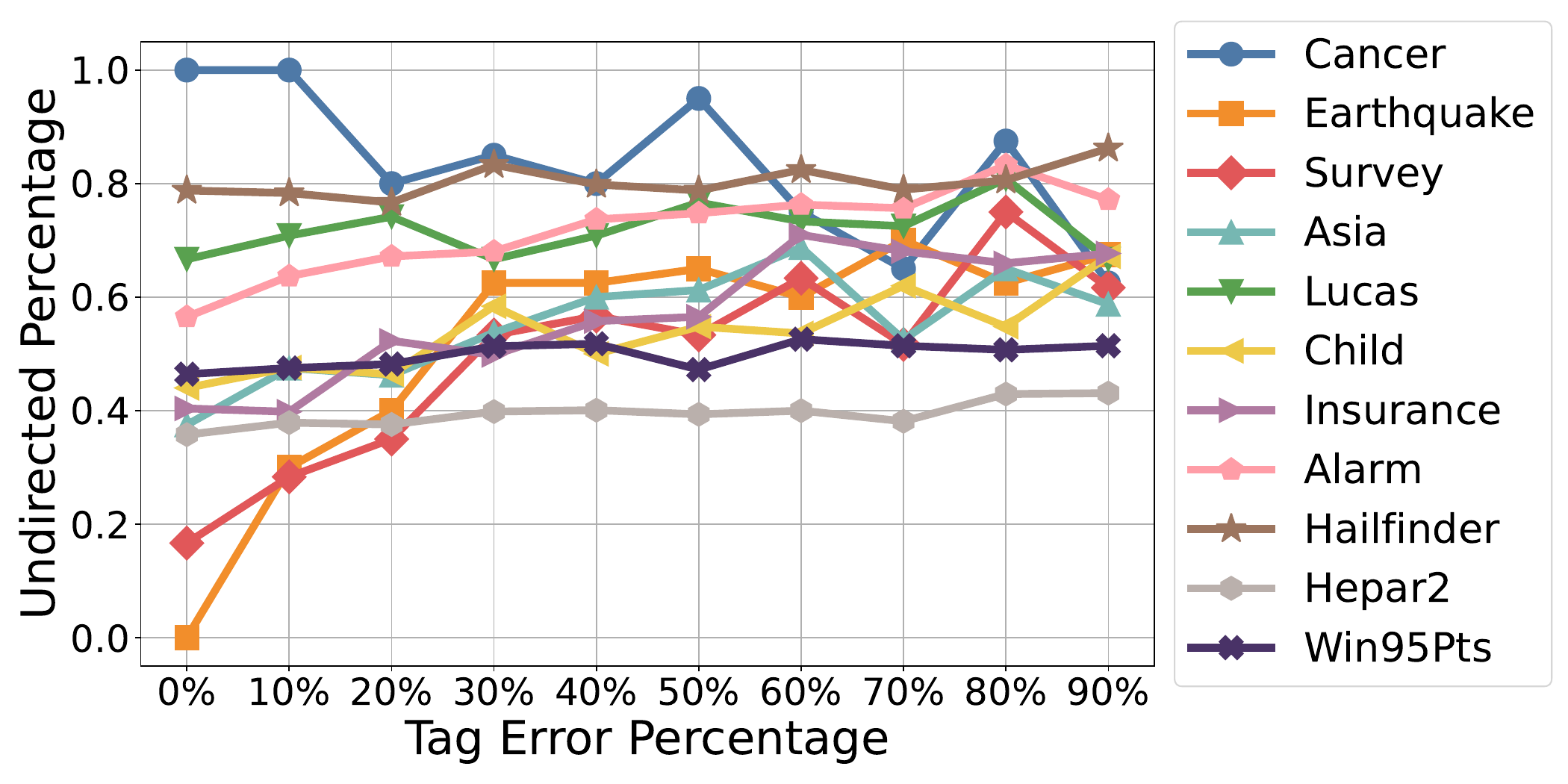}
        \caption{\textbf{Llama 3.3}}
        \label{fig:ablation_abstain_tag4}
    \end{subfigure}

    \vspace{0.5cm} %
    
    \begin{subfigure}{0.45\textwidth}
        \centering
        \includegraphics[width=\textwidth]{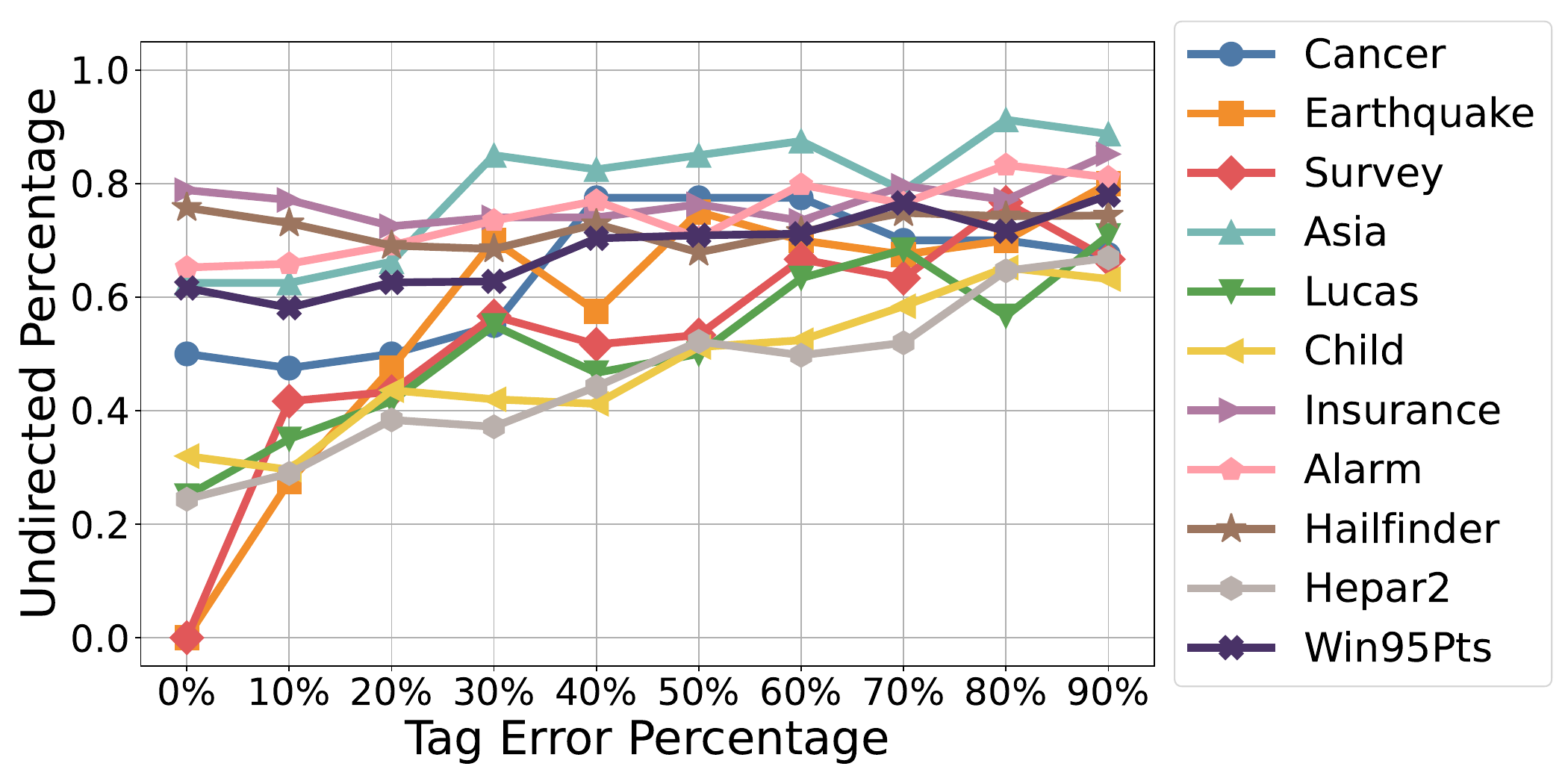}
        \caption{\textbf{Qwen 3.5}}
        \label{fig:ablation_abstain_tag5}
    \end{subfigure}
    \hfill
    \begin{subfigure}{0.45\textwidth}
        \centering
        \includegraphics[width=\textwidth]{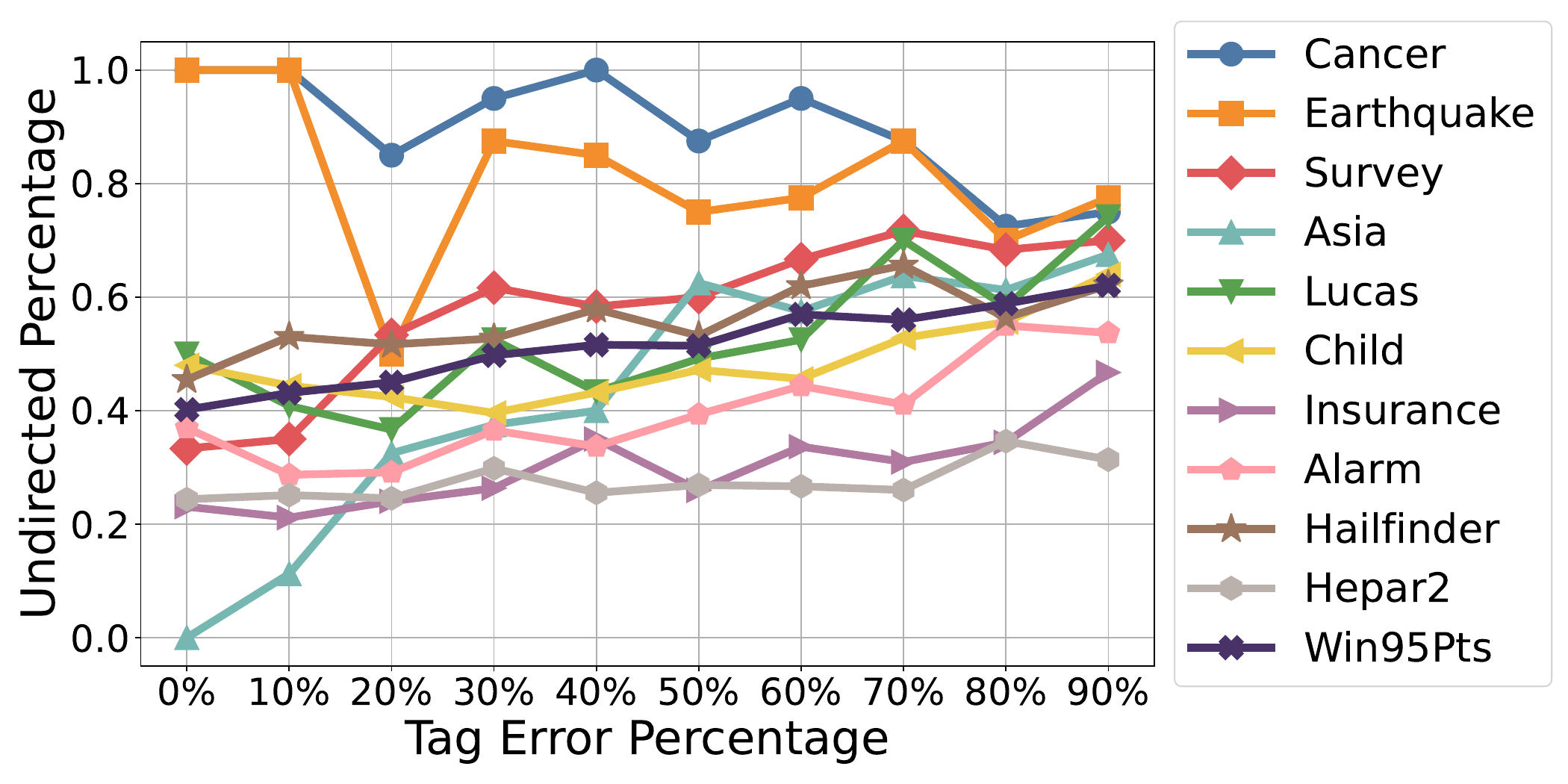}
        \caption{\textbf{GLM 5}}
        \label{fig:ablation_abstain_tag6}
    \end{subfigure}

    \vspace{0.5cm} %
    
    \begin{subfigure}{0.45\textwidth}
        \centering
        \includegraphics[width=\textwidth]{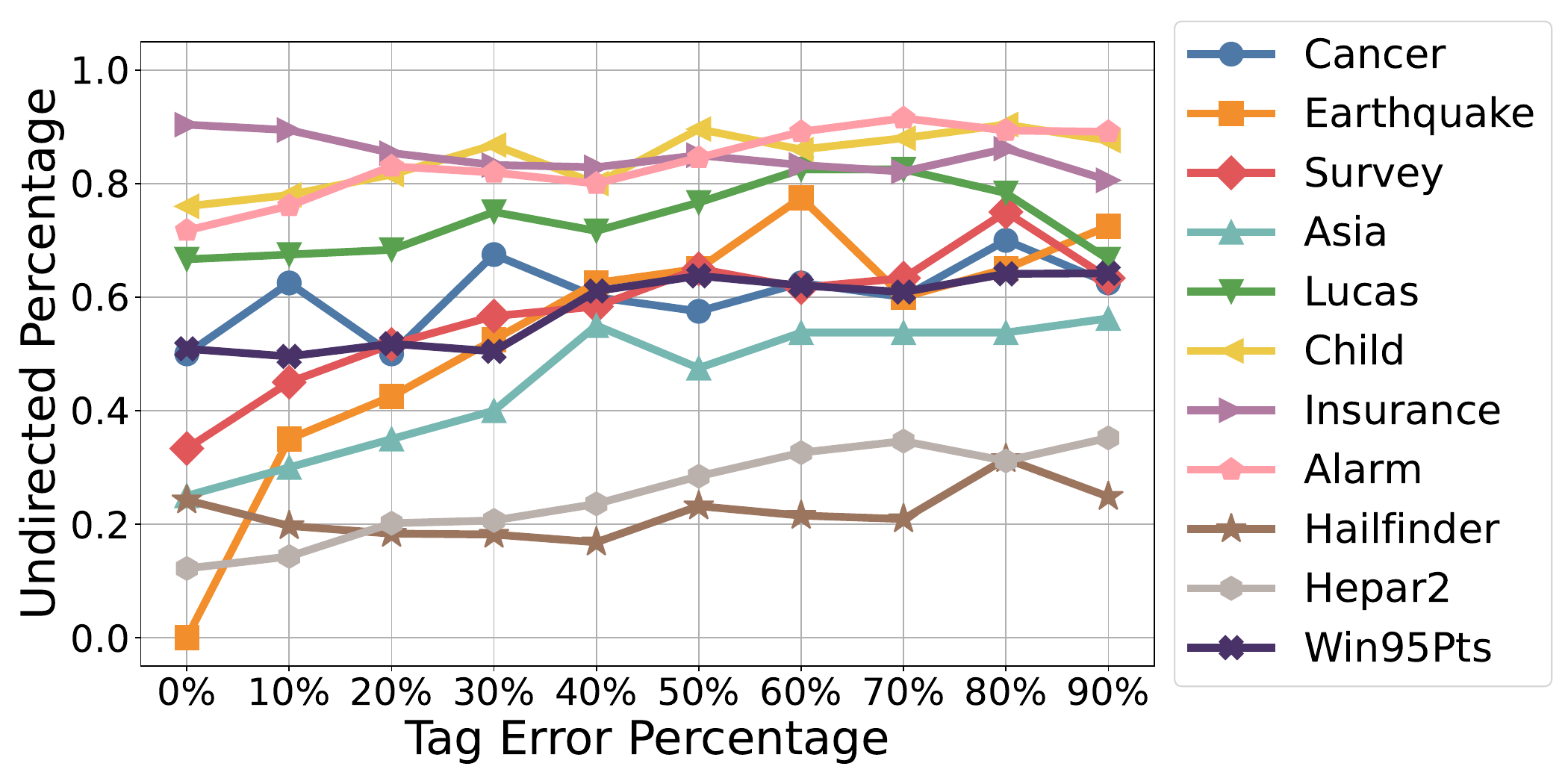}
        \caption{\textbf{Minimax 2.5}}
        \label{fig:ablation_abstain_tag7}
    \end{subfigure}
    
    \caption{\textbf{Percentage of Abstentions under Noisy Tags.} There is a clear general trend where the algorithm abstains from directing an edge more often for the noisier tag settings.}
    \label{fig:ablations_tag_abstain_all}
\end{figure}

Additionally, we run our evaluation from Section~\ref{sec:main_eval} but now with 50\% noise added in the same manner as for the aforementioned experiments.
The comparison of the main evaluation results to the noisy results can be found in Tab.~\ref{tab:noisy_eval}.
We see that the non-noisy setup performs better on average, but the reduction in performance is relatively small, even for this highly noisy setup. 

\begin{table}[]
    \centering
\resizebox{0.95\textwidth}{!}{
\begin{tabular}{l|ccccccc}
\textbf{Evaluation Results} & SHD & SHD\textsubscript{double} & SID\textsubscript{min} & SID\textsubscript{max} & Precision & Recall & F\textsubscript{1} \\
\hline \hline
Dataset Cancer & & & & & & & \\
\hline
Tagged-PC & $\mathbf{2.00} {\scriptstyle \pm 0.45}$ & $\mathbf{3.90} {\scriptstyle \pm 0.70}$ & $\mathbf{9.60} {\scriptstyle \pm 1.56}$ & $\mathbf{9.60} {\scriptstyle \pm 1.56}$ & $\mathbf{0.51} {\scriptstyle \pm 0.09}$ & $\mathbf{0.50} {\scriptstyle \pm 0.11}$ & $\mathbf{0.50} {\scriptstyle \pm 0.10}$ \\
Tagged-PC (N) & $\mathbf{2.00} {\scriptstyle \pm 0.45}$ & $\mathbf{3.90} {\scriptstyle \pm 0.70}$ & $\mathbf{9.60} {\scriptstyle \pm 1.56}$ & $\mathbf{9.60} {\scriptstyle \pm 1.56}$ & $\mathbf{0.51} {\scriptstyle \pm 0.09}$ & $\mathbf{0.50} {\scriptstyle \pm 0.11}$ & $\mathbf{0.50} {\scriptstyle \pm 0.10}$ \\
\hline
Tagged-GES & $\mathbf{0.80} {\scriptstyle \pm 1.60}$ & $\mathbf{0.80} {\scriptstyle \pm 1.60}$ & $\mathbf{0.20} {\scriptstyle \pm 0.40}$ & $\mathbf{2.20} {\scriptstyle \pm 4.40}$ & $\mathbf{0.80} {\scriptstyle \pm 0.40}$ & $\mathbf{0.80} {\scriptstyle \pm 0.40}$ & $\mathbf{0.80} {\scriptstyle \pm 0.40}$ \\
Tagged-GES (N) & $\mathbf{0.80} {\scriptstyle \pm 1.60}$ & $\mathbf{0.80} {\scriptstyle \pm 1.60}$ & $\mathbf{0.20} {\scriptstyle \pm 0.40}$ & $\mathbf{2.20} {\scriptstyle \pm 4.40}$ & $\mathbf{0.80} {\scriptstyle \pm 0.40}$ & $\mathbf{0.80} {\scriptstyle \pm 0.40}$ & $\mathbf{0.80} {\scriptstyle \pm 0.40}$ \\
\hline \hline
Dataset Earthquake & & & & & & & \\
\hline
Tagged-PC & $\mathbf{0.60} {\scriptstyle \pm 1.50}$ & $\mathbf{0.80} {\scriptstyle \pm 1.83}$ & $\mathbf{0.00} {\scriptstyle \pm 0.00}$ & $\mathbf{1.40} {\scriptstyle \pm 4.20}$ & $\mathbf{0.90} {\scriptstyle \pm 0.30}$ & $\mathbf{0.90} {\scriptstyle \pm 0.30}$ & $\mathbf{0.90} {\scriptstyle \pm 0.30}$ \\
Tagged-PC (N) & $\mathbf{0.60} {\scriptstyle \pm 1.50}$ & $\mathbf{0.80} {\scriptstyle \pm 1.83}$ & $\mathbf{0.00} {\scriptstyle \pm 0.00}$ & $\mathbf{1.40} {\scriptstyle \pm 4.20}$ & $\mathbf{0.90} {\scriptstyle \pm 0.30}$ & $\mathbf{0.90} {\scriptstyle \pm 0.30}$ & $\mathbf{0.90} {\scriptstyle \pm 0.30}$ \\
\hline
Tagged-GES & $\mathbf{0.00} {\scriptstyle \pm 0.00}$ & $\mathbf{0.00} {\scriptstyle \pm 0.00}$ & $\mathbf{0.00} {\scriptstyle \pm 0.00}$ & $\mathbf{0.00} {\scriptstyle \pm 0.00}$ & $\mathbf{1.00} {\scriptstyle \pm 0.00}$ & $\mathbf{1.00} {\scriptstyle \pm 0.00}$ & $\mathbf{1.00} {\scriptstyle \pm 0.00}$ \\
Tagged-GES (N) & $\mathbf{0.00} {\scriptstyle \pm 0.00}$ & $\mathbf{0.00} {\scriptstyle \pm 0.00}$ & $\mathbf{0.00} {\scriptstyle \pm 0.00}$ & $\mathbf{0.00} {\scriptstyle \pm 0.00}$ & $\mathbf{1.00} {\scriptstyle \pm 0.00}$ & $\mathbf{1.00} {\scriptstyle \pm 0.00}$ & $\mathbf{1.00} {\scriptstyle \pm 0.00}$ \\
\hline \hline
Dataset Survey & & & & & & & \\
\hline
Tagged-PC & $\mathbf{2.20} {\scriptstyle \pm 0.40}$ & $\mathbf{4.30} {\scriptstyle \pm 0.64}$ & $11.20 {\scriptstyle \pm 2.04}$ & $\mathbf{11.20} {\scriptstyle \pm 2.04}$ & $\mathbf{0.64} {\scriptstyle \pm 0.05}$ & $\mathbf{0.63} {\scriptstyle \pm 0.07}$ & $\mathbf{0.64} {\scriptstyle \pm 0.06}$ \\
Tagged-PC (N) & $2.30 {\scriptstyle \pm 0.64}$ & $\mathbf{4.30} {\scriptstyle \pm 0.64}$ & $\mathbf{11.10} {\scriptstyle \pm 1.92}$ & $11.80 {\scriptstyle \pm 2.44}$ & $\mathbf{0.64} {\scriptstyle \pm 0.05}$ & $0.62 {\scriptstyle \pm 0.11}$ & $0.63 {\scriptstyle \pm 0.08}$ \\
\hline
Tagged-GES & $4.20 {\scriptstyle \pm 1.89}$ & $4.30 {\scriptstyle \pm 1.79}$ & $\mathbf{8.50} {\scriptstyle \pm 2.06}$ & $15.80 {\scriptstyle \pm 7.52}$ & $\mathbf{0.48} {\scriptstyle \pm 0.48}$ & $0.30 {\scriptstyle \pm 0.31}$ & $0.36 {\scriptstyle \pm 0.37}$ \\
Tagged-GES (N) & $\mathbf{4.10} {\scriptstyle \pm 1.97}$ & $\mathbf{4.20} {\scriptstyle \pm 1.89}$ & $\mathbf{8.50} {\scriptstyle \pm 2.06}$ & $\mathbf{15.70} {\scriptstyle \pm 7.60}$ & $\mathbf{0.48} {\scriptstyle \pm 0.48}$ & $\mathbf{0.32} {\scriptstyle \pm 0.33}$ & $\mathbf{0.38} {\scriptstyle \pm 0.38}$ \\
\hline \hline
Dataset Asia & & & & & & & \\
\hline
Tagged-PC & $5.10 {\scriptstyle \pm 0.30}$ & $5.40 {\scriptstyle \pm 0.92}$ & $\mathbf{14.90} {\scriptstyle \pm 5.39}$ & $24.20 {\scriptstyle \pm 3.12}$ & $\mathbf{0.93} {\scriptstyle \pm 0.16}$ & $0.38 {\scriptstyle \pm 0.00}$ & $0.53 {\scriptstyle \pm 0.04}$ \\
Tagged-PC (N) & $\mathbf{4.10} {\scriptstyle \pm 1.22}$ & $\mathbf{4.40} {\scriptstyle \pm 1.62}$ & $15.10 {\scriptstyle \pm 4.21}$ & $\mathbf{20.80} {\scriptstyle \pm 6.65}$ & $0.91 {\scriptstyle \pm 0.16}$ & $\mathbf{0.50} {\scriptstyle \pm 0.16}$ & $\mathbf{0.63} {\scriptstyle \pm 0.16}$ \\
\hline
Tagged-GES & $3.00 {\scriptstyle \pm 0.00}$ & $3.00 {\scriptstyle \pm 0.00}$ & $\mathbf{0.80} {\scriptstyle \pm 0.40}$ & $8.80 {\scriptstyle \pm 1.60}$ & $\mathbf{1.00} {\scriptstyle \pm 0.00}$ & $0.62 {\scriptstyle \pm 0.00}$ & $0.77 {\scriptstyle \pm 0.00}$ \\
Tagged-GES (N) & $\mathbf{2.00} {\scriptstyle \pm 0.77}$ & $\mathbf{2.60} {\scriptstyle \pm 1.11}$ & $4.60 {\scriptstyle \pm 3.93}$ & $\mathbf{6.70} {\scriptstyle \pm 4.03}$ & $0.92 {\scriptstyle \pm 0.09}$ & $\mathbf{0.75} {\scriptstyle \pm 0.10}$ & $\mathbf{0.82} {\scriptstyle \pm 0.08}$ \\
\hline \hline
Dataset Lucas & & & & & & & \\
\hline
Tagged-PC & $\mathbf{0.00} {\scriptstyle \pm 0.00}$ & $\mathbf{0.00} {\scriptstyle \pm 0.00}$ & $\mathbf{0.00} {\scriptstyle \pm 0.00}$ & $\mathbf{0.00} {\scriptstyle \pm 0.00}$ & $\mathbf{1.00} {\scriptstyle \pm 0.00}$ & $\mathbf{1.00} {\scriptstyle \pm 0.00}$ & $\mathbf{1.00} {\scriptstyle \pm 0.00}$ \\
Tagged-PC (N) & $0.30 {\scriptstyle \pm 0.46}$ & $0.30 {\scriptstyle \pm 0.46}$ & $\mathbf{0.00} {\scriptstyle \pm 0.00}$ & $2.10 {\scriptstyle \pm 3.21}$ & $\mathbf{1.00} {\scriptstyle \pm 0.00}$ & $0.97 {\scriptstyle \pm 0.04}$ & $0.99 {\scriptstyle \pm 0.02}$ \\
\hline
Tagged-GES & $\mathbf{0.00} {\scriptstyle \pm 0.00}$ & $\mathbf{0.00} {\scriptstyle \pm 0.00}$ & $\mathbf{0.00} {\scriptstyle \pm 0.00}$ & $\mathbf{0.00} {\scriptstyle \pm 0.00}$ & $\mathbf{1.00} {\scriptstyle \pm 0.00}$ & $\mathbf{1.00} {\scriptstyle \pm 0.00}$ & $\mathbf{1.00} {\scriptstyle \pm 0.00}$ \\
Tagged-GES (N) & $0.60 {\scriptstyle \pm 0.49}$ & $0.70 {\scriptstyle \pm 0.64}$ & $1.00 {\scriptstyle \pm 3.00}$ & $4.50 {\scriptstyle \pm 3.77}$ & $0.99 {\scriptstyle \pm 0.03}$ & $0.95 {\scriptstyle \pm 0.04}$ & $0.97 {\scriptstyle \pm 0.03}$ \\
\hline \hline
Dataset Child & & & & & & & \\
\hline
Tagged-PC & $\mathbf{1.30} {\scriptstyle \pm 1.19}$ & $\mathbf{2.50} {\scriptstyle \pm 2.42}$ & $\mathbf{21.00} {\scriptstyle \pm 23.48}$ & $\mathbf{21.00} {\scriptstyle \pm 23.48}$ & $\mathbf{0.95} {\scriptstyle \pm 0.05}$ & $\mathbf{0.95} {\scriptstyle \pm 0.05}$ & $\mathbf{0.95} {\scriptstyle \pm 0.05}$ \\
Tagged-PC (N) & $4.10 {\scriptstyle \pm 1.51}$ & $7.50 {\scriptstyle \pm 2.33}$ & $76.00 {\scriptstyle \pm 22.26}$ & $82.60 {\scriptstyle \pm 32.13}$ & $0.86 {\scriptstyle \pm 0.04}$ & $0.84 {\scriptstyle \pm 0.06}$ & $0.85 {\scriptstyle \pm 0.05}$ \\
\hline
Tagged-GES & $3.00 {\scriptstyle \pm 0.00}$ & $\mathbf{3.00} {\scriptstyle \pm 0.00}$ & $\mathbf{0.00} {\scriptstyle \pm 0.00}$ & $44.00 {\scriptstyle \pm 0.00}$ & $\mathbf{1.00} {\scriptstyle \pm 0.00}$ & $0.88 {\scriptstyle \pm 0.00}$ & $\mathbf{0.94} {\scriptstyle \pm 0.00}$ \\
Tagged-GES (N) & $\mathbf{2.50} {\scriptstyle \pm 1.28}$ & $4.30 {\scriptstyle \pm 1.90}$ & $29.80 {\scriptstyle \pm 19.40}$ & $\mathbf{37.30} {\scriptstyle \pm 19.57}$ & $0.93 {\scriptstyle \pm 0.04}$ & $\mathbf{0.90} {\scriptstyle \pm 0.05}$ & $0.91 {\scriptstyle \pm 0.04}$ \\
\hline \hline
Dataset Alarm & & & & & & & \\
\hline
Tagged-PC & $\mathbf{7.10} {\scriptstyle \pm 3.11}$ & $\mathbf{9.20} {\scriptstyle \pm 4.85}$ & $\mathbf{72.40} {\scriptstyle \pm 45.53}$ & $\mathbf{81.70} {\scriptstyle \pm 50.81}$ & $\mathbf{0.93} {\scriptstyle \pm 0.06}$ & $\mathbf{0.88} {\scriptstyle \pm 0.05}$ & $\mathbf{0.91} {\scriptstyle \pm 0.05}$ \\
Tagged-PC (N) & $8.00 {\scriptstyle \pm 2.79}$ & $10.50 {\scriptstyle \pm 5.20}$ & $74.90 {\scriptstyle \pm 55.18}$ & $87.10 {\scriptstyle \pm 49.88}$ & $0.91 {\scriptstyle \pm 0.07}$ & $0.86 {\scriptstyle \pm 0.04}$ & $0.89 {\scriptstyle \pm 0.05}$ \\
\hline
Tagged-GES & $\mathbf{4.50} {\scriptstyle \pm 1.28}$ & $\mathbf{4.50} {\scriptstyle \pm 1.28}$ & $\mathbf{18.40} {\scriptstyle \pm 0.80}$ & $\mathbf{36.40} {\scriptstyle \pm 0.80}$ & $\mathbf{0.97} {\scriptstyle \pm 0.02}$ & $\mathbf{0.93} {\scriptstyle \pm 0.01}$ & $\mathbf{0.95} {\scriptstyle \pm 0.01}$ \\
Tagged-GES (N) & $6.70 {\scriptstyle \pm 2.24}$ & $9.00 {\scriptstyle \pm 2.97}$ & $61.80 {\scriptstyle \pm 18.25}$ & $74.00 {\scriptstyle \pm 21.15}$ & $0.92 {\scriptstyle \pm 0.04}$ & $0.88 {\scriptstyle \pm 0.03}$ & $0.90 {\scriptstyle \pm 0.03}$ \\
\hline \hline
Dataset Insurance & & & & & & & \\
\hline
Tagged-PC & $\mathbf{20.90} {\scriptstyle \pm 2.21}$ & $\mathbf{30.00} {\scriptstyle \pm 4.00}$ & $\mathbf{389.60} {\scriptstyle \pm 48.01}$ & $\mathbf{390.20} {\scriptstyle \pm 48.46}$ & $\mathbf{0.77} {\scriptstyle \pm 0.05}$ & $\mathbf{0.61} {\scriptstyle \pm 0.04}$ & $\mathbf{0.68} {\scriptstyle \pm 0.04}$ \\
Tagged-PC (N) & $21.50 {\scriptstyle \pm 2.54}$ & $31.50 {\scriptstyle \pm 4.90}$ & $392.70 {\scriptstyle \pm 47.13}$ & $395.70 {\scriptstyle \pm 50.13}$ & $0.75 {\scriptstyle \pm 0.06}$ & $0.60 {\scriptstyle \pm 0.05}$ & $0.66 {\scriptstyle \pm 0.05}$ \\
\hline
Tagged-GES & $\mathbf{18.40} {\scriptstyle \pm 4.15}$ & $\mathbf{22.40} {\scriptstyle \pm 6.55}$ & $\mathbf{250.30} {\scriptstyle \pm 37.97}$ & $\mathbf{259.30} {\scriptstyle \pm 30.59}$ & $\mathbf{0.86} {\scriptstyle \pm 0.08}$ & $\mathbf{0.69} {\scriptstyle \pm 0.06}$ & $\mathbf{0.76} {\scriptstyle \pm 0.07}$ \\
Tagged-GES (N) & $18.90 {\scriptstyle \pm 4.83}$ & $23.30 {\scriptstyle \pm 7.20}$ & $254.70 {\scriptstyle \pm 46.74}$ & $275.50 {\scriptstyle \pm 42.30}$ & $0.84 {\scriptstyle \pm 0.09}$ & $0.68 {\scriptstyle \pm 0.07}$ & $0.75 {\scriptstyle \pm 0.08}$ \\
\hline \hline
Dataset Hailfinder & & & & & & & \\
\hline
Tagged-PC & $\mathbf{41.90} {\scriptstyle \pm 3.45}$ & $\mathbf{44.70} {\scriptstyle \pm 3.93}$ & $-$ & $-$ & $\mathbf{0.80} {\scriptstyle \pm 0.04}$ & $\mathbf{0.45} {\scriptstyle \pm 0.04}$ & $\mathbf{0.58} {\scriptstyle \pm 0.04}$ \\
Tagged-PC (N) & $43.70 {\scriptstyle \pm 2.87}$ & $47.60 {\scriptstyle \pm 4.39}$ & $-$ & $-$ & $0.78 {\scriptstyle \pm 0.07}$ & $0.43 {\scriptstyle \pm 0.03}$ & $0.55 {\scriptstyle \pm 0.04}$ \\
\hline
Tagged-GES & $\mathbf{39.00} {\scriptstyle \pm 15.21}$ & $\mathbf{42.20} {\scriptstyle \pm 15.28}$ & $-$ & $-$ & $\mathbf{0.69} {\scriptstyle \pm 0.12}$ & $\mathbf{0.65} {\scriptstyle \pm 0.12}$ & $\mathbf{0.67} {\scriptstyle \pm 0.12}$ \\
Tagged-GES (N) & $\mathbf{39.00} {\scriptstyle \pm 15.21}$ & $\mathbf{42.20} {\scriptstyle \pm 15.28}$ & $-$ & $-$ & $\mathbf{0.69} {\scriptstyle \pm 0.12}$ & $\mathbf{0.65} {\scriptstyle \pm 0.12}$ & $\mathbf{0.67} {\scriptstyle \pm 0.12}$ \\
\hline \hline
Dataset Hepar2 & & & & & & & \\
\hline
Tagged-PC & $94.20 {\scriptstyle \pm 9.90}$ & $127.90 {\scriptstyle \pm 19.41}$ & $-$ & $-$ & $0.48 {\scriptstyle \pm 0.13}$ & $0.28 {\scriptstyle \pm 0.08}$ & $0.35 {\scriptstyle \pm 0.10}$ \\
Tagged-PC (N) & $\mathbf{91.50} {\scriptstyle \pm 5.55}$ & $\mathbf{123.00} {\scriptstyle \pm 11.34}$ & $-$ & $-$ & $\mathbf{0.51} {\scriptstyle \pm 0.08}$ & $\mathbf{0.30} {\scriptstyle \pm 0.04}$ & $\mathbf{0.38} {\scriptstyle \pm 0.05}$ \\
\hline
Tagged-GES & $\mathbf{50.30} {\scriptstyle \pm 1.55}$ & $\mathbf{55.00} {\scriptstyle \pm 2.32}$ & $-$ & $-$ & $\mathbf{0.94} {\scriptstyle \pm 0.02}$ & $\mathbf{0.59} {\scriptstyle \pm 0.01}$ & $\mathbf{0.73} {\scriptstyle \pm 0.01}$ \\
Tagged-GES (N) & $53.80 {\scriptstyle \pm 2.75}$ & $61.70 {\scriptstyle \pm 3.23}$ & $-$ & $-$ & $0.90 {\scriptstyle \pm 0.01}$ & $0.56 {\scriptstyle \pm 0.02}$ & $0.69 {\scriptstyle \pm 0.02}$ \\
\hline \hline
Dataset Win95Pts & & & & & & & \\
\hline
Tagged-PC & $\mathbf{48.70} {\scriptstyle \pm 3.61}$ & $\mathbf{57.00} {\scriptstyle \pm 6.36}$ & $-$ & $-$ & $\mathbf{0.84} {\scriptstyle \pm 0.04}$ & $\mathbf{0.61} {\scriptstyle \pm 0.02}$ & $\mathbf{0.71} {\scriptstyle \pm 0.03}$ \\
Tagged-PC (N) & $52.00 {\scriptstyle \pm 3.79}$ & $60.30 {\scriptstyle \pm 5.69}$ & $-$ & $-$ & $0.83 {\scriptstyle \pm 0.04}$ & $0.58 {\scriptstyle \pm 0.03}$ & $0.69 {\scriptstyle \pm 0.03}$ \\
\hline
Tagged-GES & $\mathbf{45.90} {\scriptstyle \pm 5.96}$ & $\mathbf{50.00} {\scriptstyle \pm 6.29}$ & $-$ & $-$ & $\mathbf{0.79} {\scriptstyle \pm 0.03}$ & $\mathbf{0.76} {\scriptstyle \pm 0.03}$ & $\mathbf{0.77} {\scriptstyle \pm 0.03}$ \\
Tagged-GES (N) & $49.30 {\scriptstyle \pm 5.66}$ & $59.30 {\scriptstyle \pm 6.17}$ & $-$ & $-$ & $0.74 {\scriptstyle \pm 0.03}$ & $0.73 {\scriptstyle \pm 0.03}$ & $0.73 {\scriptstyle \pm 0.03}$ \\
\hline \hline
\end{tabular}
}
    \caption{\textbf{Absolute Scores Comparing Normal and Noisy Tags.} Results for the standard tag sets are the same as in Tab.~\ref{tab:detailedOverallResultsAppx}. Below each method, we include the results for noisy (N) tags. Despite highly noisy tags, performance only decreases slightly.  ${}_{\pm y}$ indicates standard deviation.}
    \label{tab:noisy_eval}
\end{table}

In the last experiment in this section, we repeat our main evaluation once more.
However, now, instead of introducing several types of tag errors, we simply reduce the tag amount, rerunning the evaluation with $25\%$, $50\%$, and $75\%$ fewer tags per dataset.
Results are shown in Tab.~\ref{tab:reduced-tags-dataset-metrics}.
Unsurprisingly, using a smaller number of tags does not lead to better results.
This provides further evidence for our hypothesis that a larger number of tags generally only improves tagging accuracy, as noisy tags can be ignored, while useful tags are leveraged to improve prediction accuracy.

\begin{table*}[t]
\centering
\resizebox{0.95\textwidth}{!}{
\small
\begin{tabular}{l|c|ccccccc}
 & Fewer Tags & SHD & SHD\textsubscript{double} & SID\textsubscript{min} & SID\textsubscript{max} & Precision & Recall & F\textsubscript{1} \\
\hline \hline
Cancer & & & & & & & & \\
\hline
 & 0.00 & $0.80 {\scriptstyle \pm 1.60}$ & $0.80 {\scriptstyle \pm 1.60}$ & $0.20 {\scriptstyle \pm 0.40}$ & $2.20 {\scriptstyle \pm 4.40}$ & $0.80 {\scriptstyle \pm 0.40}$ & $0.80 {\scriptstyle \pm 0.40}$ & $0.80 {\scriptstyle \pm 0.40}$ \\
 & 0.25 & $0.80 {\scriptstyle \pm 1.60}$ & $0.80 {\scriptstyle \pm 1.60}$ & $0.20 {\scriptstyle \pm 0.40}$ & $2.20 {\scriptstyle \pm 4.40}$ & $0.80 {\scriptstyle \pm 0.40}$ & $0.80 {\scriptstyle \pm 0.40}$ & $0.80 {\scriptstyle \pm 0.40}$ \\
 & 0.50 & $0.80 {\scriptstyle \pm 1.60}$ & $0.80 {\scriptstyle \pm 1.60}$ & $0.20 {\scriptstyle \pm 0.40}$ & $2.20 {\scriptstyle \pm 4.40}$ & $0.80 {\scriptstyle \pm 0.40}$ & $0.80 {\scriptstyle \pm 0.40}$ & $0.80 {\scriptstyle \pm 0.40}$ \\
 & 0.75 & $0.80 {\scriptstyle \pm 1.60}$ & $0.80 {\scriptstyle \pm 1.60}$ & $0.20 {\scriptstyle \pm 0.40}$ & $2.20 {\scriptstyle \pm 4.40}$ & $0.80 {\scriptstyle \pm 0.40}$ & $0.80 {\scriptstyle \pm 0.40}$ & $0.80 {\scriptstyle \pm 0.40}$ \\
\hline \hline
Earthquake & & & & & & & & \\
\hline
 & 0.00 & $0.00 {\scriptstyle \pm 0.00}$ & $0.00 {\scriptstyle \pm 0.00}$ & $0.00 {\scriptstyle \pm 0.00}$ & $0.00 {\scriptstyle \pm 0.00}$ & $1.00 {\scriptstyle \pm 0.00}$ & $1.00 {\scriptstyle \pm 0.00}$ & $1.00 {\scriptstyle \pm 0.00}$ \\
 & 0.25 & $0.00 {\scriptstyle \pm 0.00}$ & $0.00 {\scriptstyle \pm 0.00}$ & $0.00 {\scriptstyle \pm 0.00}$ & $0.00 {\scriptstyle \pm 0.00}$ & $1.00 {\scriptstyle \pm 0.00}$ & $1.00 {\scriptstyle \pm 0.00}$ & $1.00 {\scriptstyle \pm 0.00}$ \\
 & 0.50 & $0.00 {\scriptstyle \pm 0.00}$ & $0.00 {\scriptstyle \pm 0.00}$ & $0.00 {\scriptstyle \pm 0.00}$ & $0.00 {\scriptstyle \pm 0.00}$ & $1.00 {\scriptstyle \pm 0.00}$ & $1.00 {\scriptstyle \pm 0.00}$ & $1.00 {\scriptstyle \pm 0.00}$ \\
 & 0.75 & $0.00 {\scriptstyle \pm 0.00}$ & $0.00 {\scriptstyle \pm 0.00}$ & $0.00 {\scriptstyle \pm 0.00}$ & $0.00 {\scriptstyle \pm 0.00}$ & $1.00 {\scriptstyle \pm 0.00}$ & $1.00 {\scriptstyle \pm 0.00}$ & $1.00 {\scriptstyle \pm 0.00}$ \\
\hline \hline
Survey & & & & & & & & \\
\hline
 & 0.00 & $4.20 {\scriptstyle \pm 1.89}$ & $4.30 {\scriptstyle \pm 1.79}$ & $8.50 {\scriptstyle \pm 2.06}$ & $15.80 {\scriptstyle \pm 7.52}$ & $0.48 {\scriptstyle \pm 0.48}$ & $0.30 {\scriptstyle \pm 0.31}$ & $0.36 {\scriptstyle \pm 0.37}$ \\
 & 0.25 & $4.20 {\scriptstyle \pm 1.89}$ & $4.30 {\scriptstyle \pm 1.79}$ & $8.50 {\scriptstyle \pm 2.06}$ & $15.80 {\scriptstyle \pm 7.52}$ & $0.48 {\scriptstyle \pm 0.48}$ & $0.30 {\scriptstyle \pm 0.31}$ & $0.36 {\scriptstyle \pm 0.37}$ \\
 & 0.50 & $4.20 {\scriptstyle \pm 1.89}$ & $4.30 {\scriptstyle \pm 1.79}$ & $8.50 {\scriptstyle \pm 2.06}$ & $15.80 {\scriptstyle \pm 7.52}$ & $0.48 {\scriptstyle \pm 0.48}$ & $0.30 {\scriptstyle \pm 0.31}$ & $0.36 {\scriptstyle \pm 0.37}$ \\
 & 0.75 & $4.20 {\scriptstyle \pm 1.89}$ & $4.30 {\scriptstyle \pm 1.79}$ & $8.50 {\scriptstyle \pm 2.06}$ & $15.80 {\scriptstyle \pm 7.52}$ & $0.48 {\scriptstyle \pm 0.48}$ & $0.30 {\scriptstyle \pm 0.31}$ & $0.36 {\scriptstyle \pm 0.37}$ \\
\hline \hline
Asia & & & & & & & & \\
\hline
 & 0.00 & $3.00 {\scriptstyle \pm 0.00}$ & $3.00 {\scriptstyle \pm 0.00}$ & $0.80 {\scriptstyle \pm 0.40}$ & $8.80 {\scriptstyle \pm 1.60}$ & $1.00 {\scriptstyle \pm 0.00}$ & $0.62 {\scriptstyle \pm 0.00}$ & $0.77 {\scriptstyle \pm 0.00}$ \\
 & 0.25 & $3.00 {\scriptstyle \pm 0.00}$ & $3.00 {\scriptstyle \pm 0.00}$ & $0.80 {\scriptstyle \pm 0.40}$ & $8.80 {\scriptstyle \pm 1.60}$ & $1.00 {\scriptstyle \pm 0.00}$ & $0.62 {\scriptstyle \pm 0.00}$ & $0.77 {\scriptstyle \pm 0.00}$ \\
 & 0.50 & $3.10 {\scriptstyle \pm 0.30}$ & $3.20 {\scriptstyle \pm 0.60}$ & $1.70 {\scriptstyle \pm 2.79}$ & $9.70 {\scriptstyle \pm 2.90}$ & $0.98 {\scriptstyle \pm 0.06}$ & $0.61 {\scriptstyle \pm 0.04}$ & $0.75 {\scriptstyle \pm 0.05}$ \\
 & 0.75 & $3.10 {\scriptstyle \pm 0.30}$ & $3.20 {\scriptstyle \pm 0.60}$ & $1.70 {\scriptstyle \pm 2.79}$ & $9.70 {\scriptstyle \pm 2.90}$ & $0.98 {\scriptstyle \pm 0.06}$ & $0.61 {\scriptstyle \pm 0.04}$ & $0.75 {\scriptstyle \pm 0.05}$ \\
\hline \hline
Lucas & & & & & & & & \\
\hline
 & 0.00 & $0.00 {\scriptstyle \pm 0.00}$ & $0.00 {\scriptstyle \pm 0.00}$ & $0.00 {\scriptstyle \pm 0.00}$ & $0.00 {\scriptstyle \pm 0.00}$ & $1.00 {\scriptstyle \pm 0.00}$ & $1.00 {\scriptstyle \pm 0.00}$ & $1.00 {\scriptstyle \pm 0.00}$ \\
 & 0.25 & $0.70 {\scriptstyle \pm 0.46}$ & $0.70 {\scriptstyle \pm 0.46}$ & $0.00 {\scriptstyle \pm 0.00}$ & $4.90 {\scriptstyle \pm 3.21}$ & $1.00 {\scriptstyle \pm 0.00}$ & $0.94 {\scriptstyle \pm 0.04}$ & $0.97 {\scriptstyle \pm 0.02}$ \\
 & 0.50 & $1.60 {\scriptstyle \pm 1.02}$ & $2.40 {\scriptstyle \pm 2.24}$ & $6.60 {\scriptstyle \pm 10.11}$ & $12.20 {\scriptstyle \pm 8.27}$ & $0.93 {\scriptstyle \pm 0.11}$ & $0.87 {\scriptstyle \pm 0.08}$ & $0.90 {\scriptstyle \pm 0.10}$ \\
 & 0.75 & $1.20 {\scriptstyle \pm 0.60}$ & $1.50 {\scriptstyle \pm 1.02}$ & $4.50 {\scriptstyle \pm 6.92}$ & $10.80 {\scriptstyle \pm 7.65}$ & $0.97 {\scriptstyle \pm 0.04}$ & $0.90 {\scriptstyle \pm 0.05}$ & $0.93 {\scriptstyle \pm 0.04}$ \\
\hline \hline
Child & & & & & & & & \\
\hline
 & 0.00 & $3.00 {\scriptstyle \pm 0.00}$ & $3.00 {\scriptstyle \pm 0.00}$ & $0.00 {\scriptstyle \pm 0.00}$ & $44.00 {\scriptstyle \pm 0.00}$ & $1.00 {\scriptstyle \pm 0.00}$ & $0.88 {\scriptstyle \pm 0.00}$ & $0.94 {\scriptstyle \pm 0.00}$ \\
 & 0.25 & $6.30 {\scriptstyle \pm 2.45}$ & $6.30 {\scriptstyle \pm 2.45}$ & $0.00 {\scriptstyle \pm 0.00}$ & $90.20 {\scriptstyle \pm 51.38}$ & $1.00 {\scriptstyle \pm 0.00}$ & $0.75 {\scriptstyle \pm 0.10}$ & $0.85 {\scriptstyle \pm 0.06}$ \\
 & 0.50 & $7.50 {\scriptstyle \pm 2.58}$ & $7.80 {\scriptstyle \pm 2.32}$ & $5.80 {\scriptstyle \pm 17.40}$ & $124.60 {\scriptstyle \pm 59.79}$ & $0.99 {\scriptstyle \pm 0.04}$ & $0.70 {\scriptstyle \pm 0.10}$ & $0.81 {\scriptstyle \pm 0.06}$ \\
 & 0.75 & $9.10 {\scriptstyle \pm 2.51}$ & $9.10 {\scriptstyle \pm 2.51}$ & $0.00 {\scriptstyle \pm 0.00}$ & $137.30 {\scriptstyle \pm 76.75}$ & $1.00 {\scriptstyle \pm 0.00}$ & $0.64 {\scriptstyle \pm 0.10}$ & $0.77 {\scriptstyle \pm 0.07}$ \\
\hline \hline
Alarm & & & & & & & & \\
\hline
 & 0.00 & $4.50 {\scriptstyle \pm 1.28}$ & $4.50 {\scriptstyle \pm 1.28}$ & $18.40 {\scriptstyle \pm 0.80}$ & $36.40 {\scriptstyle \pm 0.80}$ & $0.97 {\scriptstyle \pm 0.02}$ & $0.93 {\scriptstyle \pm 0.01}$ & $0.95 {\scriptstyle \pm 0.01}$ \\
 & 0.25 & $4.90 {\scriptstyle \pm 1.87}$ & $5.60 {\scriptstyle \pm 2.33}$ & $34.00 {\scriptstyle \pm 17.48}$ & $46.60 {\scriptstyle \pm 18.76}$ & $0.96 {\scriptstyle \pm 0.03}$ & $0.92 {\scriptstyle \pm 0.02}$ & $0.94 {\scriptstyle \pm 0.03}$ \\
 & 0.50 & $6.30 {\scriptstyle \pm 2.00}$ & $7.70 {\scriptstyle \pm 2.87}$ & $55.40 {\scriptstyle \pm 24.74}$ & $76.40 {\scriptstyle \pm 28.19}$ & $0.94 {\scriptstyle \pm 0.04}$ & $0.89 {\scriptstyle \pm 0.03}$ & $0.91 {\scriptstyle \pm 0.03}$ \\
 & 0.75 & $8.00 {\scriptstyle \pm 2.49}$ & $10.00 {\scriptstyle \pm 3.82}$ & $69.80 {\scriptstyle \pm 33.01}$ & $99.60 {\scriptstyle \pm 32.56}$ & $0.92 {\scriptstyle \pm 0.05}$ & $0.85 {\scriptstyle \pm 0.04}$ & $0.89 {\scriptstyle \pm 0.04}$ \\
\hline \hline
Insurance & & & & & & & & \\
\hline
 & 0.00 & $18.40 {\scriptstyle \pm 4.15}$ & $22.40 {\scriptstyle \pm 6.55}$ & $250.30 {\scriptstyle \pm 37.97}$ & $259.30 {\scriptstyle \pm 30.59}$ & $0.86 {\scriptstyle \pm 0.08}$ & $0.69 {\scriptstyle \pm 0.06}$ & $0.76 {\scriptstyle \pm 0.07}$ \\
 & 0.25 & $18.70 {\scriptstyle \pm 4.38}$ & $22.90 {\scriptstyle \pm 6.71}$ & $246.60 {\scriptstyle \pm 45.76}$ & $270.10 {\scriptstyle \pm 32.37}$ & $0.85 {\scriptstyle \pm 0.08}$ & $0.68 {\scriptstyle \pm 0.06}$ & $0.76 {\scriptstyle \pm 0.07}$ \\
 & 0.50 & $20.70 {\scriptstyle \pm 3.47}$ & $24.60 {\scriptstyle \pm 5.37}$ & $244.80 {\scriptstyle \pm 63.22}$ & $302.30 {\scriptstyle \pm 36.76}$ & $0.85 {\scriptstyle \pm 0.09}$ & $0.64 {\scriptstyle \pm 0.05}$ & $0.73 {\scriptstyle \pm 0.06}$ \\
 & 0.75 & $21.50 {\scriptstyle \pm 4.43}$ & $25.10 {\scriptstyle \pm 7.08}$ & $241.70 {\scriptstyle \pm 71.67}$ & $320.50 {\scriptstyle \pm 39.15}$ & $0.85 {\scriptstyle \pm 0.10}$ & $0.63 {\scriptstyle \pm 0.07}$ & $0.72 {\scriptstyle \pm 0.08}$ \\
\hline \hline
Hailfinder & & & & & & & & \\
\hline
 & 0.00 & $39.00 {\scriptstyle \pm 15.21}$ & $42.20 {\scriptstyle \pm 15.28}$ & - & - & $0.69 {\scriptstyle \pm 0.12}$ & $0.65 {\scriptstyle \pm 0.12}$ & $0.67 {\scriptstyle \pm 0.12}$ \\
 & 0.25 & $39.00 {\scriptstyle \pm 15.21}$ & $42.20 {\scriptstyle \pm 15.28}$ & - & - & $0.69 {\scriptstyle \pm 0.12}$ & $0.65 {\scriptstyle \pm 0.12}$ & $0.67 {\scriptstyle \pm 0.12}$ \\
 & 0.50 & $39.00 {\scriptstyle \pm 15.21}$ & $42.20 {\scriptstyle \pm 15.28}$ & - & - & $0.69 {\scriptstyle \pm 0.12}$ & $0.65 {\scriptstyle \pm 0.12}$ & $0.67 {\scriptstyle \pm 0.12}$ \\
 & 0.75 & $39.00 {\scriptstyle \pm 15.21}$ & $42.20 {\scriptstyle \pm 15.28}$ & - & - & $0.69 {\scriptstyle \pm 0.12}$ & $0.65 {\scriptstyle \pm 0.12}$ & $0.67 {\scriptstyle \pm 0.12}$ \\
\hline \hline
Hepar2 & & & & & & & & \\
\hline
 & 0.00 & $50.30 {\scriptstyle \pm 1.55}$ & $55.00 {\scriptstyle \pm 2.32}$ & - & - & $0.94 {\scriptstyle \pm 0.02}$ & $0.59 {\scriptstyle \pm 0.01}$ & $0.73 {\scriptstyle \pm 0.01}$ \\
 & 0.25 & $52.80 {\scriptstyle \pm 3.03}$ & $58.80 {\scriptstyle \pm 5.00}$ & - & - & $0.92 {\scriptstyle \pm 0.03}$ & $0.57 {\scriptstyle \pm 0.02}$ & $0.71 {\scriptstyle \pm 0.03}$ \\
 & 0.50 & $54.70 {\scriptstyle \pm 2.37}$ & $61.70 {\scriptstyle \pm 3.07}$ & - & - & $0.91 {\scriptstyle \pm 0.02}$ & $0.56 {\scriptstyle \pm 0.02}$ & $0.69 {\scriptstyle \pm 0.02}$ \\
 & 0.75 & $57.40 {\scriptstyle \pm 3.53}$ & $64.80 {\scriptstyle \pm 3.57}$ & - & - & $0.90 {\scriptstyle \pm 0.01}$ & $0.53 {\scriptstyle \pm 0.03}$ & $0.67 {\scriptstyle \pm 0.02}$ \\
\hline \hline
Win95Pts & & & & & & & & \\
\hline
 & 0.00 & $45.90 {\scriptstyle \pm 5.96}$ & $50.00 {\scriptstyle \pm 6.29}$ & - & - & $0.79 {\scriptstyle \pm 0.03}$ & $0.76 {\scriptstyle \pm 0.03}$ & $0.77 {\scriptstyle \pm 0.03}$ \\
 & 0.25 & $47.90 {\scriptstyle \pm 6.55}$ & $53.20 {\scriptstyle \pm 7.51}$ & - & - & $0.77 {\scriptstyle \pm 0.03}$ & $0.74 {\scriptstyle \pm 0.04}$ & $0.76 {\scriptstyle \pm 0.03}$ \\
 & 0.50 & $50.30 {\scriptstyle \pm 6.99}$ & $56.30 {\scriptstyle \pm 8.22}$ & - & - & $0.76 {\scriptstyle \pm 0.04}$ & $0.72 {\scriptstyle \pm 0.04}$ & $0.74 {\scriptstyle \pm 0.04}$ \\
 & 0.75 & $52.00 {\scriptstyle \pm 5.98}$ & $59.10 {\scriptstyle \pm 6.70}$ & - & - & $0.75 {\scriptstyle \pm 0.03}$ & $0.71 {\scriptstyle \pm 0.03}$ & $0.73 {\scriptstyle \pm 0.03}$ \\
\end{tabular}

}
\caption{\textbf{Reduced-Tags Evaluation Results for Tagged-GES.} Each row reports the mean over 10 seeds. The performance decreases or stays the same with fewer tags.  ${}_{\pm y}$ indicates standard deviation.}
\label{tab:reduced-tags-dataset-metrics}
\end{table*}

\subsection{Further Results on Homogeneity}
\label{sec:homogeneity_all}
We report the plots on tag homogeneity in Fig.~\ref{fig:homogeneity_all} in the same manner as in Sec.~\ref{sec:homogeneity} for all LLMs.
To highlight low-evidence cases in the homogeneity plots, elements supported by only a single edge are marked with an 'X' pattern, while those supported by exactly two edges are marked with diagonal lines ('/').
While tags differ, all LLMs show high levels of homogeneity.
In Tab.~\ref{tab:taggingRelationInspectAll}, we include all tag relations with a positive homogeneity for Claude 4.6 that have at least three edges as evidence (sorting by accuracy and only showing the top 7 relations if there are more).

\begin{figure}[h]
    \centering
    \includegraphics[width=0.9\textwidth]{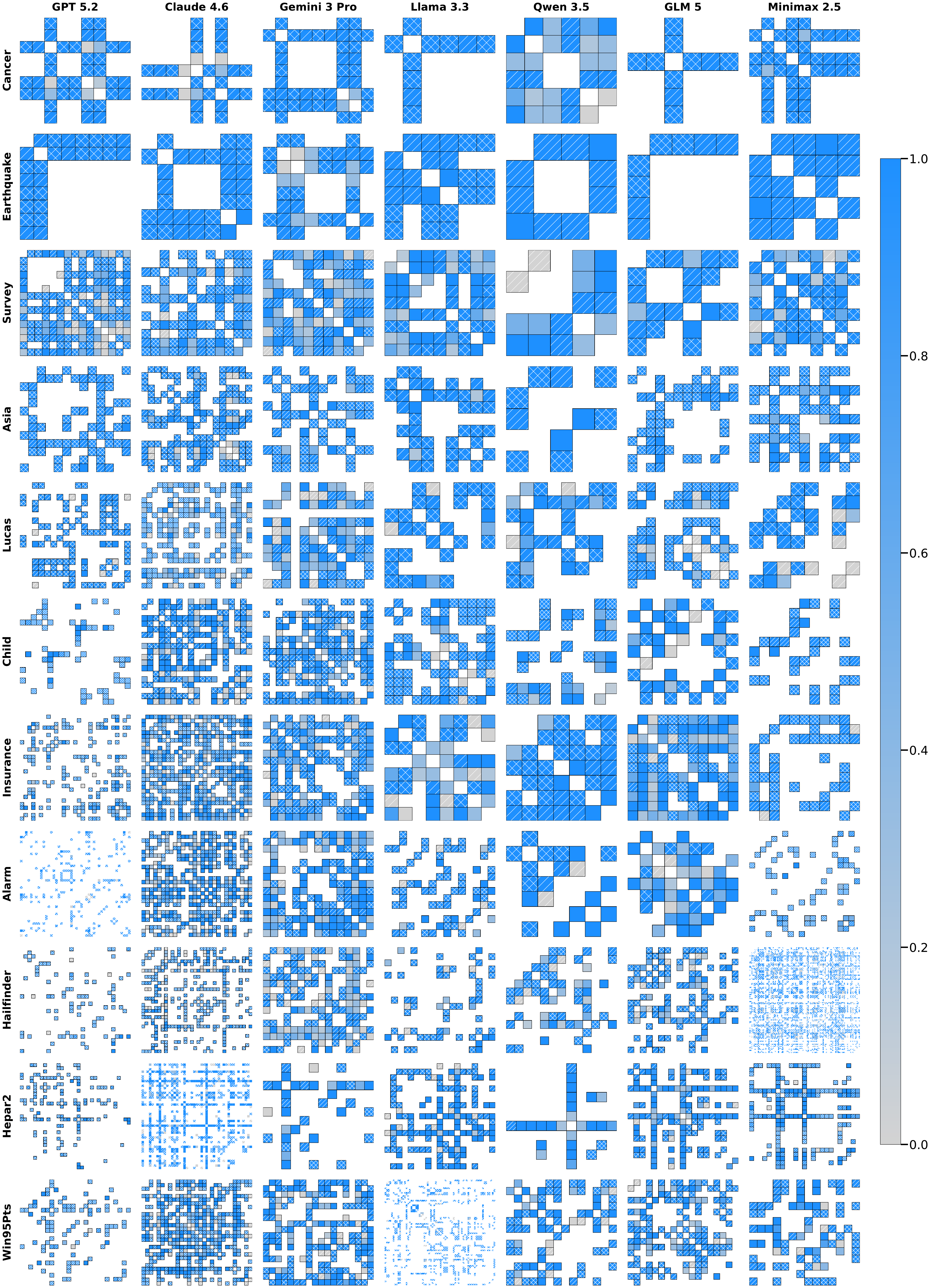}
    
    \caption{\textbf{Tag Homogeneity for all LLMs.} A tag relation has high homogeneity (1) if variables of one tag consistently cause the other and low homogeneity (0) if the cause-effect relationship between these variables is random. We observe high homogeneity throughout all datasets. Hatched areas might appear white on screen due to anti-aliasing.}
    \label{fig:homogeneity_all}
\end{figure}

\begin{table}[t]
    \centering
    \begin{align}
 & \textbf{Cancer} & & \nonumber \\
\text{``respiratory''} &~\rightarrow~ \text{``disease''} && (66\% ~/~ 3~~) \nonumber \\
 & \textbf{Earthquake} & & \nonumber \\
\text{``security\_related''} &~\rightarrow~ \text{``notification''} && (100\% ~/~ 3~~) \nonumber \\
 & \textbf{Survey} & & \nonumber \\
\text{``personal\_attribute''} &~\rightarrow~ \text{``choice\_dependent''} && (100\% ~/~ 4~~) \nonumber \\
\text{``personal\_attribute''} &~\rightarrow~ \text{``socioeconomic''} && (100\% ~/~ 4~~) \nonumber \\
\text{``socioeconomic''} &~\rightarrow~ \text{``geographic''} && (100\% ~/~ 3~~) \nonumber \\
\text{``personal\_attribute''} &~\rightarrow~ \text{``human\_capital''} && (100\% ~/~ 3~~) \nonumber \\
\text{``time\_dependent''} &~\rightarrow~ \text{``choice\_dependent''} && (80\% ~/~ 5~~) \nonumber \\
\text{``choice\_dependent''} &~\rightarrow~ \text{``geographic''} && (75\% ~/~ 4~~) \nonumber \\
\text{``human\_capital''} &~\rightarrow~ \text{``choice\_dependent''} && (75\% ~/~ 4~~) \nonumber \\
 & \textbf{Asia} & & \nonumber \\
\text{``lung-related''} &~\rightarrow~ \text{``downstream effect''} && (100\% ~/~ 4~~) \nonumber \\
\text{``downstream effect''} &~\rightarrow~ \text{``clinical observation''} && (100\% ~/~ 3~~) \nonumber \\
\text{``disease''} &~\rightarrow~ \text{``downstream effect''} && (100\% ~/~ 3~~) \nonumber \\
\text{``exogenous''} &~\rightarrow~ \text{``disease''} && (100\% ~/~ 3~~) \nonumber \\
\text{``exogenous''} &~\rightarrow~ \text{``medical outcome''} && (100\% ~/~ 3~~) \nonumber \\
\text{``exogenous''} &~\rightarrow~ \text{``respiratory''} && (100\% ~/~ 3~~) \nonumber \\
\text{``downstream effect''} &~\rightarrow~ \text{``breathing''} && (66\% ~/~ 3~~) \nonumber \\
 & \textbf{Lucas} & & \nonumber \\
\text{``respiratory''} &~\rightarrow~ \text{``symptom''} && (100\% ~/~ 5~~) \nonumber \\
\text{``heritable''} &~\rightarrow~ \text{``downstream\_effect''} && (100\% ~/~ 4~~) \nonumber \\
\text{``risk\_factor''} &~\rightarrow~ \text{``accident\_related''} && (100\% ~/~ 3~~) \nonumber \\
\text{``disease''} &~\rightarrow~ \text{``downstream\_effect''} && (100\% ~/~ 3~~) \nonumber \\
\text{``disease''} &~\rightarrow~ \text{``symptom''} && (100\% ~/~ 3~~) \nonumber \\
\text{``heritable''} &~\rightarrow~ \text{``symptom''} && (100\% ~/~ 3~~) \nonumber \\
\text{``respiratory''} &~\rightarrow~ \text{``downstream\_effect''} && (85\% ~/~ 7~~) \nonumber \\
 & \textbf{Child} & & \nonumber \\
\text{``disease\_mechanism''} &~\rightarrow~ \text{``pathophysiology''} && (100\% ~/~ 8~~) \nonumber \\
\text{``pathophysiology''} &~\rightarrow~ \text{``measurement''} && (100\% ~/~ 7~~) \nonumber \\
\text{``pathophysiology''} &~\rightarrow~ \text{``oxygenation''} && (100\% ~/~ 7~~) \nonumber \\
\text{``pathophysiology''} &~\rightarrow~ \text{``blood\_gas''} && (100\% ~/~ 6~~) \nonumber \\
\text{``disease\_mechanism''} &~\rightarrow~ \text{``pulmonary''} && (100\% ~/~ 6~~) \nonumber \\
\text{``etiology''} &~\rightarrow~ \text{``disease\_mechanism''} && (100\% ~/~ 5~~) \nonumber \\
\text{``disease\_mechanism''} &~\rightarrow~ \text{``hemodynamic''} && (100\% ~/~ 5~~) \nonumber \\
 & \textit{continued on next page} & & \nonumber %
     \end{align}
\end{table}

\begin{table}[t]
    \centering
    \begin{align}
 & \textbf{Insurance} & & \nonumber \\
\text{``demographic''} &~\rightarrow~ \text{``socioeconomic\_dependent''} && (100\% ~/~ 7~~) \nonumber \\
\text{``exogenous''} &~\rightarrow~ \text{``vehicle\_property''} && (100\% ~/~ 7~~) \nonumber \\
\text{``exogenous''} &~\rightarrow~ \text{``vehicle\_related''} && (100\% ~/~ 7~~) \nonumber \\
\text{``accident\_related''} &~\rightarrow~ \text{``downstream\_consequence''} && (100\% ~/~ 6~~) \nonumber \\
\text{``demographic''} &~\rightarrow~ \text{``age\_dependent''} && (100\% ~/~ 6~~) \nonumber \\
\text{``age\_dependent''} &~\rightarrow~ \text{``risk\_factor''} && (100\% ~/~ 6~~) \nonumber \\
\text{``socioeconomic\_dependent''} &~\rightarrow~ \text{``theft\_related''} && (100\% ~/~ 6~~) \nonumber \\
 & \textbf{Alarm} & & \nonumber \\
\text{``equipment''} &~\rightarrow~ \text{``airway''} && (100\% ~/~ 8~~) \nonumber \\
\text{``root\_cause''} &~\rightarrow~ \text{``latent\_physiology''} && (100\% ~/~ 8~~) \nonumber \\
\text{``cardiac''} &~\rightarrow~ \text{``observed\_indicator''} && (100\% ~/~ 7~~) \nonumber \\
\text{``equipment''} &~\rightarrow~ \text{``measurement''} && (100\% ~/~ 7~~) \nonumber \\
\text{``equipment''} &~\rightarrow~ \text{``observed\_indicator''} && (100\% ~/~ 7~~) \nonumber \\
\text{``pathological\_condition''} &~\rightarrow~ \text{``latent\_physiology''} && (100\% ~/~ 7~~) \nonumber \\
\text{``airway''} &~\rightarrow~ \text{``latent\_physiology''} && (100\% ~/~ 6~~) \nonumber \\
 & \textbf{Hailfinder} & & \nonumber \\
\text{``context''} &~\rightarrow~ \text{``observed\_input''} && (100\% ~/~ 10~~) \nonumber \\
\text{``observed\_input''} &~\rightarrow~ \text{``derived\_intermediate''} && (100\% ~/~ 8~~) \nonumber \\
\text{``observed\_input''} &~\rightarrow~ \text{``combined\_aggregate''} && (100\% ~/~ 6~~) \nonumber \\
\text{``context''} &~\rightarrow~ \text{``scenario\_related''} && (100\% ~/~ 5~~) \nonumber \\
\text{``derived\_intermediate''} &~\rightarrow~ \text{``scenario\_related''} && (100\% ~/~ 4~~) \nonumber \\
\text{``scenario\_related''} &~\rightarrow~ \text{``forecast\_output''} && (100\% ~/~ 4~~) \nonumber \\
\text{``observed\_input''} &~\rightarrow~ \text{``cloud''} && (100\% ~/~ 3~~) \nonumber \\
 & \textbf{Hepar2} & & \nonumber \\
\text{``liver\_disease''} &~\rightarrow~ \text{``laboratory\_value''} && (100\% ~/~ 48~~) \nonumber \\
\text{``chronic\_condition''} &~\rightarrow~ \text{``laboratory\_value''} && (100\% ~/~ 28~~) \nonumber \\
\text{``hepatitis''} &~\rightarrow~ \text{``laboratory\_value''} && (100\% ~/~ 27~~) \nonumber \\
\text{``liver\_disease''} &~\rightarrow~ \text{``liver\_enzyme''} && (100\% ~/~ 19~~) \nonumber \\
\text{``biliary\_system''} &~\rightarrow~ \text{``laboratory\_value''} && (100\% ~/~ 13~~) \nonumber \\
\text{``liver\_disease''} &~\rightarrow~ \text{``cholestatic\_marker''} && (100\% ~/~ 12~~) \nonumber \\
\text{``hepatitis''} &~\rightarrow~ \text{``liver\_enzyme''} && (100\% ~/~ 12~~) \nonumber \\
 & \textbf{Win95Pts} & & \nonumber \\
\text{``root\_cause''} &~\rightarrow~ \text{``intermediate\_state''} && (100\% ~/~ 17~~) \nonumber \\
\text{``physical\_hardware''} &~\rightarrow~ \text{``intermediate\_state''} && (100\% ~/~ 12~~) \nonumber \\
\text{``root\_cause''} &~\rightarrow~ \text{``functional\_status''} && (100\% ~/~ 11~~) \nonumber \\
\text{``root\_cause''} &~\rightarrow~ \text{``data\_flow''} && (100\% ~/~ 8~~) \nonumber \\
\text{``physical\_hardware''} &~\rightarrow~ \text{``functional\_status''} && (100\% ~/~ 8~~) \nonumber \\
\text{``software''} &~\rightarrow~ \text{``intermediate\_state''} && (100\% ~/~ 8~~) \nonumber \\
\text{``physical\_hardware''} &~\rightarrow~ \text{``local\_printing''} && (100\% ~/~ 8~~) \nonumber %
    \end{align}
    \vspace{-25pt}
        \caption{\textbf{Tag Relations for Claude 4.6.} We consider the most informative tag pairs per dataset for Claude 4.6 that have at least 3 edges of evidence. We list tag pair probability and support in brackets, sorted by accuracy.
        (A probability of 50\% corresponds to a tag homogeneity of 0; 100\% indicates full homogeneity.)
        }
    \label{tab:taggingRelationInspectAll}
\end{table}

\begin{table}[t]
\centering
\begin{tabular}{lllc}
Dataset A & Dataset B & Found Relation & Agreement \\
\midrule
Alarm  & Child     & oxygenation $\to$ blood\_gas             & \cmark \\
Alarm  & Child     & oxygenation $\to$ measurement            & \cmark \\
Alarm  & Child     & pulmonary $\to$ blood\_gas               & \cmark \\
Alarm  & Child     & pulmonary $\to$ measurement              & \cmark \\
Lucas  & Cancer    & respiratory - disease                    & \xmark \\
Lucas  & Insurance & risk\_factor $\to$ accident\_related     & \cmark \\
\end{tabular}
\caption{\textbf{Shared Tag Relations.} The table lists all tag relations (support $\geq 3$) shared between any two of the eleven considered datasets under Claude 4.6 tagging annotation. Tags across datasets were matched via exact name match.}
\label{tab:shared-tag-relations}
\end{table}

\subsection{Generality of Shared Tag Relations}
\label{app:shared-tag-relations}

To show the possible generality of tag relations, an additional analysis for finding tag relations that are shared between datasets was run. Even though the 11 datasets span different domains and feature diverse tags due to the open annotation scheme, scanning all of them yielded a small set of six shared tag relations (considering sufficient support of 3+ edges with Claude tagging and exact name matches; individual relations shown in Table~\ref{tab:shared-tag-relations}). All datasets featuring shared relations model medical settings, making shared tags more probable. Of the six relations found, five agreed on their direction across datasets, with the only exception being the ``respiratory -- disease'' relation, which points in different directions depending on the dataset (Cancer: respiratory $\to$ disease; Lucas: disease$\to$respiratory). This also illustrates the primary obstacle to full generalization: depending on whether a domain models the causal direction of a disease causing respiration, or the inference direction of (the lack of) respiration indicating a disease to a medical doctor, both directions may be valid within their respective dataset.

\end{document}